\pdfoutput=1
\PassOptionsToPackage{hidelinks}{hyperref}
\documentclass[11pt]{article}
\usepackage[final]{acl}
\usepackage{times}
\usepackage{latexsym}
\usepackage[T1]{fontenc}
\usepackage[utf8]{inputenc}
\usepackage{microtype}
\usepackage{inconsolata}
\usepackage{graphicx}
\usepackage{booktabs}
\usepackage{makecell}
\usepackage{multirow}
\usepackage{amsmath}
\usepackage{tabularx} 
\usepackage{amssymb}
\usepackage{float}
\usepackage{caption}
\usepackage{titletoc}

\title{Calibrating LLM Confidence by Probing Perturbed\\Representation Stability}

\author{
  \textbf{Reza Khanmohammadi\textsuperscript{\textnormal{1}}\thanks{Corresponding author: \href{mailto:khanreza@msu.edu}{\texttt{khanreza@msu.edu}}}},
  \textbf{Erfan Miahi\textsuperscript{\textnormal{2}}},
  \textbf{Mehrsa Mardikoraem\textsuperscript{\textnormal{2}}},
  \textbf{Simerjot Kaur\textsuperscript{\textnormal{3}}},
\\
  \textbf{Ivan Brugere\textsuperscript{\textnormal{3}}},
  \textbf{Charese H. Smiley\textsuperscript{\textnormal{3}}},
  \textbf{Kundan Thind\textsuperscript{\textnormal{4}}},
  \textbf{Mohammad M. Ghassemi\textsuperscript{\textnormal{1}}}
\\
\\
  \textsuperscript{1}Michigan State University \quad
  \textsuperscript{2}Independent AI Researcher \\
  \textsuperscript{3}JPMorgan AI Research \quad
  \textsuperscript{4}Henry Ford Health
\\
      \small\textsuperscript{1} \texttt{\{khanreza, ghassem3\}@msu.edu} \quad \textsuperscript{2} \texttt{\{mhi.erfan1, mardikorm\}@gmail.com} \\
      \small\textsuperscript{3} \texttt{\{simerjot.kaur, ivan.brugere, charese.h.smiley\}@jpmchase.com} \quad \textsuperscript{4} \texttt{kthind1@hfhs.org}
}

\begin{document}
\maketitle
\begin{abstract}
Miscalibration in Large Language Models (LLMs) undermines their reliability, highlighting the need for accurate confidence estimation. We introduce CCPS (Calibrating LLM Confidence by Probing Perturbed Representation Stability), a novel method analyzing internal representational stability in LLMs. CCPS applies targeted adversarial perturbations to final hidden states, extracts features reflecting the model's response to these perturbations, and uses a lightweight classifier to predict answer correctness. CCPS was evaluated on LLMs from 8B to 32B parameters (covering Llama, Qwen, and Mistral architectures) using MMLU and MMLU-Pro benchmarks in both multiple-choice and open-ended formats. Our results show that CCPS significantly outperforms current approaches. Across four LLMs and three MMLU variants, CCPS reduces Expected Calibration Error by approximately 55\% and Brier score by 21\%, while increasing accuracy by 5 percentage points, Area Under the Precision-Recall Curve by 4 percentage points, and Area Under the Receiver Operating Characteristic Curve by 6 percentage points, all relative to the strongest prior method. CCPS delivers an efficient, broadly applicable, and more accurate solution for estimating LLM confidence, thereby improving their trustworthiness.
\end{abstract}
{\normalsize \textbf{Code} --- {\small \href{https://github.com/ledengary/CCPS}{https://github.com/ledengary/CCPS}}} \\
{\normalsize \textbf{Data} --- {\small \href{https://huggingface.co/datasets/ledengary/CCPS}{https://huggingface.co/datasets/ledengary/CCPS}}}

\section{Introduction}
\label{sec:introduction}
Despite their impressive performance, large language models (LLMs) consistently struggle with confidence calibration \cite{guo,geng-etal-2024-survey}. Their confidence—the model’s internally estimated probability that a given response is correct—frequently misaligns with actual outcomes: LLMs often assign high confidence to wrong answers and low confidence to right ones. This unreliability is particularly acute in high-stakes domains like medicine, finance, and law. For example, in a medical task like symptom extraction for cancer toxicity assessment, even if an LLM often produces correct information, it might do so with inappropriately low confidence, or conversely, express high confidence for incorrect outputs. If such confidence scores are not dependable guides to actual correctness, human experts may be forced to meticulously review every LLM-generated instance, significantly diminishing the practical benefits of automation. Accurate confidence estimation for each specific response is therefore essential, as it provides a vital mechanism for managing risk, enabling users to prioritize human oversight, selectively rely on LLM outputs, and foster more responsible and effective LLM integration.

Existing approaches to LLM confidence estimation include direct self-evaluation \cite{kadavath2022languagemodelsmostlyknow}, post-hoc adjustments \cite{jiang-etal-2021-know}, internal state probing with lightweight classifiers \cite{azaria-mitchell-2023-internal, liu2024litcablightweightlanguagemodel}, and model fine-tuning \cite{kapoor-etal-2024-calibration}. These methods often struggle to consistently deliver on multiple desirable properties simultaneously, namely achieving strong calibration (e.g., low Expected Calibration Error (ECE)) and high discriminative power (e.g., high Area Under the Precision-Recall Curve (AUCPR) or Area Under the Receiver Operating Characteristic Curve (AUROC)) while maintaining computational efficiency and generalizability across the diverse set of LLM architectures and families. Many methods excel in some of these desirable properties but make trade-offs in others; for instance, fine-tuning approaches like Calibration-Tuning (CT) \cite{kapoor-etal-2024-calibration} often achieve strong calibration in ECE but may not consistently lead in discriminative metrics like AUROC, while lightweight methods such as LitCab \cite{liu2024litcablightweightlanguagemodel} can demonstrate strong AUROC but sometimes show variable ECE performance across different LLM families. This leaves a need for more holistically effective solutions.

In this work, we introduce \textbf{CCPS} (\textbf{C}alibrating LLM \textbf{C}onfidence by Probing \textbf{P}erturbed Representation \textbf{S}tability), a novel method that addresses these challenges by assessing LLM confidence through the stability of its internal representations. CCPS operates on frozen base LLMs, applying targeted adversarial perturbations to the final hidden states that generate an answer's tokens. From the LLM's response to these perturbations, we extract a rich feature set and train a lightweight classifier to predict answer correctness. This model-agnostic probing offers an efficient confidence proxy without modifying the base LLM.

Comprehensive evaluations demonstrate CCPS's significant advantages over existing confidence estimation approaches. Tested across four modern LLMs (8B to 32B parameters, spanning three architectural families) on MMLU and MMLU-Pro benchmarks in both multiple-choice and open-ended formats, CCPS consistently achieves superior performance across key calibration (e.g., ECE, Brier score) and discrimination metrics (e.g., ACC, AUCPR, AUROC). Our findings reveal that by quantifying LLM representational stability through targeted internal perturbations, CCPS achieves substantial improvements over other state-of-the-art confidence estimation methods; for instance, CCPS reduces average ECE by approximately 55\% (up to 88\%) and Brier score by 21\% (up to 45\%), while also increasing average Accuracy (ACC) by 5 percentage points (pp) (up to +14 pp), AUCPR by 4 pp (up to +13 pp), and AUROC by 6 pp (up to +17 pp), relative to the best performing baseline. The key contributions of this work include:
\begin{itemize}
    \item A novel, model-agnostic, parameter-efficient, and scalable framework (CCPS) offering a fresh perspective on LLM confidence estimation by quantifying it through the stability of internal representations under targeted perturbations.
    \item Demonstration of CCPS's substantial improvements in both key calibration (ECE, Brier score) and discrimination (ACC, AUCPR, AUROC) metrics.
    \item Evidence of CCPS's generalizability across diverse LLM architectures (Llama, Qwen, Mistral; 8B to 32B).
    \item Extensive benchmarking of confidence estimation methods on MMLU and MMLU-Pro in both multiple-choice and open-ended formats.
\end{itemize}
These contributions establish CCPS as an effective method for improving LLM confidence estimation, helping to make LLM applications more trustworthy, especially in critical domains where reliability is crucial.

\section{Related Work}
\label{sec:related_work}
\noindent\textbf{Calibration in LLMs} \quad A model is considered well-calibrated when its expressed confidence in a prediction aligns with the empirical likelihood of that prediction being correct. In the context of LLMs, calibration efforts broadly diverge into two streams. The first targets calibration of next-token predictions and responses to reduce hallucinations. This direction is exemplified by the work of \citet{zhou2025hademif}, which focuses on hallucination mitigation through comprehensive model calibration. The second stream, more aligned with the present work, focuses on developing and calibrating explicit confidence estimation mechanisms that assess the correctness of statements generated by LLMs.

\vspace{5pt}
\noindent\textbf{Confidence Estimation in LLMs} \quad Several approaches have been proposed for estimating an LLM's confidence in its assertions. One vein of research explores probing the internal states of LLMs. For instance, \citet{azaria-mitchell-2023-internal} train an auxiliary linear classifier on hidden layer activations from an LLM to predict the truthfulness of statements. While this can reveal internal knowledge, its efficacy depends on identifying the optimal representational layer and may vary across evaluation metrics. Another approach involves eliciting the model's inherent self-assessment. \citet{kadavath2022languagemodelsmostlyknow} introduced concepts like P(True), the probability an LLM assigns to its generated answer being correct (often derived from probabilities of ``True'' or ``False'' tokens when prompted to evaluate its own previous answer), and P(IK), the probability the model assigns to its own ability to answer a given question correctly, estimated before attempting to generate the answer. These methods assess the model's intrinsic confidence without external classifiers but rely on the LLM's inherent, and often uncalibrated, self-evaluation capabilities.

\vspace{5pt}
\noindent\textbf{Improving Confidence Calibration in LLMs} \quad Other research adapts the LLM or its outputs to produce more reliable confidence scores. Logit Temperature Scaling (LTS) \cite{jiang-etal-2021-know} is a post-hoc method that adjusts output logits using a learned temperature parameter; however, its performance can degrade under distributional shifts between calibration and test data \cite{kapoor-etal-2024-calibration}. More intensive methods involve fine-tuning. CT \cite{kapoor2024large,kapoor-etal-2024-calibration} builds on the P(True) concept, prompting the LLM to assess its own answers and then fine-tuning it on this self-evaluation using methods like LoRA. This can achieve strong ECE but may face challenges in efficient class discrimination (e.g., AUROC) and can be computationally demanding. In contrast, LitCab \cite{liu2024litcablightweightlanguagemodel} offers a lightweight approach by training a single linear layer to predict a bias term added to the LLM's output logits. While LitCab shows strong discrimination, our experiments reveal variable ECE across LLM families. These diverse strategies highlight an ongoing trade-off in achieving robust calibration, discriminative power, computational efficiency, and generalization in LLM confidence estimation.

\section{Method}
\label{sec:method}
\begin{figure*}[t]
  \centering
  \includegraphics[width=\textwidth]{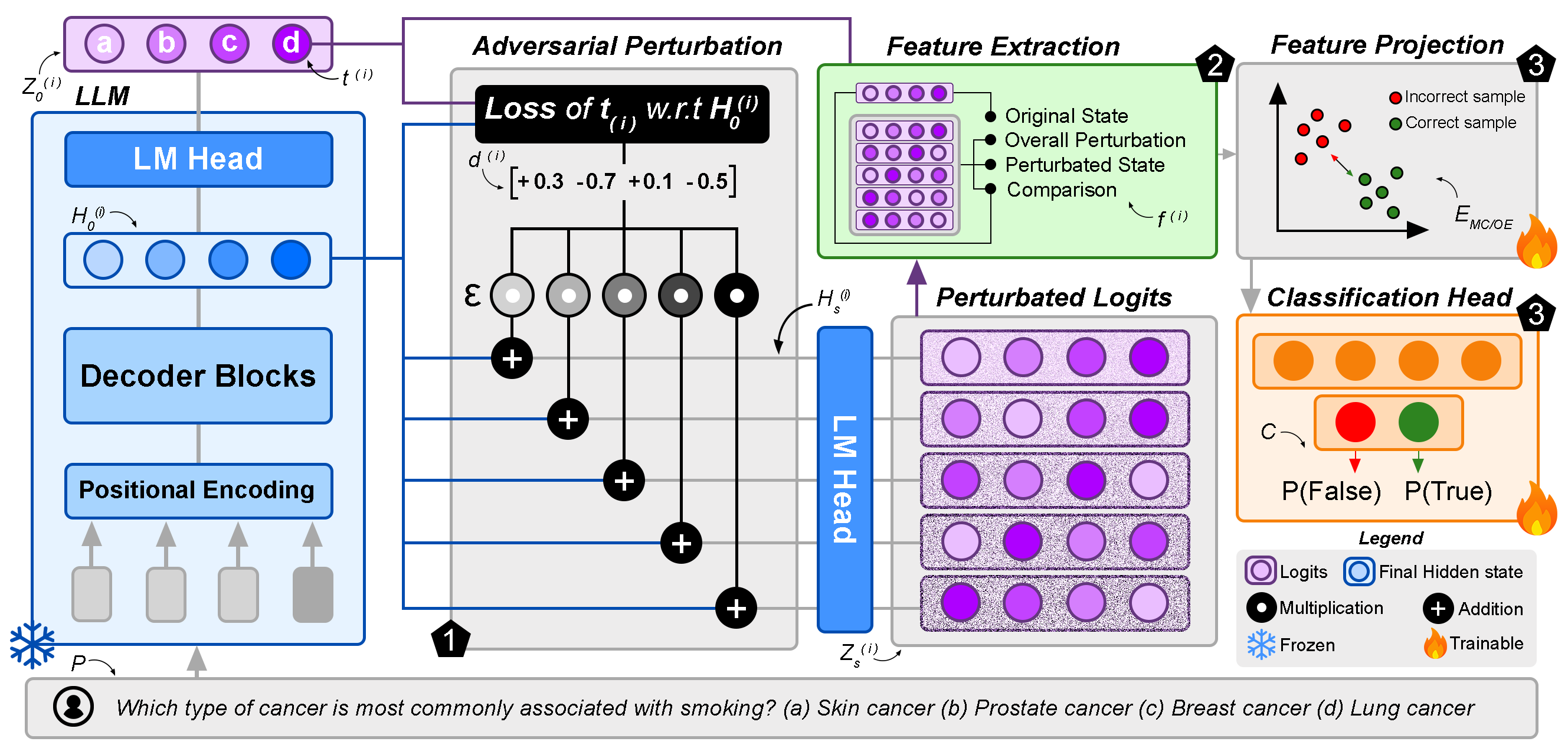}
\caption{Overview of the CCPS method, illustrating its three primary stages. \textbf{(1)} For each token $t_i$ (e.g., 'd' in the example) from a frozen LLM's response to an input prompt $P$, its original final hidden state $H_0^{(i)}$ is systematically perturbed (details in \S\ref{sec:probing_stability}). This involves moving $H_0^{(i)}$ along a derived adversarial direction $d^{(i)}$ with varying magnitudes $\epsilon_s$ (visually represented by lighter to darker shades for increasing $\epsilon_s$) to yield a trajectory of perturbed hidden states $H_s^{(i)}$ and their corresponding logits $Z_s^{(i)}$ via the LM Head. \textbf{(2)} A $D_f$-dimensional feature vector $f^{(i)}$ is then engineered (\S\ref{sec:feature_engineering}), encompassing original state characteristics, perturbation stability indicators, and trajectory divergence statistics derived from the original and perturbed representational data. \textbf{(3)} This feature vector $f^{(i)}$ is subsequently processed by a trainable feature projection network ($E_{\text{MC/OE}}$) and a classification head ($C$) (\S\ref{sec:confidence_classification}) to output the final confidence score, $P(\text{True})$, indicating the likelihood of the LLM's answer being correct.}
\label{fig:CCPS-main-figure}
\end{figure*}

Our approach to LLM confidence estimation is centered on evaluating the internal stability of the model's representations when its generated answer is produced. We hypothesize that an LLM's confidence correlates with the robustness of its internal states; specifically, the final hidden states that lead to the tokens of a high-confidence answer should exhibit greater stability when subjected to targeted perturbations. This internal probing of representational stability offers an efficient alternative to methods relying on multiple generation passes for consistency checking. Notably, output consistency has been identified as a strong indicator of LLM reliability \cite{zhou2025hademif}, but external checks involve significant computational overhead, which our internal analysis aims to mitigate while leveraging a similar underlying principle of stability.

The methodology involves three primary stages, applied while the base LLM (whose confidence is being estimated) remains frozen: (1) token-level adversarial perturbation of the LLM's final hidden states along a defined trajectory, (2) extraction of features that quantify the impact of these perturbations, and (3) a classification architecture that maps these features to a confidence score, representing the answer's probability of correctness. These three stages are illustrated in Figure~\ref{fig:CCPS-main-figure}.

In this work, we use the term \textit{adversarial} in a restricted sense: it refers to targeted, gradient-informed perturbations that are designed to systematically probe representational stability by challenging the generation of token $t_i$. This usage is distinct from adversarial attacks aimed at inducing misclassification with minimal input changes, or from adversarial training schemes intended to improve model robustness.

\subsection{Probing Internal Stability}
\label{sec:probing_stability}
For a given input prompt $P$ (which includes few-shot exemplars and the target question) and an answer $A = (t_1, t_2, \ldots, t_L)$ generated by the base LLM, where $t_i$ is the $i$-th token, we analyze each token individually:

\vspace{5pt}
\noindent\textbf{Original State Identification} \quad For each token $t_i$ in $A$, we first identify the original final hidden state $H_0^{(i)} \in \mathbb{R}^{d_h}$ from the LLM's last transformer layer that immediately led to the generation of $t_i$. This is obtained by feeding $P$ and any preceding generated tokens $t_{<i}$ into the LLM. The corresponding original logits are $Z_0^{(i)} = \text{LM\_Head}(H_0^{(i)})$.

\vspace{5pt}
\noindent\textbf{Adversarial Perturbation Trajectory Direction} \quad To define a systematic perturbation trajectory that challenges the LLM's generation of the \textit{observed token} $t_i$, we utilize the gradient of the loss associated with $t_i$ with respect to its generating hidden state $H_0^{(i)}$. Let $P(t_i | H_0^{(i)})$ be the probability of the token $t_i$ given $H_0^{(i)}$. We define the loss as the negative log-likelihood: $\mathcal{L}^{(i)} = -\log P(t_i | H_0^{(i)})$. The Jacobian vector $J^{(i)} = \nabla_{H_0^{(i)}} \mathcal{L}^{(i)}$ then indicates the direction in the hidden state space where this loss $\mathcal{L}^{(i)}$ increases most rapidly; equivalently, this is the direction where the probability of token $t_i$ decreases most steeply. We normalize this vector to obtain the unit direction $d^{(i)} = J^{(i)} / ||J^{(i)}||_2$. If $J^{(i)}$ is a zero vector, $d^{(i)}$ is also set to zero. Perturbing along this direction $d^{(i)}$ is an adversarial act aimed at making the original token $t_i$ less likely. This contrasts with standard LLM training, where one steps in the negative gradient direction (e.g., $-\nabla \mathcal{L}$) to \textit{reduce} loss for a target token. Here, by moving along the positive gradient of $\mathcal{L}^{(i)}$, we are adversarially probing the stability of the LLM's initial choice $t_i$ by actively trying to dislodge it.

\vspace{5pt}
\noindent\textbf{Iterative Adversarial Perturbation} \quad We then explore the stability of $H_0^{(i)}$ by applying $S$ discrete adversarial perturbations along the direction $d^{(i)}$. The maximum extent of this exploration is defined by a radius $\epsilon_{\text{max}}$ and the number of steps is $S$. We deliberately used a fixed set of hyperparameters ($\epsilon_{\text{max}}=20.0, S=5$) across all models and benchmarks to test the method's robustness. These values were chosen based on empirical observation; $\epsilon_{\text{max}}=20.0$ was a practical upper bound where the integrity of the LLM's output distribution began to degrade into noise or "gibberish" consistently across all models, while $S=5$ was selected as a pragmatic trade-off between capturing the trajectory's dynamics and managing the computational cost. The $s$-th perturbation magnitude is $\epsilon_s = s \cdot (\epsilon_{\text{max}} / S)$, for $s \in \{1, \ldots, S\}$. The $s$-th perturbed hidden state is:
$$H_s^{(i)} = H_0^{(i)} + \epsilon_s \cdot d^{(i)}$$
For each $H_s^{(i)}$, we compute the corresponding perturbed logits $Z_s^{(i)} = \text{LM\_Head}(H_s^{(i)})$. This creates a trajectory of hidden states and their resulting output distributions under these adversarial nudges.

\subsection{Quantifying Perturbation Impact}
\label{sec:feature_engineering}
From the original hidden state $H_0^{(i)}$ and its corresponding logits $Z_0^{(i)}$, along with the trajectory of $S$ perturbed hidden states $\{H_s^{(i)}\}_{s=1}^S$ and their respective logits $\{Z_s^{(i)}\}_{s=1}^S$, we extract a $D_f$-dimensional feature vector $f^{(i)}$ for each token $t_i$. These features are designed to capture the LLM's initial output characteristics for $t_i$ and how these characteristics evolve under systematic adversarial perturbation. Detailed definitions of all features are provided in Appendix \ref{app:feature_set_description}. The primary categories are:

\vspace{5pt}
\noindent\textbf{Original State Features} \quad This set quantifies the LLM's baseline predictive characteristics for token $t_i$ prior to any perturbation, including measures of output probabilities, logits, distribution entropy, and prediction margins.

\vspace{5pt}
\noindent\textbf{Overall Perturbation Features} \quad This category comprises scalar metrics reflecting key aspects of the perturbation process itself or its integrated effects, such as the L2 norm of the Jacobian vector $J^{(i)}$, the perturbation magnitude required to change the LLM's top predicted token from $t_i$ (epsilon-to-flip), and the Perturbation Energy Integral (PEI) value which summarizes the impact of perturbations on the log-probability of $t_i$.

\vspace{5pt}
\noindent\textbf{Perturbed State Features} \quad These features consist of statistical summaries (e.g., mean, standard deviation, min, max, across the $S$ perturbation steps) of the LLM's output characteristics (such as token probabilities and distribution entropy) evaluated \textit{after} its hidden states have been perturbed.

\vspace{5pt}
\noindent\textbf{Comparison Features} \quad This group includes statistical summaries of metrics that quantify the differences or relationships (e.g., distributional divergences like Kullback–Leibler and Jensen–Shannon, cosine similarities) between the LLM's original state (hidden states, logits, probability distributions) and its perturbed states across the trajectory.

\vspace{5pt}
\subsection{Confidence Classification Architecture}
\label{sec:confidence_classification}
The per-token feature vectors serve as input to a neural network designed to predict the correctness of the entire answer $A$. This architecture comprises a feature projection network and a classification head.

\vspace{5pt}
\noindent\textbf{Feature Projection Network} \quad The network structure adapts to the answer format. For Multiple-Choice (MC) answers, which are typically single-token responses, the feature vector $f^{(1)}$ is processed by a Multi-Layer Perceptron (MLP), denoted as $E_{\text{MC}}$, to yield an embedding $e = E_{\text{MC}}(f^{(1)})$. In contrast, for Open-Ended (OE) answers consisting of $L$ tokens, the sequence of feature vectors $(f^{(1)}, \ldots, f^{(L)})$ is passed through an encoder $E_{\text{OE}}$ composed of 1D convolutional layers and adaptive pooling, resulting in a sequence embedding $e = E_{\text{OE}}(f^{(1)}, \ldots, f^{(L)})$. 

Both $E_{\text{MC}}$ and $E_{\text{OE}}$ are pre-trained using a Max-Margin contrastive loss. Specifically, given a correct answer embedding $e^{+}$ and an incorrect answer embedding $e^{-}$ for the same question, the loss encourages the distance between $e^{+}$ and $e^{-}$ to exceed a margin $\gamma > 0$, formulated as

{\small
\[
\mathcal{L}_{\text{max-margin}} = \max\!\big(0, \, \gamma - (\| e^{+} - e^{-} \|_2 - \| e^{+} - e^{+} \|_2 )\big)
\]
}

This objective pushes embeddings of correct answers closer together while enforcing separation from incorrect ones. The choice of loss is aimed at learning discriminative embeddings, a strategy also found effective in other confidence estimation works such as \citet{liu2024litcablightweightlanguagemodel}. The objective of this pre-training is to map features from correctly answered questions to regions in the embedding space that are separable from those associated with incorrect answers, supervised by the ground truth correctness of $A$.

\vspace{5pt}
\noindent\textbf{Classification Head} \quad The embedding $e$ is then passed to an MLP classification head, $C$. This head outputs a 2-dimensional logit vector, $Z_{\text{conf}} = C(e)$. This architectural choice for binary correctness prediction (incorrect vs. correct) is similar to that used by \citet{kapoor-etal-2024-calibration}. The final confidence score, $P(\text{correct}|A)$, is obtained via a softmax function applied to $Z_{\text{conf}}$.

\vspace{5pt}
\noindent\textbf{Training Procedure} \quad Following the contrastive pre-training of the projection network, the projection network ($E_{\text{MC}}$ or $E_{\text{OE}}$) and the classification head $C$ are jointly fine-tuned. This stage employs a standard cross-entropy loss, again supervised by the ground truth correctness of answer $A$. Further implementation details of both $E_{\text{MC}}$ and $E_{\text{OE}}$,
including the results of our hyperparameter search and the finalized layer
configurations, are provided in Appendix~\ref{app:ccps_architecture}.

\section{Experimental Setup}
\label{sec:experiments}
This section details the experimental setup designed to empirically evaluate CCPS. Our evaluation framework provides a comprehensive and rigorous comparison, benchmarking CCPS against a wide array of recent confidence estimation methods. This benchmark is conducted across four modern LLMs of varying architectures and scales, on knowledge-based question-answering (QA) datasets including MMLU and MMLU-Pro in both multiple-choice and open-ended formats. The following subsections detail the specific language models, datasets, training configurations, baselines, and evaluation metrics employed in our study.

\vspace{5pt}
\noindent\textbf{Datasets} \quad For training and validating our confidence estimation models, we utilize the \texttt{CT-CHOICE} and \texttt{CT-OE} datasets for multiple-choice and open-ended QA formats, respectively. These datasets, generated following the exact methodology detailed by \citet{kapoor-etal-2024-calibration} (Apache License 2.0), comprise a large collection of commonly used QA datasets from the literature. To assess generalization and performance, we evaluate on tasks from the Massive Multitask Language Understanding (MMLU) benchmark \cite{mmlu} (MIT License). We created multiple-choice and open-ended versions of these tasks, namely \texttt{MMLU-CHOICE} and \texttt{MMLU-OE}, using the same data processing approach as \citet{kapoor-etal-2024-calibration} to ensure consistency. Additionally, we employ \texttt{MMLU-PRO-CHOICE} (Apache License 2.0), a multiple-choice version of the MMLU-Pro dataset \cite{mmlupro}, for further rigorous testing. All dataset instances across training, validation, and testing incorporate 5-shot exemplars within the input prompt $P$ to contextualize the LLMs. Additional details on the dataset characteristics, response generation process, and labeling procedures are provided in Appendix~\ref{app:datasets}.

\vspace{5pt}
\noindent\textbf{Training Details} \quad The full architectural details of our projection networks and classification head are provided in Appendix~\ref{app:ccps_architecture}. To ensure fair comparisons, training configurations were kept consistent across all methods, including baselines. The main classification/fine-tuning stage for all models involved a total of 10,000 training steps. For our proposed method, the contrastive feature projection network ($E_{\text{MC}}$ or $E_{\text{OE}}$) was pre-trained for 5,000 steps. Subsequently, the confidence classification model was trained for an additional 5,000 steps. Key hyperparameters for the AdamW optimizer \cite{loshchilov2019decoupledweightdecayregularization}, such as a learning rate of $1 \times 10^{-4}$, were aligned with those reported by \citet{kapoor-etal-2024-calibration}. Training was conducted with a batch size of 32. A weight decay of 0.1 was uniformly applied across all training stages and methods. 

\vspace{5pt}
\noindent\textbf{Baselines} \quad We compare our method (CCPS) against a comprehensive set of established confidence estimation techniques. These include P(True) and P(IK) \cite{kadavath2022languagemodelsmostlyknow}, Logit Temperature Scaling (LTS) \cite{jiang-etal-2021-know}, Instruction Tuning (IT) \cite{flan1} on the uncertainty query, SAPLMA \cite{azaria-mitchell-2023-internal} (with variants SAPLMA-F, SAPLMA-M, and SAPLMA-UM corresponding to different layer inputs), Calibration Tuning (CT) \cite{kapoor-etal-2024-calibration}, and LitCab \cite{liu2024litcablightweightlanguagemodel}. Detailed descriptions of these baseline methods are provided in Appendix~\ref{app:baselines_desc}. Information regarding the computational setup and resources utilized for all methods is available in Appendix~\ref{app:computational_resources}. Furthermore, a comparative analysis of the additional trainable parameters introduced by each method is presented in Appendix~\ref{app:appendix_params}, underscoring the parameter efficiency of our CCPS approach.

\vspace{5pt}
\noindent\textbf{Evaluation Metrics} \quad We focus on two primary metrics in the main text. For calibration, we use the Expected Calibration Error (ECE), which measures how well predicted confidences align with actual accuracies. Intuitively, if a model assigns 70\% confidence to a set of answers, then about 70\% of those answers should be correct. To compute this, the $n$ samples are partitioned into $b=10$ equally spaced bins $\{B_j\}_{j=1}^b$, and we compare the average predicted confidence $\text{conf}(B_j)$ against the empirical accuracy $\text{acc}(B_j)$ within each bin:
\[
\text{ECE} = \sum_{j=1}^{b} \frac{|B_j|}{n} \, \big|\text{conf}(B_j) - \text{acc}(B_j)\big|.
\]
Smaller values indicate that predicted probabilities more faithfully reflect true correctness rates.  

In addition, we report the Brier Score, which directly measures the squared error between the predicted confidence $p_k$ and the ground-truth outcome $o_k \in \{0,1\}$ for each sample $k$:
\[
\text{Brier} = \frac{1}{n} \sum_{k=1}^{n} (p_k - o_k)^2.
\]
This metric captures both calibration and the sharpness of predictions, with lower scores reflecting more reliable and informative confidence estimates.  

Additional classification-oriented metrics—including ACC, AUCPR, and AUROC—are reported and formally defined in Appendix~\ref{app:evaluation_metrics}.

\vspace{5pt}
\noindent\textbf{Scientific Artifacts} \quad A detailed discussion regarding the scientific artifacts utilized and developed in this study, including our adherence to their intended use and the intended applications of our created artifacts, can be found in Appendix~\ref{app:artifacts}.

\section{Results}
\label{sec:results}
The performance of CCPS compared to baseline methods across different LLMs and MMLU benchmark variants is presented in Table \ref{tab:performance_metrics}. Our method, CCPS, consistently demonstrates notable improvements in both calibration and discriminative power.

On the standard multiple-choice benchmark, \texttt{MMLU-CHOICE}, CCPS consistently achieves superior performance across all four base LLMs. For instance, ECE scores for CCPS are typically in the range of 5.8-6.5\%, representing substantial reductions compared to both LitCab and CT, which often exhibit much higher ECEs (e.g., LitCab's ECE of 45.6\% and CT's 45.2\% on \texttt{Qwen2.5-14B} and \texttt{Qwen2.5-32B} respectively, against CCPS's 6.3\% on both). CCPS shows similar gains in Brier score and discriminative metrics like AUCPR and AUROC, often matching or outperforming baselines.

\begin{table*}[t]
\centering
\resizebox{\textwidth}{!}{
\begin{tabular}{lccccc|lccccc}
\toprule
\multicolumn{12}{c}{\textbf{\textit{MMLU-CHOICE}}} \\
\midrule
\multicolumn{6}{c}{\textit{Meta-Llama-3.1-8B-Instruct}} & \multicolumn{6}{c}{\textit{Qwen2.5-14B-Instruct}} \\
\textbf{\textit{Method}} & \textbf{\textit{ECE $\downarrow$}} & \textbf{\textit{BRIER $\downarrow$}} & \textbf{\textit{ACC $\uparrow$}} & \textbf{\textit{AUCPR $\uparrow$}} & \textbf{\textit{AUROC $\uparrow$}} &
\textbf{\textit{Method}} & \textbf{\textit{ECE $\downarrow$}} & \textbf{\textit{BRIER $\downarrow$}} & \textbf{\textit{ACC $\uparrow$}} & \textbf{\textit{AUCPR $\uparrow$}} & \textbf{\textit{AUROC $\uparrow$}} \\
\midrule
LitCab & 10.9 & 18.1 & 73.2 & 84.0 & \textbf{77.1} &
LitCab & 45.6 & 20.0 & 78.3 & 83.7 & 65.3 \\
CT & 10.7 & 21.1 & 67.8 & 74.2 & 62.8 &
CT & 12.1 & 17.0 & 78.6 & 84.7 & 64.8 \\
\textbf{CCPS} & \textbf{6.5} & \textbf{17.1} & \textbf{73.4} & \textbf{84.1} & \textbf{77.1} &
\textbf{CCPS} & \textbf{6.3} & \textbf{13.1} & \textbf{80.2} & \textbf{92.1} & \textbf{81.6} \\
\midrule
\multicolumn{6}{c}{\textit{Mistral-Small-24B-Instruct-2501}} & \multicolumn{6}{c}{\textit{Qwen2.5-32B-Instruct}} \\
\textbf{\textit{Method}} & \textbf{\textit{ECE $\downarrow$}} & \textbf{\textit{BRIER $\downarrow$}} & \textbf{\textit{ACC $\uparrow$}} & \textbf{\textit{AUCPR $\uparrow$}} & \textbf{\textit{AUROC $\uparrow$}} &
\textbf{\textit{Method}} & \textbf{\textit{ECE $\downarrow$}} & \textbf{\textit{BRIER $\downarrow$}} & \textbf{\textit{ACC $\uparrow$}} & \textbf{\textit{AUCPR $\uparrow$}} & \textbf{\textit{AUROC $\uparrow$}} \\
\midrule
LitCab & 13.5 & 15.1 & 79.5 & 91.5 & 78.2 &
LitCab & 43.2 & 15.9 & 82.6 & 87.9 & 67.2 \\
CT & 8.2 & 15.5 & 79.6 & 83.3 & 56.5 &
CT & 45.2 & 46.9 & 37.2 & 84.3 & 51.6 \\
\textbf{CCPS} & \textbf{5.8} & \textbf{11.5} & \textbf{83.0} & \textbf{93.1} & \textbf{83.3} &
\textbf{CCPS} & \textbf{6.3} & \textbf{10.8} & \textbf{84.1} & \textbf{94.1} & \textbf{82.8} \\
\midrule\midrule

\multicolumn{12}{c}{\textbf{\textit{MMLU-PRO-CHOICE}}} \\
\midrule
\multicolumn{6}{c}{\textit{Meta-Llama-3.1-8B-Instruct}} & \multicolumn{6}{c}{\textit{Qwen2.5-14B-Instruct}} \\
\textbf{\textit{Method}} & \textbf{\textit{ECE $\downarrow$}} & \textbf{\textit{BRIER $\downarrow$}} & \textbf{\textit{ACC $\uparrow$}} & \textbf{\textit{AUCPR $\uparrow$}} & \textbf{\textit{AUROC $\uparrow$}} &
\textbf{\textit{Method}} & \textbf{\textit{ECE $\downarrow$}} & \textbf{\textit{BRIER $\downarrow$}} & \textbf{\textit{ACC $\uparrow$}} & \textbf{\textit{AUCPR $\uparrow$}} & \textbf{\textit{AUROC $\uparrow$}} \\
\midrule
LitCab & 16.6 & 24.7 & 66.1 & 51.7 & 63.6 &
LitCab & 49.7 & 38.3 & 55.3 & 66.2 & 68.0 \\
CT & 21.5 & 29.8 & 50.4 & 43.7 & 57.3 &
CT & 20.4 & 28.7 & 55.6 & 59.4 & 56.6 \\
\textbf{CCPS} & \textbf{4.5} & \textbf{20.0} & \textbf{70.4} & \textbf{55.2} & \textbf{67.9} &
\textbf{CCPS} & \textbf{4.2} & \textbf{20.1} & \textbf{69.2} & \textbf{75.8} & \textbf{74.0} \\
\midrule
\multicolumn{6}{c}{\textit{Mistral-Small-24B-Instruct-2501}} & \multicolumn{6}{c}{\textit{Qwen2.5-32B-Instruct}} \\
\textbf{\textit{Method}} & \textbf{\textit{ECE $\downarrow$}} & \textbf{\textit{BRIER $\downarrow$}} & \textbf{\textit{ACC $\uparrow$}} & \textbf{\textit{AUCPR $\uparrow$}} & \textbf{\textit{AUROC $\uparrow$}} &
\textbf{\textit{Method}} & \textbf{\textit{ECE $\downarrow$}} & \textbf{\textit{BRIER $\downarrow$}} & \textbf{\textit{ACC $\uparrow$}} & \textbf{\textit{AUCPR $\uparrow$}} & \textbf{\textit{AUROC $\uparrow$}} \\
\midrule
LitCab & 32.2 & 34.6 & 57.0 & 66.2 & 60.1 &
LitCab & 48.4 & 33.7 & 60.8 & 72.7 & 70.3 \\
CT & 17.8 & 27.4 & 58.2 & 60.1 & 54.3 &
CT & 38.0 & 41.6 & 44.8 & 60.5 & 49.9 \\
\textbf{CCPS} & \textbf{4.5} & \textbf{18.6} & \textbf{71.3} & \textbf{79.5} & \textbf{77.2} &
\textbf{CCPS} & \textbf{4.6} & \textbf{18.5} & \textbf{71.8} & \textbf{82.4} & \textbf{77.8} \\
\midrule\midrule

\multicolumn{12}{c}{\textbf{\textit{MMLU-OE}}} \\
\midrule
\multicolumn{6}{c}{\textit{Meta-Llama-3.1-8B-Instruct}} & \multicolumn{6}{c}{\textit{Qwen2.5-14B-Instruct}} \\
\textbf{\textit{Method}} & \textbf{\textit{ECE $\downarrow$}} & \textbf{\textit{BRIER $\downarrow$}} & \textbf{\textit{ACC $\uparrow$}} & \textbf{\textit{AUCPR $\uparrow$}} & \textbf{\textit{AUROC $\uparrow$}} &
\textbf{\textit{Method}} & \textbf{\textit{ECE $\downarrow$}} & \textbf{\textit{BRIER $\downarrow$}} & \textbf{\textit{ACC $\uparrow$}} & \textbf{\textit{AUCPR $\uparrow$}} & \textbf{\textit{AUROC $\uparrow$}} \\
\midrule
LitCab & 8.8 & 22.5 & 65.3 & 46.2 & 66.0 &
LitCab & 34.4 & 37.0 & 49.4 & 56.8 & 62.5 \\
CT & 8.8 & 21.1 & 65.3 & 48.9 & \textbf{70.9} &
CT & 9.4 & 22.6 & 63.4 & \textbf{61.7} & \textbf{69.3} \\
\textbf{CCPS} & \textbf{8.0} & \textbf{20.2} & \textbf{69.5} & \textbf{49.4} & 69.3 &
\textbf{CCPS} & \textbf{6.7} & \textbf{22.5} & \textbf{63.6} & 59.0 & 66.6 \\
\midrule
\multicolumn{6}{c}{\textit{Mistral-Small-24B-Instruct-2501}} & \multicolumn{6}{c}{\textit{Qwen2.5-32B-Instruct}} \\
\textbf{\textit{Method}} & \textbf{\textit{ECE $\downarrow$}} & \textbf{\textit{BRIER $\downarrow$}} & \textbf{\textit{ACC $\uparrow$}} & \textbf{\textit{AUCPR $\uparrow$}} & \textbf{\textit{AUROC $\uparrow$}} &
\textbf{\textit{Method}} & \textbf{\textit{ECE $\downarrow$}} & \textbf{\textit{BRIER $\downarrow$}} & \textbf{\textit{ACC $\uparrow$}} & \textbf{\textit{AUCPR $\uparrow$}} & \textbf{\textit{AUROC $\uparrow$}} \\
\midrule
LitCab & 11.2 & 24.6 & 60.2 & 60.5 & 66.4 &
LitCab & 28.4 & 33.2 & 52.7 & 60.2 & 62.3 \\
CT & 10.8 & 22.8 & 62.2 & 60.7 & 68.2 &
CT & 22.9 & 31.1 & 57.1 & 52.9 & 56.3 \\
\textbf{CCPS} & \textbf{6.8} & \textbf{20.8} & \textbf{67.6} & \textbf{64.7} & \textbf{71.4} &
\textbf{CCPS} & \textbf{8.7} & \textbf{23.3} & \textbf{62.6} & \textbf{62.0} & \textbf{66.4} \\
\bottomrule
\end{tabular}
}
\caption{Average performance on MMLU variants across tasks per LLM. Arrows indicate whether lower ($\downarrow$) or higher ($\uparrow$) values are better. All values are percentages. Best values per method-block are bolded. For brevity, only the two best-performing baselines are shown here; full results are provided in Appendix~\ref{app:extended_results}.}
\label{tab:performance_metrics}
\end{table*}

When evaluated on the more challenging \texttt{MMLU-PRO-CHOICE} dataset, CCPS further extends its performance advantages, particularly in calibration. CCPS consistently achieves ECE values around 4.5\% across all tested LLMs, a significant improvement over LitCab (ECEs ranging from 16.6\% to 49.7\%) and CT (ECEs from 17.8\% to 38.0\%). This strong calibration is paired with top scores in Brier, ACC, AUCPR, and AUROC, showing CCPS’s robustness on more difficult questions. For example, with \texttt{Mistral-24B}, CCPS records an ECE of 4.5\% and an AUROC of 77.2\%, compared to LitCab's 32.2\% ECE and 60.1\% AUROC, and CT's 17.8\% ECE and 54.3\% AUROC.

In the open-ended generation setting (\texttt{MMLU-OE}), CCPS generally maintains strong calibration, consistently achieving the best ECE and Brier scores, especially with larger models like \texttt{Mistral-24B} and \texttt{Qwen2.5-32B} where it leads across all metrics. For smaller models on \texttt{MMLU-OE}, while CCPS leads in calibration, CT demonstrates competitive discriminative performance in AUCPR and AUROC (e.g., for \texttt{Llama-3.1-8B}, CT's AUROC is 70.9\% vs. CCPS's 69.3\%; for \texttt{Qwen2.5-14B}, CT leads in AUCPR and AUROC). However, CCPS's calibration advantage remains evident, for example, achieving an ECE of 6.7\% with \texttt{Qwen2.5-14B} compared to CT's 9.4\%. To further assess the cross-domain robustness of CCPS, we conducted an additional evaluation on the specialized MedMCQA benchmark; these results are detailed in Appendix~\ref{app:medmcqa_results}.

In summary, CCPS consistently delivers substantial improvements in confidence estimation, excelling in both calibration and the ability to discriminate between correct and incorrect responses across diverse LLMs and task formats, particularly on challenging multiple-choice benchmarks. The findings in Table \ref{tab:performance_metrics} are further detailed in Appendix \ref{app:extended_results}, which includes comprehensive results for all baselines (mean and standard deviation scores, comparative bar charts, per-task breakdowns, and feature importance analyses).

\section{Discussion}
\label{sec:discussion}
\noindent\textbf{CCPS Excels in Both Calibration and Discrimination.} \quad A significant finding is the ability of CCPS to simultaneously achieve strong calibration (low ECE and Brier scores) and high discriminative power (high AUCPR and AUROC), as evidenced in Table \ref{tab:performance_metrics}. This contrasts with observations for some baselines; for instance, while LitCab often demonstrates good discrimination, its ECE can be variable, particularly with certain LLM families (e.g., Qwen models). Conversely, Calibration Tuning (CT) generally achieves good ECE but can lag in discriminative metrics compared to CCPS. Our method's dual strength suggests that the features extracted from internal perturbation trajectories effectively capture signals relevant to both the reliability and the correctness of an LLM's answer.

\vspace{5pt}
\noindent\textbf{The CCPS Framework Provides an Efficient and Scalable Approach to Confidence Estimation.} \quad CCPS is designed to be lightweight. Once features are extracted, the confidence estimation model itself consists of relatively small MLPs or CNNs (as detailed in Appendix \ref{app:ccps_architecture}), making its training and inference efficient. Specifically for our OE models, the convolutional architecture proves to be an effective design choice; a Token Masking Impact Analysis (detailed in Appendix \ref{app:token_importance}) demonstrates that the model successfully learns to prioritize semantically meaningful tokens by being significantly more sensitive to content words than to grammatical filler words, all within a compact architecture. This efficiency contrasts sharply with methods like CT, which, despite using LoRA, require fine-tuning larger portions of the base LLM and can be resource-intensive (e.g., CT reportedly takes $\sim$4 GPU days on an NVIDIA V100). Furthermore, CCPS avoids some scalability concerns present in other methods. For example, LitCab's projection layer size (hidden\_dim $\times$ vocabulary\_size) can become very large for LLMs with extensive vocabularies, and its reliance on multiple negative samples per question for its contrastive learning imposes specific data curation requirements. CCPS, on the other hand, uses more compact projection networks and only requires labels of correctness for the LLM's generated answers.

\vspace{5pt}
\noindent\textbf{Probing Internal Representational Stability Forms the Core of CCPS’s Mechanism.}
The methodological foundation of CCPS lies in quantifying internal consistency. Prior work has shown that external output consistency is a useful reliability signal—for example, generating multiple responses and measuring consensus, as in the Self-Consistency (SC) method~\cite{selfconsistency}. Such approaches, however, are computationally costly because they require repeated full-generation passes \cite{zhou2025hademif}. CCPS internalizes this idea by perturbing hidden state representations instead. The premise is that if an LLM is truly confident, its internal decision-making process for a token should remain stable under small, targeted perturbations. Our results suggest that features derived from this stability serve as effective proxies for confidence. Furthermore, our direct empirical comparison in Appendix~\ref{app:self_consistency_comparison} shows that CCPS consistently outperforms SC, particularly in calibration, validating our approach as a more efficient and effective way to measure consistency. To further isolate the source of these gains, Appendix~\ref{app:disentangling_feature_contributions} presents a detailed ablation study contrasting feature sets from unperturbed states, perturbation-derived features, and their combination. The results confirm that perturbation features are the dominant driver of CCPS’s strong performance, especially on more difficult tasks.

\vspace{5pt}
\noindent\textbf{Perturbation-Derived Features Offer Key Insights into LLM Confidence Signals.} \quad The SHAP value analyses (Appendix \ref{app:shap} provide insights into which features derived from our perturbation process are most influential. Consistently across different LLMs and datasets, the \textit{original entropy} of the LLM's output distribution for a token emerges as an important feature. As expected, higher original entropy typically shows a negative correlation with the prediction of correctness (meaning higher entropy contributes to predicting the answer as incorrect), signifying that greater initial uncertainty in the LLM's choice is indicative of a potentially incorrect answer. More revealingly, many of the top-ranking features are those derived from the \textit{perturbed states}. For instance, the \textit{margin between the logits of the top-ranked and second-ranked tokens after perturbation} often shows a positive correlation with correctness; a larger margin, even under adversarial stress, indicates a more decisive and less ambiguous output from the LLM, which CCPS learns as a sign of confidence. Similarly, a higher \textit{epsilon-to-flip} value, indicating that a larger perturbation magnitude is needed to make the LLM change its predicted token, consistently contributes positively to the confidence score. These findings affirm that the dynamic response to perturbation, not just the initial state, provides critical signals for confidence estimation. To further validate these findings and rigorously quantify the importance of each feature, we conducted a comprehensive leave-one-out ablation study, detailed in Appendix \ref{app:loo_feature_ablation}.

\vspace{5pt}
\noindent\textbf{CCPS Demonstrates Consistent Efficacy Across Diverse LLM Architectures.} \quad The strong performance of CCPS is not confined to a specific model architecture or size, as it demonstrates effectiveness across \texttt{Llama}, \texttt{Qwen}, and \texttt{Mistral} families (8B to 32B parameters). This consistency, particularly when compared to methods like LitCab which showed variable ECE performance across LLM families in our experiments (Table \ref{tab:performance_metrics}), suggests that the feature set derived from our internal perturbation methodology captures fundamental aspects of LLM decision-making relevant to confidence, regardless of the specific base model.

\section{Conclusion}
\label{sec:conclusion}
In this work, we introduced CCPS, a novel method for estimating LLM confidence by evaluating the stability of their internal representations when subjected to targeted adversarial perturbations, using features derived from this process with a lightweight classifier. Our approach demonstrated significant improvements over existing methods, consistently achieving superior calibration (measured by ECE and Brier scores) and discriminative ability (evidenced by strong AUCPR and AUROC results). This effectiveness was observed across a diverse range of LLMs, various MMLU and MMLU-Pro task formats (including multiple-choice and open-ended question answering), and differing levels of difficulty. The features derived from the LLM's response to adversarial nudges proved highly indicative of confidence. CCPS offers an effective and lightweight way to assess LLM reliability, requiring no changes to generation or extensive fine-tuning, and marks a promising step toward more trustworthy, interpretable systems.

\section*{Limitations}
Despite its strong performance, CCPS has limitations. Firstly, the pre-processing stage of quantifying features from perturbation impacts incurs a computational cost. For each token in an answer, this cost includes an initial Jacobian calculation and subsequently, for each of the $S$ perturbations, processing the perturbed hidden state through the LLM's head to obtain perturbed logits. Access to model internals is also a prerequisite for this feature extraction phase. Secondly, feature effectiveness depends on perturbation hyperparameters (e.g., $\epsilon_{\text{max}}$, $S$), which, though optimized in our experiments, may need retuning for different models or tasks. Lastly, the quality of extracted features inherently relies on the meaningfulness of the base LLM's internal representations; if an LLM's hidden states do not systematically encode information related to its certainty, the efficacy of any method probing these states might be constrained.

These limitations also highlight opportunities for improvement. One avenue is using the learned stability signals not just for post-hoc estimation but to directly inform and calibrate the generation process, potentially reducing hallucinations. Additionally, while this work perturbs only the final hidden state, exploring perturbations across different transformer layers may yield richer or complementary indicators of confidence.

\section*{Ethical Considerations}
While CCPS is developed with the aim of enhancing the reliability and trustworthiness of LLMs, several ethical considerations are relevant to its application and interpretation. A primary concern is the potential for over-reliance on the confidence scores produced. Although CCPS demonstrates improved calibration and discrimination, it is crucial to recognize that no confidence estimation method is perfect. In high-stakes domains, such as medicine, finance, or law, an uncritical acceptance of automated confidence scores without appropriate human judgment and oversight could lead to adverse outcomes if the underlying LLM makes an error that is not perfectly flagged by the confidence score.

Secondly, the fairness of CCPS across diverse demographic groups and data distributions warrants careful attention during deployment. If the base LLMs, from which internal representations are extracted, contain inherent biases or exhibit differential performance characteristics for certain populations, CCPS's confidence assessments could potentially reflect or even inadvertently amplify these disparities. This could result in confidence scores that are less reliable for some groups than for others, potentially leading to inequitable or unfair consequences. Therefore, the deployment of any confidence estimation method, including CCPS, especially in sensitive applications, should be accompanied by rigorous testing for fairness, ongoing monitoring of its performance across relevant subgroups, and a clear framework emphasizing its role as an assistive tool to augment, not replace, human expertise and critical decision-making.

\section*{Acknowledgments}
This work was supported in part by the Henry Ford Health + Michigan State University Health Sciences Cancer Seed Funding Program and by the JPMorgan Chase AI Research Faculty Research Award. The authors are solely responsible for the contents of this paper; the opinions expressed do not necessarily reflect those of the funding organizations. The authors also acknowledge the use of Large Language Models to assist in polishing the language and grammar of this manuscript.

\section*{Disclaimer}
This paper was prepared for informational purposes by the Artificial Intelligence Research group of JPMorgan Chase \& Co and its affiliates (“JP Morgan”), and is not a product of the Research Department of JP Morgan. JP Morgan makes no representation and warranty whatsoever and disclaims all liability, for the completeness, accuracy or reliability of the information contained herein. This document is not intended as investment research or investment advice, or a recommendation, offer or solicitation for the purchase or sale of any security, financial instrument, financial product or service, or to be used in any way for evaluating the merits of participating in any transaction, and shall not constitute a solicitation under any jurisdiction or to any person, if such solicitation under such jurisdiction or to such person would be unlawful.

\bibliography{custom}

\clearpage
\appendix
\begin{center}
\huge\textbf{Appendix}
\end{center}
\label{sec:appendix}

\begin{center}
\Large\textbf{Table of Contents}
\rule{\linewidth}{1pt}
\end{center}
\startcontents[sections]
\printcontents[sections]{}{1}{}

\rule{\linewidth}{1pt}

\section{Artifact Usage and Creation}
\label{app:artifacts}
\noindent\textbf{Consistency with Intended Use of Existing Artifacts:} All existing scientific artifacts employed in this research, including pre-trained LLMs, benchmark datasets (\texttt{MMLU}, \texttt{MMLU-Pro}, and the constituent datasets of \texttt{CT-CHOICE}/\texttt{CT-OE}), and software libraries, were used in a manner consistent with their specified intended uses, primarily for academic research, evaluation, and the development of new methodologies within the field of Natural Language Processing. The use of proprietary models like \texttt{GPT-4o-mini} for data labeling was conducted in accordance with its API terms of service for research applications.

\vspace{5pt}
\noindent\textbf{Intended Use of Created Artifacts:} The scientific artifacts created as part of this work—including the source code for the CCPS method, our trained confidence estimation models, and the derived feature sets—are primarily intended to support academic research. Their release aims to ensure the reproducibility of our findings, encourage further investigation into LLM confidence estimation techniques, and allow the community to build upon our contributions. The use and distribution of any created artifacts that are derivative of existing datasets or models will be governed by terms compatible with the original access conditions and licenses of those foundational resources, particularly ensuring that derivatives of artifacts intended for research remain within research contexts where applicable.
\section{Feature Set Description}
\label{app:feature_set_description}
This appendix details the features extracted for analyzing the language model's token-level generative behavior. Our final model uses a 75-dimensional feature vector ($D_f=75$) per token, which is derived from 30 base metric definitions. These features are calculated from the model's internal states and its responses to systematic perturbations, where a token's hidden state is incrementally moved along an adversarial direction. The 75 features are composed of 15 single scalar values (12 Original State features and 3 Overall Perturbation features), and 60 features derived from statistical summaries (mean, standard deviation, minimum, and maximum) of the remaining 15 base metrics that are computed across the perturbation trajectory. This process yields a total of $15 + (15 \times 4) = 75$ features per token. The definitions for the base metrics are provided in Table~\ref{tab:all_features}.

\subsection{Original State Features} 
This feature set quantifies the model's baseline predictive characteristics for each token prior to experimental perturbation. These include measures of output probabilities, logits, distribution entropy, prediction margins, and vector norms of internal representations. These features establish a reference for evaluating perturbation effects.

\subsection{Overall Perturbation Features} 
This group comprises scalar Features quantifying specific properties related to the perturbation mechanism itself or its direct consequences. These include the L2 norm of the Jacobian vector, the perturbation magnitude required to alter the model's top-1 predicted token (\texttt{epsilon\_to\_flip\_token}), and the integrated effect of perturbations on the log-probability of the token guiding the perturbation direction (PEI value).

\subsection{Perturbed State Features} 
These features describe the model's output characteristics (e.g., token probabilities, distribution entropy, decision margins, as listed in Table~\ref{tab:all_features}) evaluated after its hidden states are perturbed. The base metrics are calculated at each discrete perturbation step. Statistical summaries (minimum, maximum, mean, standard deviation) of these per-step metrics are then computed across all applied perturbation magnitudes for a given token. This process summarizes the model's output behavior under varying degrees of targeted hidden state modification.

\subsection{Comparison Features (Original vs.\ Perturbed)} 
This feature set quantifies the differences between the model's original state (hidden states, logits, probability distributions) and its state after each perturbation step. Base comparison metrics are detailed in Table~\ref{tab:all_features}. These metrics, such as changes in log-probabilities, distributional divergences (KL, JS), and vector similarities/distances, are statistically summarized (minimum, maximum, mean, standard deviation) across all perturbation magnitudes. The summaries indicate the extent of change in model representations and outputs due to the applied perturbations.

A total of $D_f=75$ such features are extracted per token.

\begin{table*}[t]
\centering
\small
\resizebox{\textwidth}{!}{
\begin{tabular}{p{0.4\textwidth}p{0.55\textwidth}}
\toprule
\multicolumn{2}{c}{\textbf{Original State Features}} \\
\midrule
\texttt{original\_log\_prob\_actual} & Log-probability of the actual token based on the model’s original (unperturbed) output distribution, i.e.\ $\log P_{\text{original}}(\text{actual\_token})$. \\
\texttt{original\_prob\_actual} & Probability of the actual token based on the model’s original output distribution, i.e.\ $P_{\text{original}}(\text{actual\_token})$. \\
\texttt{original\_logit\_actual} & Logit value of the actual token from the model’s original output. \\
\texttt{original\_prob\_argmax} & Highest probability assigned to any token by the original model, i.e.\ $P_{\text{original}}(\text{argmax\_token})$. \\
\texttt{original\_logit\_argmax} & Highest logit value assigned to any token by the original model. \\
\texttt{original\_entropy} & Entropy of the original predictive distribution: $-\sum_i P_{\text{original}}(i)\log P_{\text{original}}(i)$. \\
\texttt{original\_margin\_logit\_top1\_top2} & Difference between top-1 and top-2 logits in the original output. \\
\texttt{original\_margin\_prob\_top1\_top2} & Difference between top-1 and top-2 probabilities in the original output. \\
\texttt{original\_norm\_logits\_L2} & L2 norm of the original logit vector. \\
\texttt{original\_std\_logits} & Standard deviation of the original logit values. \\
\texttt{original\_norm\_hidden\_state\_L2} & L2 norm of the original last hidden state vector. \\
\texttt{is\_actual\_token\_original\_argmax} & Indicator (1/0) if the actual token is the argmax under the original model. \\
\midrule
\multicolumn{2}{c}{\textbf{Overall Perturbation Features}} \\
\midrule
\texttt{jacobian\_norm\_token} & L2 norm of the Jacobian of the token’s log-prob w.r.t.\ the original hidden state (sensitivity measure). \\
\texttt{epsilon\_to\_flip\_token} & Minimum perturbation magnitude along the Jacobian direction to change the top-1 token. \\
\texttt{pei\_value\_token} & Perturbation Energy Integral (PEI): total normalized drop in log-prob of the actual token over all perturbation steps. \\
\midrule
\multicolumn{2}{c}{\textbf{Perturbed State Features}} \\
\midrule
\texttt{perturbed\_log\_prob\_actual} & Log-prob of the actual token after hidden-state perturbation, $\log P_{\text{perturbed}}(\text{actual\_token})$. \\
\texttt{perturbed\_prob\_actual} & Probability of the actual token after perturbation, $P_{\text{perturbed}}(\text{actual\_token})$. \\
\texttt{perturbed\_logit\_actual} & Logit value of the actual token after perturbation. \\
\texttt{perturbed\_prob\_argmax} & Highest probability assigned after perturbation. \\
\texttt{perturbed\_logit\_argmax} & Highest logit value assigned after perturbation. \\
\texttt{perturbed\_entropy} & Entropy of the perturbed predictive distribution. \\
\texttt{perturbed\_margin\_logit\_top1\_top2} & Difference between top-1 and top-2 logits post-perturbation. \\
\texttt{perturbed\_norm\_logits\_L2} & L2 norm of the perturbed logit vector. \\
\midrule
\multicolumn{2}{c}{\textbf{Comparison Features (Original vs.\ Perturbed)}} \\
\midrule
\texttt{delta\_log\_prob\_actual\_from\_original} & Change in log-prob: $\log P_{\text{original}} - \log P_{\text{perturbed}}$ for the actual token. \\
\texttt{did\_argmax\_change\_from\_original} & Indicator (1/0) if the argmax token changed after perturbation. \\
\texttt{kl\_div\_perturbed\_from\_original} & KL divergence $D_{KL}(P_{\text{original}}\parallel P_{\text{perturbed}})$. \\
\texttt{js\_div\_perturbed\_from\_original} & Jensen-Shannon divergence between original and perturbed distributions. \\
\texttt{cosine\_sim\_logits\_perturbed\_to\_original} & Cosine similarity of logit vectors before vs.\ after perturbation. \\
\texttt{cosine\_sim\_hidden\_perturbed\_to\_original} & Cosine similarity of hidden-state vectors before vs.\ after perturbation. \\
\texttt{l2\_dist\_hidden\_perturbed\_from\_original} & L2 distance between hidden-state vectors before vs.\ after perturbation. \\
\bottomrule
\end{tabular}
}
\caption{Definitions of features employed in this study, grouped by feature set type.}
\label{tab:all_features}
\end{table*}
\section{Datasets}
\label{app:datasets}

This section provides further details on the datasets used for training, validation, and evaluation of our confidence estimation models. All datasets employed in this study are in English. For comprehensive information regarding the original construction, specific domain coverage, linguistic characteristics, and any available demographic details for the underlying public benchmarks (such as MMLU, MMLU-Pro, and the constituent datasets of \texttt{CT-CHOICE} and \texttt{CT-OE}), we refer readers to their respective original publications, which are cited upon their introduction in the subsequent subsections.

\subsection{Training and Validation Datasets}
For training and validating our confidence estimation models, we utilize the \texttt{CT-CHOICE} and \texttt{CT-OE} datasets, designed for multiple-choice and open-ended QA formats, respectively. These datasets, generated following the methodology of \citet{kapoor-etal-2024-calibration}, aggregate a diverse collection of commonly used public QA datasets. Instances from these datasets were formatted to ensure a maximum input sequence length of 1,600 tokens during our training process. The underlying datasets include:
\begin{itemize}
    \item AI2 Reasoning Challenge (ARC) \cite{arc}
    \item Boolean Questions (BoolQ) \cite{boolq}
    \item CommonsenseQA \cite{commonsenseqa}
    \item CosmosQA \cite{cosmos}
    \item HellaSwag \cite{hellaswag}
    \item MathQA \cite{mathqa}
    \item Recognizing Textual Entailment (RTE/SNLI) \cite{snli}
    \item Adversarial NLI \cite{adversarialnli}
    \item OpenBookQA \cite{openbook}
    \item PIQA \cite{piqa}
    \item SciQ \cite{sciq}
    \item The CommitmentBank (CB) \cite{cb}
    \item Multi-Sentence Reading Comprehension (MultiRC) \cite{multirc}
    \item Choice of Plausible Alternatives (CoPA) \cite{copa}
    \item TREC \cite{trec}
    \item Adversarial Winograd (Winogrande) \cite{winogrande}
\end{itemize}

\subsection{Evaluation Datasets}
Our evaluation suite consists of variants of the Massive Multitask Language Understanding (MMLU) \cite{mmlu} and MMLU-Pro \cite{mmlupro} benchmarks, formatted for both multiple-choice and open-ended evaluation.

\noindent\textbf{MMLU-CHOICE and MMLU-OE:} These datasets are derived from the standard MMLU benchmark, which covers 57 diverse tasks spanning STEM, humanities, social sciences, and other areas. We created multiple-choice (\texttt{MMLU-CHOICE}) and open-ended (\texttt{MMLU-OE}) versions following the data processing approach of \citet{kapoor-etal-2024-calibration}. The constituent tasks and their respective sample sizes for MMLU are listed in Table \ref{tab:mmlu_tasks}.

\begin{table*}[t]
\centering
\small
\begin{tabular}{lr|lr}
\toprule
\textbf{Task Name} & \textbf{Size} & \textbf{Task Name} & \textbf{Size} \\
\midrule
Abstract Algebra & 100 & High School Statistics & 216 \\
Anatomy & 135 & High School Us History & 204 \\
Astronomy & 152 & High School World History & 237 \\
Business Ethics & 100 & Human Aging & 223 \\
Clinical Knowledge & 265 & Human Sexuality & 131 \\
College Biology & 144 & International Law & 121 \\
College Chemistry & 100 & Jurisprudence & 108 \\
College Computer Science & 100 & Logical Fallacies & 163 \\
College Mathematics & 100 & Machine Learning & 112 \\
College Medicine & 173 & Management & 103 \\
College Physics & 102 & Marketing & 234 \\
Computer Security & 100 & Medical Genetics & 100 \\
Conceptual Physics & 235 & Miscellaneous & 783 \\
Econometrics & 114 & Moral Disputes & 346 \\
Electrical Engineering & 145 & Moral Scenarios & 895 \\
Elementary Mathematics & 378 & Nutrition & 306 \\
Formal Logic & 126 & Philosophy & 311 \\
Global Facts & 100 & Prehistory & 324 \\
High School Biology & 310 & Professional Accounting & 282 \\
High School Chemistry & 203 & Professional Law & 1,534 \\
High School Computer Science & 100 & Professional Medicine & 272 \\
High School European History & 165 & Professional Psychology & 612 \\
High School Geography & 198 & Public Relations & 110 \\
High School Government And Politics & 193 & Security Studies & 245 \\
High School Macroeconomics & 390 & Sociology & 201 \\
High School Mathematics & 270 & US Foreign Policy & 100 \\
High School Microeconomics & 238 & Virology & 166 \\
High School Physics & 151 & World Religions & 171 \\
High School Psychology & 545 & & \\
\midrule
\multicolumn{3}{r}{\textbf{Total}} & \textbf{14,042} \\
\bottomrule
\end{tabular}
\caption{Tasks and sample sizes in the MMLU benchmark.}
\label{tab:mmlu_tasks}
\end{table*}

\noindent\textbf{MMLU-PRO-CHOICE:} This dataset is the multiple-choice version of MMLU-Pro \cite{mmlupro}, which includes 14 tasks designed with more challenging questions that often require deeper domain knowledge. Unlike the standard MMLU, the structure of MMLU-Pro questions often makes the provided choices an indispensable part of the question's context, meaning it could not be meaningfully converted to an open-ended format without fundamentally altering the nature of the problems. Furthermore, the answer options in MMLU-Pro frequently extend beyond the typical A-D choices, sometimes including E, F, or more. The tasks and their sample sizes for MMLU-Pro are detailed in Table \ref{tab:mmlu_pro_tasks}.

\begin{table*}[t!]
\centering
\small
\begin{tabular}{lr}
\toprule
\textbf{Task Name} & \textbf{Size} \\
\midrule
Biology & 717 \\
Business & 789 \\
Chemistry & 1,132 \\
Computer Science & 410 \\
Economics & 844 \\
Engineering & 969 \\
Health & 818 \\
History & 381 \\
Law & 1,101 \\
Math & 1,351 \\
Other & 924 \\
Philosophy & 499 \\
Physics & 1,299 \\
Psychology & 798 \\
\midrule
\textbf{Total} & \textbf{12,032} \\
\bottomrule
\end{tabular}
\caption{Tasks and sample sizes in the MMLU-Pro benchmark.}
\label{tab:mmlu_pro_tasks}
\end{table*}

\subsection{Response Generation and Labeling}
\label{app:response_generation_labeling}

For all datasets described above, responses from the base LLMs were first generated to create the instances for our confidence estimation task. The user prompt, which includes the question and any contextual information (such as few-shot exemplars), was constructed following the methodology of \citet{kapoor-etal-2024-calibration}, to which we refer the reader for further details. We employed specific system prompts for guiding the base LLMs during response generation, as detailed in Table \ref{tab:generation_prompts}. These prompts are similar to those used by \citet{kapoor-etal-2024-calibration} but were slightly refined for improved clarity to the LLMs. In line with their approach, for multiple-choice QA datasets (\texttt{CT-CHOICE}, \texttt{MMLU-CHOICE}, \texttt{MMLU-PRO-CHOICE}), answers were generated with a maximum token limit of 1, corresponding to the chosen option letter. For open-ended datasets (\texttt{CT-OE}, \texttt{MMLU-OE}), responses were generated using greedy decoding with a maximum length of 30 tokens.

\begin{table*}[t!]
\centering
\small
\begin{tabular}{p{0.15\textwidth}|p{0.75\textwidth}}
\toprule
\textbf{Format} & \textbf{System Prompt} \\
\midrule
Multiple-Choice & \texttt{You are an expert who responds with concise, correct answers. For multiple-choice questions, respond only with the letter of the correct option (e.g., a, b, c, d, ...). Do not include any explanation or additional text.} \\
\midrule
Open-Ended & \texttt{You are an expert who responds with concise, correct answers. Directly state the answer without phrases like 'the correct answer is'.} \\
\bottomrule
\end{tabular}
\caption{System prompts used for base LLM response generation.}
\label{tab:generation_prompts}
\end{table*}

Each generated response was subsequently labeled as correct or incorrect. For multiple-choice questions, correctness was determined by a straightforward string match between the LLM’s generated option letter and the ground truth option. For open-ended responses, assessing semantic equivalence requires a more nuanced approach. To this end, and consistent with recent literature, we employed a powerful auxiliary LLM not as a knowledge oracle, but as a semantic equivalence assessor. The grader model's task was constrained: for each question, it was provided with the ground-truth answer available in the dataset and the LLM's generated response, and was prompted to determine if the two answers were semantically equivalent. The reliability of using a powerful LLM for this specific equivalence task has been validated by \citet{kapoor-etal-2024-calibration}, who conducted a comparative analysis against human evaluations. Their study found that \texttt{GPT-4} assessments exhibited a low average absolute difference of 4.5\% in accuracy estimation compared to human annotators. Building upon their findings, and given the availability of even more capable models since their study, we utilized the more recent \texttt{GPT-4o-mini} model to ensure the highest quality semantic equivalence judgments. The prompts used for this grading task are detailed in Table \ref{tab:grading_prompts}.

\begin{table*}[t!]
\centering
\small
\begin{tabular}{p{0.15\textwidth}|p{0.75\textwidth}}
\toprule
\textbf{Prompt Type} & \textbf{Content} \\
\midrule
System Prompt & 
\begin{minipage}[t]{\linewidth}
\texttt{You are an automated grading assistant helping a teacher grade student answers.}
\end{minipage} \\
\midrule
User Prompt & 
\begin{minipage}[t]{\linewidth}
\texttt{The problem is: "\{question\}"}\\
\texttt{ }\\
\texttt{The correct answer for this problem is: "\{gt\_answer\}"}\\
\texttt{ }\\
\texttt{A student submitted the answer: "\{llm\_answer\}"}\\
\texttt{ }\\
\texttt{The student's answer should be semantically equivalent to the correct answer---that is, it should express the same meaning, even if the wording or format is slightly different. However, answers that are ambiguous, incorrect, or include conflicting or multiple answers should not be considered equivalent. Do not penalize superficial differences (e.g., spelling, synonyms, or phrasing), but ensure the core meaning is preserved.}\\
\texttt{ }\\
\texttt{Did the student provide a semantically equivalent answer to the ground truth? Please answer yes or no without any explanation:}
\end{minipage} \\
\bottomrule
\end{tabular}
\caption{Prompts used for GPT-4o-mini-based grading of open-ended responses.}
\label{tab:grading_prompts}
\end{table*}

The distribution of these correct and incorrect LLM responses across all datasets, for each base model used in our experiments, is detailed in Table \ref{tab:llm_performance}.

\begin{table*}[t]
\centering
\small
\resizebox{\textwidth}{!}{%
\begin{tabular}{l|ccc|ccc}
\multicolumn{7}{c}{\textbf{CT-CHOICE}} \\
\midrule
\multirow{2}{*}{\textbf{Model}} & \multicolumn{3}{c|}{\textbf{Train}} & \multicolumn{3}{c}{\textbf{Validation}} \\
 & \textbf{Correct} & \textbf{Incorrect} & \textbf{Total} & \textbf{Correct} & \textbf{Incorrect} & \textbf{Total} \\
\midrule
\texttt{Meta-Llama-3.1-8B-Instruct} & 12,654 (67.8\%) & 5,996 (32.1\%) & 18,650 & 1,688 (84.4\%) & 312 (15.6\%) & 2,000 \\
\texttt{Qwen2.5-14B-Instruct} & 15,116 (81.0\%) & 3,534 (18.9\%) & 18,650 & 1,796 (89.8\%) & 204 (10.2\%) & 2,000 \\
\texttt{Mistral-Small-24B-Instruct-2501} & 15,255 (81.8\%) & 3,395 (18.2\%) & 18,650 & 1,787 (89.3\%) & 213 (10.7\%) & 2,000 \\
\texttt{Qwen2.5-32B-Instruct} & 15,724 (84.3\%) & 2,926 (15.7\%) & 18,650 & 1,828 (91.4\%) & 172 (8.6\%) & 2,000 \\
\bottomrule
\end{tabular}
}

\vspace{1em}
\resizebox{\textwidth}{!}{%
\begin{tabular}{l|ccc|ccc}
\multicolumn{7}{c}{\textbf{CT-OE}} \\
\midrule
\multirow{2}{*}{\textbf{Model}} & \multicolumn{3}{c|}{\textbf{Train}} & \multicolumn{3}{c}{\textbf{Validation}} \\
 & \textbf{Correct} & \textbf{Incorrect} & \textbf{Total} & \textbf{Correct} & \textbf{Incorrect} & \textbf{Total} \\
\midrule
\texttt{Meta-Llama-3.1-8B-Instruct} & 9,165 (49.5\%) & 9,369 (50.5\%) & 18,534 & 1,014 (50.7\%) & 986 (49.3\%) & 2,000 \\
\texttt{Qwen2.5-14B-Instruct} & 11,656 (62.9\%) & 6,878 (37.1\%) & 18,534 & 1,221 (61.0\%) & 779 (39.0\%) & 2,000 \\
\texttt{Mistral-Small-24B-Instruct-2501} & 10,532 (56.8\%) & 8,002 (43.2\%) & 18,534 & 1,145 (57.2\%) & 855 (42.8\%) & 2,000 \\
\texttt{Qwen2.5-32B-Instruct} & 12,083 (65.2\%) & 6,451 (34.8\%) & 18,534 & 1,201 (60.0\%) & 799 (40.0\%) & 2,000 \\
\bottomrule
\end{tabular}
}

\vspace{1em}
\begin{tabular}{l|ccc}
\multicolumn{4}{c}{\textbf{MMLU-CHOICE}} \\
\midrule
\multirow{2}{*}{Model} & \multicolumn{3}{c}{Test} \\
 & Correct & Incorrect & Total \\
\midrule
\texttt{Meta-Llama-3.1-8B-Instruct} & 9,041 (64.4\%) & 5,001 (35.6\%) & 14,042 \\
\texttt{Qwen2.5-14B-Instruct} & 10,898 (77.6\%) & 3,144 (22.4\%) & 14,042 \\
\texttt{Mistral-Small-24B-Instruct-2501} & 11,231 (80.0\%) & 2,811 (20.0\%) & 14,042 \\
\texttt{Qwen2.5-32B-Instruct} & 11,488 (81.8\%) & 2,554 (18.2\%) & 14,042 \\
\bottomrule
\end{tabular}

\vspace{1em}
\begin{tabular}{l|ccc}
\multicolumn{4}{c}{\textbf{MMLU-PRO-CHOICE}} \\
\midrule
\multirow{2}{*}{Model} & \multicolumn{3}{c}{Test} \\
 & Correct & Incorrect & Total \\
\midrule
\texttt{Meta-Llama-3.1-8B-Instruct} & 4,135 (34.4\%) & 7,897 (65.6\%) & 12,032 \\
\texttt{Qwen2.5-14B-Instruct} & 6,187 (51.4\%) & 5,845 (48.6\%) & 12,032 \\
\texttt{Mistral-Small-24B-Instruct-2501} & 6,523 (54.2\%) & 5,509 (45.8\%) & 12,032 \\
\texttt{Qwen2.5-32B-Instruct} & 6,870 (57.1\%) & 5,162 (42.9\%) & 12,032 \\
\bottomrule
\end{tabular}

\vspace{1em}
\begin{tabular}{l|ccc}
\multicolumn{4}{c}{\textbf{MMLU-OE}} \\
\midrule
\multirow{2}{*}{Model} & \multicolumn{3}{c}{Test} \\
 & Correct & Incorrect & Total \\
\midrule
\texttt{Meta-Llama-3.1-8B-Instruct} & 4,225 (30.1\%) & 9,817 (69.9\%) & 14,042 \\
\texttt{Qwen2.5-14B-Instruct} & 6,386 (45.5\%) & 7,656 (54.5\%) & 14,042 \\
\texttt{Mistral-Small-24B-Instruct-2501} & 6,338 (45.1\%) & 7,704 (54.9\%) & 14,042 \\
\texttt{Qwen2.5-32B-Instruct} & 6,814 (48.5\%) & 7,228 (51.5\%) & 14,042 \\
\bottomrule
\end{tabular}
\caption{Distribution of correct and incorrect responses across CT-CHOICE, CT-OE, and MMLU variants.}
\label{tab:llm_performance}
\end{table*}
\section{Baseline Method Details}
\label{app:baselines_desc}
This section details the baseline methods implemented for comparison against our proposed CCPS method. Our selection of baselines was guided by the aim to provide a comprehensive benchmark against prominent, recent, and state-of-the-art techniques in LLM confidence estimation, many of which are established through peer-reviewed publications in highly regarded scientific venues. While the work introducing CT by \citet{kapoor-etal-2024-calibration} provided a valuable starting point by evaluating methods such as P(True), Instruction Tuning (IT), Logit Temperature Scaling (LTS), and a specific variant of SAPLMA (SAPLMA-F), our study expands significantly on this comparison. We include P(IK), which was not part of their direct comparison, and additional SAPLMA variants (SAPLMA-M, SAPLMA-UM) to explore signals from different representational depths. Furthermore, our evaluation framework encompasses a broader range of test conditions, including comprehensive training and testing on both multiple-choice and open-ended formats, and performance on datasets like \texttt{MMLU-PRO-CHOICE}, aspects not exhaustively covered for all these prior methods in the context of confidence estimation by \citet{kapoor-etal-2024-calibration}. We also incorporate LitCab \cite{liu2024litcablightweightlanguagemodel}, another significant and well-regarded recent contribution in lightweight white-box confidence estimation also originating from a top-tier conference, which provides an important additional point of comparison. For all established baseline methods, we adhered to the architectural descriptions and training configurations reported in their original publications. Common training hyperparameters, such as total steps and optimizer settings, are described in Section \ref{sec:experiments} (Training Details).

\subsection{P(True)} Introduced by \citet{kadavath2022languagemodelsmostlyknow}, P(True) assesses an LLM's self-evaluation of a generated answer. After an LLM generates an answer to an input prompt $P$, it is presented with the question, "Is the proposed answer correct? a) no b) yes" (referred to as the uncertainty query). The probabilities assigned by the original, frozen LLM to options 'a' and 'b' are then normalized (e.g., via softmax) to derive the confidence score, representing the probability of correctness. This method requires no additional training.

\subsection{P(IK)} Also from \citet{kadavath2022languagemodelsmostlyknow}, P(IK) (short for "I Know") estimates the LLM's probability of correctly answering a given question \textit{before} it generates a specific response. This typically involves training a lightweight classifier head on a hidden state representation from the LLM (e.g., the final hidden state after processing the input prompt $P$) to predict correctness. The output probabilities from this classifier serve as the confidence score.

\subsection{Logit Temperature Scaling (LTS)} As described by \citet{jiang-etal-2021-know}, LTS is a post-hoc calibration technique that adjusts a model's output probabilities. It introduces a scalar temperature parameter $\tau > 0$ which is applied to the logits before the LLM's final softmax function. In our application, after the LLM responds to the uncertainty query, the temperature $\tau$ is applied to the logits corresponding to the 'a' and 'b' options. The calibrated probability is then $\text{softmax}(\text{logits}_{\text{uncertainty query}}/\tau)$. The temperature $\tau$ is optimized on a held-out development set. LTS is computationally very light as it involves learning only a single parameter.

\subsection{Instruction Tuning (IT)} Instruction tuning, as introduced by \cite{flan1}, involves fine-tuning language models on a collection of tasks framed as natural language instructions. In our setting, this baseline involves fine-tuning the base LLM to respond to the uncertainty query more accurately. The model is trained using Low-Rank Adaptation (LoRA) \cite{lora}, a parameter-efficient fine-tuning technique, to predict the correct option ('a' or 'b') for the uncertainty query, based on ground-truth labels derived from the answer grading phase. While LoRA makes this more efficient than full fine-tuning of all parameters, it remains more resource-intensive than non-fine-tuning methods.

\subsection{SAPLMA} SAPLMA (Statement Accuracy Prediction based on Language Model Activations) \cite{azaria-mitchell-2023-internal} trains a lightweight feedforward classifier on LLM hidden state activations to predict statement truthfulness, while the LLM itself remains frozen. SAPLMA’s classifier employs a feedforward neural network featuring three hidden layers with decreasing numbers of hidden units (256, 128, 64), each followed by a ReLU activation. Their studies suggest that signals related to an LLM's internal assessment of truthfulness or confidence can manifest at different network depths depending on the model architecture and task. Therefore, while a common approach is to use final hidden states (\textbf{SAPLMA-F}), we also implemented variants using activations from the middle layer (\textbf{SAPLMA-M}) and an upper-middle layer (\textbf{SAPLMA-UM}) of the LLM to explore these potentially richer representational layers. The output probabilities from these classifiers are used as confidence scores.

\subsection{Calibration-Tuning (CT)} Proposed by \citet{kapoor-etal-2024-calibration}, CT fine-tunes an LLM (using LoRA) to explicitly predict its answer's correctness in response to the uncertainty query. It uses a classification loss combined with a divergence-based regularizer (such as Jensen-Shannon or KL Divergence) to help maintain the LLM's original generation capabilities. While LoRA reduces the training burden compared to full fine-tuning, CT can still be resource-intensive, reportedly taking about 4 GPU days on an NVIDIA V100 for their experiments. The divergence term, particularly with longer sequences in open-ended tasks, can also be memory-demanding.

\subsection{LitCab} This lightweight calibration method by \citet{liu2024litcablightweightlanguagemodel} involves a trainable linear layer of size (hidden\_dim $\times$ vocabulary\_size) that is attached to the LLM's final hidden states. This layer predicts a bias term which is added to the original output logits of the LLM. LitCab is trained using a contrastive max-margin loss, which typically requires multiple incorrect answer examples per question. The confidence score is then derived from the geometric mean of the adjusted probabilities of the response tokens.
\section{CCPS Architecture Details}
\label{app:ccps_architecture}
Our CCPS approach employs a feature projection network ($E_{\text{MC}}$ for multiple-choice, $E_{\text{OE}}$ for open-ended) followed by a classifier head ($C$). The specific architectures for these components were determined through a systematic hyperparameter search for both MC and OE formats, aimed at optimizing for the loss on validation data. Key training hyperparameters such as learning rate ($1 \times 10^{-4}$), weight decay (0.1), batch size (32), and training steps were kept consistent during this search, aligned with those detailed in Section \ref{sec:experiments} (Training Details). The finalized best-performing architectures are detailed below.

\subsection{Multiple-Choice Question Answering}
For the Multiple-Choice (MC) CCPS model, the hyperparameter search explored various configurations for the contrastive encoder ($E_{\text{MC}}$), including different embedding dimensions, the number and size of hidden layers, and a range of activation functions (ReLU, GeLU, SiLU, ELU, Leaky ReLU). Similarly, various hidden layer structures and activation functions were evaluated for the MLP-based classifier head ($C$). The selected architecture, which yielded the optimal balance of performance metrics, is as follows: the contrastive encoder ($E_{\text{MC}}$) is an MLP that processes the $D_f$-dimensional feature vector. It consists of a sequence of linear layers with output dimensions 64, 32, 16, and a final 8-dimensional embedding layer. ELU activation is applied after each layer except the output embedding layer. The subsequent classifier head receives the 8-dimensional embedding and passes it through an MLP with layers having output dimensions 48, 24, 12, each followed by ELU activation, and concludes with a final linear layer producing 2 output logits for classification.

\subsection{Open-Ended Question Answering}
For the Open-Ended (OE) CCPS model, the hyperparameter search for the contrastive encoder ($E_{\text{OE}}$) covered different embedding dimensions, the number and size of hidden channels for its 1D convolutional layers, various kernel sizes for these convolutional layers, and a range of activation functions (ReLU, GeLU, SiLU, ELU, Leaky ReLU). The MLP-based classifier head ($C$) also underwent a search over its hidden layer structures and activation functions. The best-performing configuration found is detailed here: the contrastive encoder ($E_{\text{OE}}$) processes sequences of $D_f$-dimensional token features. It employs two 1D convolutional layers; the first maps the input features to 64 channels (kernel size 3), and the second maps from 64 to 32 channels (kernel size 3). ReLU activation is applied after each convolutional layer. An adaptive max-pooling layer then reduces the sequence to a fixed-size representation, which is projected by a linear layer to a 16-dimensional embedding. The classifier head takes this 16-dimensional embedding, passes it through a linear layer to a 32-dimensional representation with ReLU activation, and finally to an output linear layer producing 2 logits for classification.
\section{Computational Setup and Resources}
\label{app:computational_resources}

All computational experiments were conducted on a GPU cluster equipped with NVIDIA A100-SXM (48GB VRAM) and NVIDIA H200 (141GB VRAM) GPUs. The allocation of GPU resources and specific setup details for the different confidence estimation methods are outlined below.

\subsection{P(True):} This method involves no training. Inference to obtain responses to the uncertainty query was performed using a single NVIDIA A100 GPU for the \texttt{Meta\allowbreak -Llama\allowbreak -3.1\allowbreak -8B\allowbreak -Instruct} and \texttt{Qwen\allowbreak 2.5\allowbreak -14B\allowbreak -Instruct} models, and a single NVIDIA H200 GPU for the \texttt{Mistral\allowbreak -Small\allowbreak -24B\allowbreak -Instruct\allowbreak -2501} and \texttt{Qwen\allowbreak 2.5\allowbreak -32B\allowbreak -Instruct} models.

\subsection{P(IK), SAPLMA, LitCab, and CCPS:}
\noindent\textbf{Hidden State / Feature Extraction:} For these methods, the initial stage of extracting hidden states or features (including perturbation processes for CCPS) from the base LLMs was performed using a single NVIDIA A100 GPU for the \texttt{Meta\allowbreak -Llama\allowbreak -3.1\allowbreak -8B\allowbreak -Instruct} and \texttt{Qwen\allowbreak 2.5\allowbreak -14B\allowbreak -Instruct} models. Due to their larger size, a single NVIDIA H200 GPU was used for the \texttt{Mistral\allowbreak -Small\allowbreak -24B\allowbreak -Instruct\allowbreak -2501} and \texttt{Qwen\allowbreak 2.5\allowbreak -32B\allowbreak -Instruct} models. This allocation ensured that each base LLM could be loaded onto an appropriate GPU.

\noindent\textbf{Training of Confidence Modules:} The subsequent training of the lightweight confidence modules for P(IK), SAPLMA variants, LitCab, and our CCPS classifiers (which typically comprise fewer than 1 million trainable parameters) was conducted on a single NVIDIA A100 GPU for all base LLMs.

\subsection{IT and LTS:} For IT, the LoRA-based fine-tuning of the base LLMs on the uncertainty query, and for LTS, the optimization of the temperature parameter, were performed on a single NVIDIA A100 GPU for \texttt{Meta\allowbreak -Llama\allowbreak -3.1\allowbreak -8B\allowbreak -Instruct} and \texttt{Qwen\allowbreak 2.5\allowbreak -14B\allowbreak -Instruct}. For the larger \texttt{Mistral\allowbreak -Small\allowbreak -24B\allowbreak -Instruct\allowbreak -2501} and \texttt{Qwen\allowbreak 2.5\allowbreak -32B\allowbreak -Instruct} models, these processes utilized a single NVIDIA H200 GPU.

\subsection{CT} The LoRA-based fine-tuning process for CT was conducted using 4 NVIDIA A100 GPUs operating in parallel for each combination of base LLM and dataset. This multi-GPU setup, managed with libraries such as Hugging Face Accelerate \cite{accelerate} and DeepSpeed \cite{deepspeed} (Zero Redundancy Optimizer Stage 2), was implemented in accordance with the original CT methodology to handle its more intensive training requirements.

\section{Analysis of Additional Trainable Parameters}
\label{app:appendix_params}

This appendix quantifies and compares the \emph{additional} learnable parameters introduced by each evaluated confidence estimation method, including our proposed CCPS, when applied to a base LLM. We first detail the architectural parameters of the base LLMs used, then provide the formulas for calculating additional trainable parameters for each confidence estimation method, followed by the exact parameter counts for the specific LLMs analyzed in our experiments. This analysis supports our claim regarding the lightweight nature of CCPS. All parameter counts include biases unless otherwise specified for asymptotic estimates.

\subsection{Base LLM Architectural Parameters}
The key architectural dimensions of the base Large Language Models (LLMs) utilized in this study, which influence the number of trainable parameters for certain confidence estimation methods, are provided in Table \ref{tab:app_model_dims}. These include the hidden size ($d_h$), tokenizer vocabulary size ($V$), and the number of decoder layers ($L$).

\begin{table*}[t]
  \centering
  \small
  \caption{Architectural dimensions for the base LLMs used.}
  \label{tab:app_model_dims}
  \begin{tabular}{lrrr}
    \toprule
    \textbf{Base LLM} & \textbf{$d_{h}$} & \textbf{$V$} & \textbf{$L$} \\
    \midrule
    \texttt{Meta\allowbreak -Llama\allowbreak -3.1\allowbreak -8B\allowbreak -Instruct}   & 4,096 & 128,256 & 32 \\
    \texttt{Qwen\allowbreak 2.5\allowbreak -14B\allowbreak -Instruct}    & 5,120 & 152,064 & 48 \\
    \texttt{Mistral\allowbreak -Small\allowbreak -24B\allowbreak -Instruct\allowbreak -2501} & 5,120 & 131,072 & 40 \\
    \texttt{Qwen\allowbreak 2.5\allowbreak -32B\allowbreak -Instruct}    & 5,120 & 152,064 & 64 \\
    \bottomrule
  \end{tabular}
\end{table*}

\subsection{Formulation of Additional Trainable Parameters}
The number of additional trainable parameters for each confidence estimation method is determined as follows (Table \ref{tab:app_param_formulas}). We define $D_f = 75$ as the input feature dimension for CCPS, and $r=8$ as the rank for LoRA implementations.

\begin{table*}[t]
  \centering
  \small
  \caption{Formulas for additional trainable parameters introduced by each method.}
  \label{tab:app_param_formulas}
  \begin{tabular}{p{2.7cm}p{5.3cm}p{5.3cm}}
    \toprule
    \textbf{Method} & \textbf{Trainable Component(s)} & \textbf{Formula for Parameters (incl. Biases)} \\
    \midrule
    P(True)         & None (prompting only)       & 0 \\
    LTS             & Temperature scalar $\tau$     & 1 \\
    P(IK)           & Linear layer ($d_h \to 2$)    & $2d_h + 2$ \\
    SAPLMA          & MLP ($d_h \to 256 \to 128 \to 64 \to 2$) & $256d_h + (256 \!\times\! 128 + 128) + (128 \!\times\! 64 + 64) + (64 \!\times\! 2 + 2)$ \\
                    &                                 & $= 256d_h + 41,282$ \\
    IT \& CT (LoRA) & LoRA layers (adapting Q \& V matrices in all $L$ layers, rank $r$) & $2L \cdot (d_h r + r d_h) = 4Ld_hr$ \\
    LitCab          & Linear bias layer ($d_h \to V$) & $d_hV + V$ \\
    CCPS (MC)       & Encoder $E_{\text{MC}}$ + Head $C_{\text{MC}}$ (MLPs) & $\sum (h_{i}h_{i+1} + h_{i+1}) + \sum (g_{j}g_{j+1} + g_{j+1})$ \\
                    & $E_{\text{MC}}$ widths: $(D_f,64,32,16,8)$ & \\
                    & $C_{\text{MC}}$ widths: $(8,48,24,12,2)$ & \\
    CCPS (OE)       & Encoder $E_{\text{OE}}$ + Head $C_{\text{OE}}$ & (See text for detailed breakdown) \\
  \bottomrule
  \end{tabular}
\end{table*}

For CCPS (MC), the encoder $E_{\text{MC}}$ layers are $(D_f,64)$, $(64,32)$, $(32,16)$, $(16,8)$, and classifier $C_{\text{MC}}$ layers are $(8,48)$, $(48,24)$, $(24,12)$, $(12,2)$. The sum of $h_i h_{i+1}$ (weights) and $h_{i+1}$ (biases) for $E_{\text{MC}}$, and $g_j g_{j+1}$ (weights) and $g_{j+1}$ (biases) for $C_{\text{MC}}$ yields the total.
For CCPS (OE), the encoder $E_{\text{OE}}$ consists of two 1D convolutional layers (first: $D_f$ to 64 channels, kernel 3; second: 64 to 32 channels, kernel 3) and a linear projection layer (32 to 16 dimensions). The classifier head $C_{\text{OE}}$ is an MLP (16 to 32 dimensions, then 32 to 2 outputs). The exact calculation for CCPS (OE), including convolutional layer parameters (weights and biases) and MLP parameters, results in approximately 22,000 parameters, as detailed in Appendix \ref{app:ccps_architecture}.

\subsection{Exact Additional Trainable Parameter Counts}
Based on the formulations above and the LLM dimensions in Table \ref{tab:app_model_dims}, the exact number of additional trainable parameters introduced by each method when applied to the different base LLMs is presented in Table \ref{tab:app_exact_param_counts}. For methods like IT and CT, LoRA with rank $r=8$ is applied to the Query (Q) and Value (V) matrices within each of the $L$ attention blocks of the base LLMs.

\begin{table*}[t]
  \centering
  \small
  \caption{Additional trainable parameters introduced by each confidence estimation method per base LLM (CCPS values for MC variant; LoRA rank $r=8$ adapting Q and V matrices in all $L$ layers).}
  \label{tab:app_exact_param_counts}
  \begin{tabular}{lrrrrrrr} 
    \toprule
    \textbf{Base LLM} & \textbf{P(True)} & \textbf{LTS} & \textbf{P(IK)} & \textbf{SAPLMA} & \textbf{IT/CT (LoRA-$r$)} & \textbf{LitCab} & \textbf{CCPS (MC)} \\
    \midrule
    \texttt{Meta\allowbreak -Llama\allowbreak -3.1\allowbreak -8B\allowbreak -Instruct}   & 0 & 1 & 8,194   & 1,089,858 & 4,194,304  & 525,464,832 & \textbf{9,542} \\
    \texttt{Qwen\allowbreak 2.5\allowbreak -14B\allowbreak -Instruct}    & 0 & 1 & 10,242  & 1,352,002 & 7,864,320  & 778,719,744 & \textbf{9,542} \\
    \texttt{Mistral\allowbreak -Small\allowbreak -24B\allowbreak -Instruct} & 0 & 1 & 10,242  & 1,352,002 & 6,553,600  & 671,219,712 & \textbf{9,542} \\
    \texttt{Qwen\allowbreak 2.5\allowbreak -32B\allowbreak -Instruct}    & 0 & 1 & 10,242  & 1,352,002 & 10,485,760 & 778,719,744 & \textbf{9,542} \\
    \bottomrule
  \end{tabular}
\end{table*}

\subsection{Discussion of Parameter Efficiency}
The results presented in Table \ref{tab:app_exact_param_counts} highlight the parameter efficiency of CCPS. Irrespective of the base LLM's size, our CCPS (MC) method introduces only 9,542 trainable parameters, and the CCPS (OE) variant introduces approximately 22,000 parameters. This contrasts sharply with other methods. For instance, LitCab requires hundreds of millions of parameters (e.g., over 525 million for \texttt{Meta\allowbreak -Llama\allowbreak -3.1\allowbreak -8B\allowbreak -Instruct}) due to its vocabulary-sized projection. LoRA-based fine-tuning (IT/CT with $r=8$) adds several million parameters (e.g., 4.2 million to 10.5 million). SAPLMA, with its MLP architecture, introduces a moderate number of parameters (e.g., approximately 1.1 million for \texttt{Meta\allowbreak -Llama\allowbreak -3.1\allowbreak -8B\allowbreak -Instruct}), while simpler probes like P(IK) remain very light (e.g., 8,194 for the same LLM). CCPS remains significantly more parameter-efficient than SAPLMA, LoRA-based methods, and LitCab.

To further illustrate this, Table \ref{tab:app_ratios_vs_ccps} shows the relative parameter budgets compared to CCPS (MC). CCPS (MC) is approximately 440 to 1,100 times smaller than LoRA-based IT/CT, and 55,000 to 81,000 times smaller than LitCab for the LLMs tested. This extreme parameter efficiency, combined with CCPS’s strong performance demonstrated in the main paper, underscores its suitability as a highly scalable solution for confidence estimation on large, frozen LLMs.

\begin{table*}[t]
\centering
\small
\caption{Relative trainable parameter budgets with respect to CCPS (MC variant; $\downarrow$ indicates better/fewer parameters).}
\label{tab:app_ratios_vs_ccps}
\begin{tabular}{lcc}
\toprule
\textbf{Base LLM} & \textbf{LitCab} $\div$ \textbf{CCPS} & \textbf{IT/CT LoRA-$r$} $\div$ \textbf{CCPS} \\
\midrule
\texttt{Meta\allowbreak -Llama\allowbreak -3.1\allowbreak -8B\allowbreak -Instruct} & $55,069 \times$ & $440 \times$ \\
\texttt{Qwen\allowbreak 2.5\allowbreak -14B\allowbreak -Instruct}    & $81,610 \times$ & $824 \times$ \\
\texttt{Mistral\allowbreak -Small\allowbreak -24B\allowbreak -Instruct\allowbreak -2501} & $70,344 \times$ & $687 \times$ \\
\texttt{Qwen\allowbreak 2.5\allowbreak -32B\allowbreak -Instruct}    & $81,610 \times$ & $1,099 \times$ \\
\bottomrule
\end{tabular}
\end{table*}
\section{Evaluation Metrics}
\label{app:evaluation_metrics}
We assess the performance of our confidence estimation method using a suite of standard metrics. This comprehensive set allows for a nuanced understanding beyond ECE and ACC, which can be less informative for imbalanced datasets often encountered in correctness prediction.

\subsection{Expected Calibration Error (ECE)} A model’s uncertainties are well-calibrated if they align with empirical probabilities—i.e., an event assigned probability $p$ occurs at rate $p$ in reality. Following \citet{kapoor-etal-2024-calibration}, we estimate ECE by binning the predicted confidence score (probability of correctness) for each of $n$ samples into $b$ equally-spaced bins $B = \{B_j\}_{j=1}^b$. The empirical ECE estimator is given by:
$$\text{ECE} = \sum_{j=1}^{b} \frac{|B_j|}{n} |\text{conf}(B_j) - \text{acc}(B_j)|$$
where $\text{conf}(B_j)$ is the average predicted confidence of samples in bin $B_j$ and $\text{acc}(B_j)$ is the corresponding ACC (fraction of correct LLM answers) within that bin. Consistent with common practice, we use $b=10$ bins. An ECE of 0 signifies perfect calibration.

\subsection{Brier Score} This measures the mean squared difference between the predicted probability of correctness $p_k$ for sample $k$ and its actual binary outcome $o_k$ (1 if correct, 0 if incorrect), summed over all $N$ samples:
$$ \text{Brier Score} = \frac{1}{N} \sum_{k=1}^{N} (p_k - o_k)^2 $$
It provides a measure of both calibration and refinement, with lower scores being better.

\subsection{Accuracy (ACC)} Refers to the proportion of the LLM's answers that are correct on the given task. While our method estimates confidence in these answers rather than altering them, ACC provides context for the difficulty of the underlying task.

\subsection{Area Under the Precision-Recall Curve (AUCPR)} This metric summarizes the trade-off between precision (the proportion of positively predicted instances that are truly positive, $\text{TP}/(\text{TP}+\text{FP})$) and recall (the proportion of actual positive instances that are correctly predicted, $\text{TP}/(\text{TP}+\text{FN})$) for the binary correctness classification task. The confidence score is used as the discrimination threshold, varied to plot the curve. AUCPR is particularly informative for imbalanced datasets where the number of incorrect answers might significantly outweigh correct ones, or vice-versa.

\subsection{Area Under the Receiver Operating Characteristic Curve (AUROC)} This evaluates the discriminative ability of the confidence score to distinguish between correct and incorrect answers. It plots the true positive rate (Recall) against the false positive rate ($\text{FP}/(\text{FP}+\text{TN})$) at various threshold settings of the confidence score. An AUROC of 1.0 indicates perfect discrimination, while 0.5 suggests random guessing.
\section{Evaluation on a High-Stakes Domain: MedMCQA}
\label{app:medmcqa_results}

To further assess the cross-domain robustness of CCPS and validate its performance in a critical, high-stakes setting as motivated in our introduction, we conducted an additional set of experiments on the \texttt{MedMCQA} benchmark~\cite{medmcqa}. This medical QA dataset provides a specialized domain to test the generalizability of our method. We compared CCPS against the strong Calibration Tuning (CT) baseline on all four base LLMs, using the identical experimental settings and fixed hyperparameters from our main evaluations to ensure a fair comparison. The performance on both multiple-choice and open-ended formats is presented in Table~\ref{tab:medmcqa_combined}.

\begin{table*}[t]
\centering
\small
\resizebox{\textwidth}{!}{
\begin{tabular}{lllccccc}
\toprule
\textbf{Dataset} & \textbf{Model} & \textbf{Method} & \textbf{ECE $\downarrow$} & \textbf{Brier $\downarrow$} & \textbf{ACC $\uparrow$} & \textbf{AUCPR $\uparrow$} & \textbf{AUROC $\uparrow$} \\
\midrule
\multirow{8}{*}{\textbf{MedMCQA (Multiple-Choice)}} 
& \multirow{2}{*}{Meta-Llama-3.1-8B} 
 & CT & 26.8 & 31.7 & 51.1 & 56.4 & 54.7 \\
& & CCPS & \textbf{11.0} & \textbf{20.9} & \textbf{67.7} & \textbf{74.0} & \textbf{73.9} \\
\cmidrule{2-8}
& \multirow{2}{*}{Qwen2.5-14B} 
 & CT & 28.0 & 30.1 & 62.2 & 68.2 & 55.3 \\
& & CCPS & \textbf{13.8} & \textbf{20.5} & \textbf{66.1} & \textbf{80.0} & \textbf{73.9} \\
\cmidrule{2-8}
& \multirow{2}{*}{Mistral-Small-24B} 
 & CT & 23.4 & 27.9 & 64.3 & 69.4 & 53.7 \\
& & CCPS & \textbf{21.2} & \textbf{25.4} & \textbf{65.2} & \textbf{70.8} & \textbf{54.5} \\
\cmidrule{2-8}
& \multirow{2}{*}{Qwen2.5-32B} 
 & CT & 41.1 & 41.8 & 44.0 & 72.4 & 52.4 \\
& & CCPS & \textbf{12.7} & \textbf{19.0} & \textbf{70.2} & \textbf{87.2} & \textbf{76.8} \\
\midrule
\multirow{8}{*}{\textbf{MedMCQA (Open-Ended)}} 
& \multirow{2}{*}{Meta-Llama-3.1-8B} 
 & CT & 12.1 & 18.7 & 71.8 & 38.7 & 66.4 \\
& & CCPS & \textbf{9.1} & \textbf{18.9} & \textbf{74.3} & \textbf{45.0} & \textbf{74.7} \\
\cmidrule{2-8}
& \multirow{2}{*}{Qwen2.5-14B} 
 & CT & 14.0 & 22.2 & 59.3 & 41.2 & 65.5 \\
& & CCPS & \textbf{11.5} & \textbf{21.2} & \textbf{68.5} & \textbf{44.5} & \textbf{68.6} \\
\cmidrule{2-8}
& \multirow{2}{*}{Mistral-Small-24B} 
 & CT & 14.6 & 29.7 & 63.4 & 36.6 & 53.9 \\
& & CCPS & \textbf{12.4} & \textbf{22.3} & \textbf{69.3} & \textbf{45.0} & \textbf{62.4} \\
\cmidrule{2-8}
& \multirow{2}{*}{Qwen2.5-32B} 
 & CT & 18.5 & 24.6 & 67.0 & 31.8 & 56.2 \\
& & CCPS & \textbf{18.3} & \textbf{21.9} & \textbf{72.4} & \textbf{43.5} & \textbf{70.1} \\
\bottomrule
\end{tabular}
}
\caption{Performance on MedMCQA across two setups: Multiple-Choice and Open-Ended. Arrows indicate whether lower ($\downarrow$) or higher ($\uparrow$) values are better. Best results per model are bolded.}
\label{tab:medmcqa_combined}
\end{table*}

The results on this specialized medical dataset show that CCPS's performance advantages are not only maintained but often amplified. In the multiple-choice setting, CCPS substantially improves upon CT across all metrics, reducing ECE by over 59\% for Llama-8B and nearly 69\% for Qwen-32B. In the open-ended setting, CCPS again demonstrates superior calibration and discrimination across nearly all models and metrics. These new results provide strong evidence that the advantages of CCPS generalize robustly beyond standard knowledge benchmarks to this critical, high-stakes domain.
\section{Extended Results and Analyses}
\label{app:extended_results}
This section provides supplementary results and analyses that further substantiate the findings presented in the main paper. We include comprehensive performance comparisons across all baseline methods, detailed per-LLM and per-task breakdowns, calibration curve visualizations, and feature importance analyses for our CCPS model.

\subsection{Per-Dataset Aggregate Performance Tables}
To offer a comprehensive comparison of all evaluated methods, including all baselines, Tables \ref{tab:appendix_mmluchoice}, \ref{tab:appendix_mmluprochoice}, and \ref{tab:appendix_mmluoe} present aggregate performance metrics for the \texttt{MMLU-CHOICE}, \texttt{MMLU-PRO-CHOICE}, and \texttt{MMLU-OE} datasets, respectively. Unlike the main paper's Table \ref{tab:performance_metrics} which shows mean scores across tasks for selected methods, these tables detail the mean $\pm$ standard deviation for all methods across all evaluated LLMs for each metric, providing insight into the consistency of performance.
\begin{table*}[t]
\centering
\small
\resizebox{\textwidth}{!}{
\begin{tabular}{llccccc}
\toprule
\textbf{Model} & \textbf{Method} & \textbf{ECE $\downarrow$} & \textbf{BRIER $\downarrow$} & \textbf{ACC $\uparrow$} & \textbf{AUCPR $\uparrow$} & \textbf{AUROC $\uparrow$} \\
\midrule
\multirow{10}{*}{\textbf{Meta-Llama-3.1-8B-Instruct}}
 & P(True) & 35.9\tiny{$\pm$5.7} & 39.4\tiny{$\pm$4.2} & 45.4\tiny{$\pm$5.9} & 66.0\tiny{$\pm$14.6} & 49.2\tiny{$\pm$5.7} \\
& P(IK) & 18.9\tiny{$\pm$9.6} & 25.4\tiny{$\pm$2.4} & 63.9\tiny{$\pm$14.8} & 65.3\tiny{$\pm$14.9} & 49.8\tiny{$\pm$1.8} \\
& LTS & 28.9\tiny{$\pm$6.6} & 34.5\tiny{$\pm$3.7} & 44.6\tiny{$\pm$6.9} & 66.6\tiny{$\pm$14.1} & 50.1\tiny{$\pm$4.3} \\
& IT & 33.4\tiny{$\pm$5.3} & 37.5\tiny{$\pm$3.7} & 47.2\tiny{$\pm$4.9} & 66.5\tiny{$\pm$14.7} & 49.8\tiny{$\pm$5.0} \\
& SAPLMA-M & 17.9\tiny{$\pm$9.7} & 24.8\tiny{$\pm$2.7} & 64.9\tiny{$\pm$15.1} & 64.9\tiny{$\pm$15.3} & 49.5\tiny{$\pm$3.1} \\
& SAPLMA-UM & 18.1\tiny{$\pm$9.8} & 24.9\tiny{$\pm$2.7} & 64.9\tiny{$\pm$15.1} & 64.6\tiny{$\pm$15.4} & 49.3\tiny{$\pm$3.3} \\
& SAPLMA-F & 18.2\tiny{$\pm$9.8} & 24.9\tiny{$\pm$2.6} & 64.9\tiny{$\pm$15.0} & 65.0\tiny{$\pm$15.0} & 49.6\tiny{$\pm$2.3} \\
& CT & 10.7\tiny{$\pm$6.7} & 21.1\tiny{$\pm$5.7} & 67.8\tiny{$\pm$12.2} & 74.2\tiny{$\pm$15.5} & 62.8\tiny{$\pm$8.0} \\
& LitCab & 10.9\tiny{$\pm$4.8} & 18.1\tiny{$\pm$5.5} & 73.2\tiny{$\pm$8.7} & 84.0\tiny{$\pm$13.5} & \textbf{77.1}\tiny{$\pm$8.2} \\
& CCPS & \textbf{6.5}\tiny{$\pm$3.9} & \textbf{17.1}\tiny{$\pm$4.7} & \textbf{73.4}\tiny{$\pm$8.5} & \textbf{84.1}\tiny{$\pm$13.5} & \textbf{77.1}\tiny{$\pm$8.5} \\
\midrule
\multirow{10}{*}{\textbf{Qwen2.5-14B-Instruct}}
 & P(True) & 47.0\tiny{$\pm$6.2} & 47.0\tiny{$\pm$4.8} & 41.3\tiny{$\pm$6.4} & 79.2\tiny{$\pm$12.5} & 51.2\tiny{$\pm$5.8} \\
& P(IK) & 25.1\tiny{$\pm$13.0} & 24.1\tiny{$\pm$3.1} & 76.8\tiny{$\pm$12.2} & 78.3\tiny{$\pm$12.1} & 49.9\tiny{$\pm$2.4} \\
& LTS & 41.5\tiny{$\pm$6.5} & 43.0\tiny{$\pm$4.3} & 38.6\tiny{$\pm$6.2} & 78.9\tiny{$\pm$12.6} & 49.7\tiny{$\pm$5.7} \\
& IT & 44.7\tiny{$\pm$5.9} & 44.0\tiny{$\pm$5.2} & 45.7\tiny{$\pm$7.1} & 79.4\tiny{$\pm$12.4} & 50.4\tiny{$\pm$6.7} \\
& SAPLMA-M & 23.8\tiny{$\pm$13.2} & 23.0\tiny{$\pm$4.0} & 78.1\tiny{$\pm$12.1} & 78.4\tiny{$\pm$12.3} & 50.5\tiny{$\pm$3.0} \\
& SAPLMA-UM & 23.7\tiny{$\pm$13.2} & 23.0\tiny{$\pm$4.0} & 78.2\tiny{$\pm$12.1} & 78.4\tiny{$\pm$12.1} & 50.3\tiny{$\pm$2.4} \\
& SAPLMA-F & 24.0\tiny{$\pm$12.9} & 23.0\tiny{$\pm$3.7} & 78.1\tiny{$\pm$12.1} & 78.5\tiny{$\pm$12.2} & 50.3\tiny{$\pm$3.0} \\
& CT & 12.1\tiny{$\pm$8.1} & 17.0\tiny{$\pm$8.1} & 78.6\tiny{$\pm$11.5} & 84.7\tiny{$\pm$10.9} & 64.8\tiny{$\pm$9.1} \\
& LitCab & 45.6\tiny{$\pm$11.3} & 20.0\tiny{$\pm$10.8} & 78.3\tiny{$\pm$12.0} & 83.7\tiny{$\pm$10.2} & 65.3\tiny{$\pm$5.4} \\
& CCPS & \textbf{6.3}\tiny{$\pm$3.7} & \textbf{13.1}\tiny{$\pm$5.8} & \textbf{80.2}\tiny{$\pm$9.5} & \textbf{92.1}\tiny{$\pm$8.1} & \textbf{81.6}\tiny{$\pm$7.0} \\
\midrule
\multirow{10}{*}{\textbf{Mistral-Small-24B-Instruct-2501}}
 & P(True) & 42.1\tiny{$\pm$8.5} & 43.3\tiny{$\pm$5.7} & 38.1\tiny{$\pm$7.9} & 80.5\tiny{$\pm$12.1} & 49.3\tiny{$\pm$8.0} \\
& P(IK) & 12.4\tiny{$\pm$9.3} & 17.8\tiny{$\pm$8.7} & 73.9\tiny{$\pm$18.8} & 82.6\tiny{$\pm$12.2} & 56.3\tiny{$\pm$8.9} \\
& LTS & 36.2\tiny{$\pm$9.0} & 38.3\tiny{$\pm$4.7} & 36.1\tiny{$\pm$8.4} & 80.2\tiny{$\pm$12.6} & 49.2\tiny{$\pm$6.2} \\
& IT & 37.3\tiny{$\pm$7.3} & 39.4\tiny{$\pm$5.0} & 42.9\tiny{$\pm$7.7} & 81.3\tiny{$\pm$12.0} & 49.8\tiny{$\pm$7.9} \\
& SAPLMA-M & 24.5\tiny{$\pm$14.0} & 22.5\tiny{$\pm$4.0} & 79.8\tiny{$\pm$12.9} & 79.9\tiny{$\pm$12.8} & 49.8\tiny{$\pm$2.0} \\
& SAPLMA-UM & 24.6\tiny{$\pm$14.1} & 22.5\tiny{$\pm$4.1} & 79.8\tiny{$\pm$12.9} & 80.1\tiny{$\pm$12.9} & 50.6\tiny{$\pm$2.9} \\
& SAPLMA-F & 25.2\tiny{$\pm$14.3} & 22.9\tiny{$\pm$4.1} & 79.8\tiny{$\pm$12.9} & 79.8\tiny{$\pm$12.9} & 49.8\tiny{$\pm$2.3} \\
& CT & 8.2\tiny{$\pm$7.4} & 15.5\tiny{$\pm$7.8} & 79.6\tiny{$\pm$13.1} & 83.3\tiny{$\pm$11.5} & 56.5\tiny{$\pm$7.6} \\
& LitCab & 13.5\tiny{$\pm$6.7} & 15.1\tiny{$\pm$7.4} & 79.5\tiny{$\pm$9.8} & 91.5\tiny{$\pm$8.4} & 78.2\tiny{$\pm$8.0} \\
& CCPS & \textbf{5.8}\tiny{$\pm$3.2} & \textbf{11.5}\tiny{$\pm$6.0} & \textbf{83.0}\tiny{$\pm$10.3} & \textbf{93.1}\tiny{$\pm$7.8} & \textbf{83.3}\tiny{$\pm$7.6} \\
\midrule
\multirow{10}{*}{\textbf{Qwen2.5-32B-Instruct}}
 & P(True) & 44.0\tiny{$\pm$7.0} & 45.7\tiny{$\pm$5.5} & 41.9\tiny{$\pm$7.4} & 84.0\tiny{$\pm$10.3} & 52.1\tiny{$\pm$7.3} \\
& P(IK) & 28.6\tiny{$\pm$12.7} & 23.5\tiny{$\pm$4.2} & 81.7\tiny{$\pm$10.7} & 82.6\tiny{$\pm$10.4} & 49.9\tiny{$\pm$2.9} \\
& LTS & 37.1\tiny{$\pm$6.7} & 40.2\tiny{$\pm$4.4} & 41.9\tiny{$\pm$7.4} & 84.1\tiny{$\pm$10.3} & 52.2\tiny{$\pm$7.3} \\
& IT & 41.9\tiny{$\pm$7.6} & 44.0\tiny{$\pm$6.2} & 44.6\tiny{$\pm$8.3} & 84.6\tiny{$\pm$10.5} & 54.8\tiny{$\pm$7.6} \\
& SAPLMA-M & 27.3\tiny{$\pm$13.2} & 22.7\tiny{$\pm$4.7} & 82.3\tiny{$\pm$10.6} & 82.4\tiny{$\pm$10.6} & 49.7\tiny{$\pm$4.2} \\
& SAPLMA-UM & 27.7\tiny{$\pm$12.8} & 22.8\tiny{$\pm$4.7} & 82.3\tiny{$\pm$10.6} & 82.3\tiny{$\pm$10.7} & 49.4\tiny{$\pm$3.6} \\
& SAPLMA-F & 27.2\tiny{$\pm$12.9} & 22.5\tiny{$\pm$4.5} & 82.3\tiny{$\pm$10.6} & 82.4\tiny{$\pm$10.7} & 49.9\tiny{$\pm$2.8} \\
& CT & 45.2\tiny{$\pm$7.0} & 46.9\tiny{$\pm$5.1} & 37.2\tiny{$\pm$6.1} & 84.3\tiny{$\pm$10.1} & 51.6\tiny{$\pm$8.0} \\
& LitCab & 43.2\tiny{$\pm$11.0} & 15.9\tiny{$\pm$9.3} & 82.6\tiny{$\pm$10.4} & 87.9\tiny{$\pm$7.9} & 67.2\tiny{$\pm$6.5} \\
& CCPS & \textbf{6.3}\tiny{$\pm$3.1} & \textbf{10.8}\tiny{$\pm$5.2} & \textbf{84.1}\tiny{$\pm$8.9} & \textbf{94.1}\tiny{$\pm$5.9} & \textbf{82.8}\tiny{$\pm$6.9} \\
\bottomrule
\end{tabular}
}
\caption{Complete performance metrics for the MMLU-CHOICE dataset. Arrows indicate whether lower ($\downarrow$) or higher ($\uparrow$) values are better. All values are percentages and show mean $\pm$ standard deviation. Best values per model are bolded.}
\label{tab:appendix_mmluchoice}
\end{table*}

\begin{table*}[t]
\centering
\small
\resizebox{\textwidth}{!}{
\begin{tabular}{llccccc}
\toprule
\textbf{Model} & \textbf{Method} & \textbf{ECE $\downarrow$} & \textbf{BRIER $\downarrow$} & \textbf{ACC $\uparrow$} & \textbf{AUCPR $\uparrow$} & \textbf{AUROC $\uparrow$} \\
\midrule
\multirow{10}{*}{\textbf{Meta-Llama-3.1-8B-Instruct}}
 & P(True) & 25.3\tiny{$\pm$6.7} & 33.1\tiny{$\pm$4.5} & 54.8\tiny{$\pm$6.8} & 37.1\tiny{$\pm$11.7} & 49.8\tiny{$\pm$2.1} \\
& P(IK) & 41.7\tiny{$\pm$15.6} & 44.1\tiny{$\pm$11.5} & 38.2\tiny{$\pm$11.7} & 37.3\tiny{$\pm$13.7} & 49.9\tiny{$\pm$3.2} \\
& LTS & 17.0\tiny{$\pm$6.7} & 29.1\tiny{$\pm$3.5} & 55.4\tiny{$\pm$7.0} & 36.9\tiny{$\pm$12.3} & 49.9\tiny{$\pm$2.2} \\
& IT & 26.8\tiny{$\pm$4.7} & 33.8\tiny{$\pm$2.9} & 52.8\tiny{$\pm$4.8} & 37.7\tiny{$\pm$12.2} & 50.0\tiny{$\pm$2.7} \\
& SAPLMA-M & 40.4\tiny{$\pm$14.0} & 40.3\tiny{$\pm$8.6} & 36.7\tiny{$\pm$12.7} & 37.3\tiny{$\pm$13.0} & 50.1\tiny{$\pm$1.8} \\
& SAPLMA-UM & 41.0\tiny{$\pm$14.3} & 41.0\tiny{$\pm$9.1} & 36.7\tiny{$\pm$12.7} & 37.5\tiny{$\pm$13.3} & 50.3\tiny{$\pm$3.0} \\
& SAPLMA-F & 40.2\tiny{$\pm$14.9} & 40.7\tiny{$\pm$10.0} & 36.8\tiny{$\pm$12.7} & 37.2\tiny{$\pm$12.9} & 50.3\tiny{$\pm$1.8} \\
& CT & 21.5\tiny{$\pm$11.5} & 29.8\tiny{$\pm$5.9} & 50.4\tiny{$\pm$11.7} & 43.7\tiny{$\pm$14.4} & 57.3\tiny{$\pm$4.4} \\
& LitCab & 16.6\tiny{$\pm$2.9} & 24.7\tiny{$\pm$2.6} & 66.1\tiny{$\pm$4.2} & 51.7\tiny{$\pm$18.4} & 63.6\tiny{$\pm$9.0} \\
& CCPS & \textbf{4.5}\tiny{$\pm$2.1} & \textbf{20.0}\tiny{$\pm$2.2} & \textbf{70.4}\tiny{$\pm$4.0} & \textbf{55.2}\tiny{$\pm$19.4} & \textbf{67.9}\tiny{$\pm$8.1} \\
\midrule
\multirow{10}{*}{\textbf{Qwen2.5-14B-Instruct}}
 & P(True) & 33.7\tiny{$\pm$7.1} & 38.6\tiny{$\pm$4.8} & 49.9\tiny{$\pm$6.0} & 55.4\tiny{$\pm$12.6} & 51.4\tiny{$\pm$1.3} \\
& P(IK) & 27.3\tiny{$\pm$11.4} & 33.9\tiny{$\pm$8.2} & 53.6\tiny{$\pm$11.5} & 53.5\tiny{$\pm$13.4} & 49.1\tiny{$\pm$2.3} \\
& LTS & 26.7\tiny{$\pm$7.4} & 34.5\tiny{$\pm$4.5} & 49.3\tiny{$\pm$6.7} & 54.6\tiny{$\pm$11.8} & 50.7\tiny{$\pm$2.2} \\
& IT & 33.5\tiny{$\pm$6.0} & 38.3\tiny{$\pm$3.6} & 50.5\tiny{$\pm$4.2} & 55.6\tiny{$\pm$12.1} & 51.1\tiny{$\pm$2.4} \\
& SAPLMA-M & 28.1\tiny{$\pm$13.2} & 33.4\tiny{$\pm$8.6} & 53.4\tiny{$\pm$12.5} & 54.1\tiny{$\pm$13.3} & 50.1\tiny{$\pm$3.0} \\
& SAPLMA-UM & 27.4\tiny{$\pm$13.5} & 33.0\tiny{$\pm$8.6} & 53.5\tiny{$\pm$12.5} & 53.8\tiny{$\pm$13.0} & 49.9\tiny{$\pm$3.1} \\
& SAPLMA-F & 25.7\tiny{$\pm$12.3} & 32.1\tiny{$\pm$7.7} & 53.4\tiny{$\pm$12.5} & 53.8\tiny{$\pm$12.5} & 49.3\tiny{$\pm$2.8} \\
& CT & 20.4\tiny{$\pm$10.3} & 28.7\tiny{$\pm$6.3} & 55.6\tiny{$\pm$11.4} & 59.4\tiny{$\pm$12.9} & 56.6\tiny{$\pm$3.5} \\
& LitCab & 49.7\tiny{$\pm$4.2} & 38.3\tiny{$\pm$8.8} & 55.3\tiny{$\pm$11.6} & 66.2\tiny{$\pm$10.1} & 68.0\tiny{$\pm$3.7} \\
& CCPS & \textbf{4.2}\tiny{$\pm$1.8} & \textbf{20.1}\tiny{$\pm$2.9} & \textbf{69.2}\tiny{$\pm$5.4} & \textbf{75.8}\tiny{$\pm$10.5} & \textbf{74.0}\tiny{$\pm$4.8} \\
\midrule
\multirow{10}{*}{\textbf{Mistral-Small-24B-Instruct-2501}}
 & P(True) & 32.0\tiny{$\pm$8.1} & 37.2\tiny{$\pm$5.0} & 46.9\tiny{$\pm$7.3} & 57.5\tiny{$\pm$12.3} & 50.2\tiny{$\pm$2.2} \\
& P(IK) & 32.3\tiny{$\pm$11.4} & 36.3\tiny{$\pm$9.7} & 56.1\tiny{$\pm$10.9} & 57.4\tiny{$\pm$13.4} & 50.6\tiny{$\pm$2.1} \\
& LTS & 24.7\tiny{$\pm$7.6} & 32.7\tiny{$\pm$3.7} & 46.2\tiny{$\pm$7.2} & 56.6\tiny{$\pm$12.3} & 49.2\tiny{$\pm$1.7} \\
& IT & 31.2\tiny{$\pm$6.7} & 36.2\tiny{$\pm$3.9} & 47.0\tiny{$\pm$6.1} & 58.4\tiny{$\pm$12.0} & 50.3\tiny{$\pm$2.8} \\
& SAPLMA-M & 24.5\tiny{$\pm$12.1} & 30.7\tiny{$\pm$8.0} & 56.7\tiny{$\pm$12.4} & 57.0\tiny{$\pm$13.3} & 49.9\tiny{$\pm$2.8} \\
& SAPLMA-UM & 24.5\tiny{$\pm$11.7} & 30.7\tiny{$\pm$8.0} & 56.7\tiny{$\pm$12.4} & 57.6\tiny{$\pm$13.4} & 50.6\tiny{$\pm$3.1} \\
& SAPLMA-F & 25.1\tiny{$\pm$12.8} & 31.4\tiny{$\pm$8.5} & 56.7\tiny{$\pm$12.4} & 56.8\tiny{$\pm$12.2} & 49.8\tiny{$\pm$2.2} \\
& CT & 17.8\tiny{$\pm$9.7} & 27.4\tiny{$\pm$5.9} & 58.2\tiny{$\pm$11.6} & 60.1\tiny{$\pm$13.1} & 54.3\tiny{$\pm$3.1} \\
& LitCab & 32.2\tiny{$\pm$3.1} & 34.6\tiny{$\pm$3.2} & 57.0\tiny{$\pm$3.7} & 66.2\tiny{$\pm$12.8} & 60.1\tiny{$\pm$5.0} \\
& CCPS & \textbf{4.5}\tiny{$\pm$1.9} & \textbf{18.6}\tiny{$\pm$3.3} & \textbf{71.3}\tiny{$\pm$6.4} & \textbf{79.5}\tiny{$\pm$9.4} & \textbf{77.2}\tiny{$\pm$5.2} \\
\midrule
\multirow{10}{*}{\textbf{Qwen2.5-32B-Instruct}}
 & P(True) & 34.6\tiny{$\pm$6.8} & 39.5\tiny{$\pm$4.9} & 46.1\tiny{$\pm$5.9} & 60.1\tiny{$\pm$12.2} & 50.3\tiny{$\pm$2.7} \\
& P(IK) & 23.6\tiny{$\pm$9.9} & 30.8\tiny{$\pm$7.7} & 58.0\tiny{$\pm$10.8} & 59.5\tiny{$\pm$11.9} & 50.2\tiny{$\pm$2.5} \\
& LTS & 27.5\tiny{$\pm$6.7} & 34.8\tiny{$\pm$3.9} & 46.1\tiny{$\pm$5.9} & 60.1\tiny{$\pm$12.2} & 50.3\tiny{$\pm$2.7} \\
& IT & 36.6\tiny{$\pm$6.7} & 40.9\tiny{$\pm$5.3} & 45.9\tiny{$\pm$5.9} & 60.1\tiny{$\pm$12.1} & 51.0\tiny{$\pm$2.7} \\
& SAPLMA-M & 24.8\tiny{$\pm$12.0} & 30.5\tiny{$\pm$8.3} & 59.3\tiny{$\pm$11.8} & 59.9\tiny{$\pm$12.1} & 49.9\tiny{$\pm$2.8} \\
& SAPLMA-UM & 26.9\tiny{$\pm$12.1} & 31.8\tiny{$\pm$8.9} & 59.3\tiny{$\pm$11.8} & 60.2\tiny{$\pm$12.0} & 49.8\tiny{$\pm$3.3} \\
& SAPLMA-F & 23.7\tiny{$\pm$11.2} & 30.0\tiny{$\pm$7.9} & 59.3\tiny{$\pm$11.8} & 59.4\tiny{$\pm$11.6} & 49.5\tiny{$\pm$2.7} \\
& CT & 38.0\tiny{$\pm$8.5} & 41.6\tiny{$\pm$6.4} & 44.8\tiny{$\pm$7.2} & 60.5\tiny{$\pm$11.3} & 49.9\tiny{$\pm$2.7} \\
& LitCab & 48.4\tiny{$\pm$3.5} & 33.7\tiny{$\pm$8.7} & 60.8\tiny{$\pm$11.0} & 72.7\tiny{$\pm$8.8} & 70.3\tiny{$\pm$4.7} \\
& CCPS & \textbf{4.6}\tiny{$\pm$2.1} & \textbf{18.5}\tiny{$\pm$3.4} & \textbf{71.8}\tiny{$\pm$6.1} & \textbf{82.4}\tiny{$\pm$7.7} & \textbf{77.8}\tiny{$\pm$4.7} \\
\bottomrule
\end{tabular}
}
\caption{Complete performance metrics for the MMLU-PRO-CHOICE dataset. Arrows indicate whether lower ($\downarrow$) or higher ($\uparrow$) values are better. All values are percentages and show mean $\pm$ standard deviation. Best values per model are bolded.}
\label{tab:appendix_mmluprochoice}
\end{table*}

\begin{table*}[t]
\centering
\small
\resizebox{\textwidth}{!}{
\begin{tabular}{llccccc}
\toprule
\textbf{Model} & \textbf{Method} & \textbf{ECE $\downarrow$} & \textbf{BRIER $\downarrow$} & \textbf{ACC $\uparrow$} & \textbf{AUCPR $\uparrow$} & \textbf{AUROC $\uparrow$} \\
\midrule
\multirow{10}{*}{\textbf{Meta-Llama-3.1-8B-Instruct}}
 & P(True) & 25.9\tiny{$\pm$7.0} & 32.0\tiny{$\pm$5.2} & 56.0\tiny{$\pm$7.7} & 29.9\tiny{$\pm$12.5} & 46.2\tiny{$\pm$5.8} \\
& P(IK) & 22.6\tiny{$\pm$12.0} & 26.6\tiny{$\pm$5.0} & 30.5\tiny{$\pm$11.8} & 29.9\tiny{$\pm$11.5} & 49.8\tiny{$\pm$1.1} \\
& LTS & 27.9\tiny{$\pm$5.6} & 34.0\tiny{$\pm$3.5} & 47.8\tiny{$\pm$4.9} & 31.5\tiny{$\pm$13.6} & 47.5\tiny{$\pm$5.9} \\
& IT & 27.8\tiny{$\pm$5.9} & 33.2\tiny{$\pm$4.6} & 53.9\tiny{$\pm$6.1} & 30.6\tiny{$\pm$13.3} & 47.2\tiny{$\pm$6.0} \\
& SAPLMA-M & 23.0\tiny{$\pm$12.5} & 26.6\tiny{$\pm$5.2} & 67.0\tiny{$\pm$15.7} & 29.6\tiny{$\pm$11.3} & 49.9\tiny{$\pm$0.9} \\
& SAPLMA-UM & 22.8\tiny{$\pm$11.7} & 26.2\tiny{$\pm$3.5} & 29.6\tiny{$\pm$11.3} & 29.6\tiny{$\pm$11.3} & 49.9\tiny{$\pm$0.9} \\
& SAPLMA-F & 22.5\tiny{$\pm$11.5} & 26.1\tiny{$\pm$3.5} & 29.7\tiny{$\pm$11.4} & 29.7\tiny{$\pm$11.4} & 49.8\tiny{$\pm$1.4} \\
& CT & 8.8\tiny{$\pm$6.4} & 21.1\tiny{$\pm$4.9} & 65.3\tiny{$\pm$11.4} & 48.9\tiny{$\pm$16.8} & \textbf{70.9}\tiny{$\pm$7.5} \\
& LitCab & 8.8\tiny{$\pm$7.6} & 22.5\tiny{$\pm$4.8} & 65.3\tiny{$\pm$9.1} & 46.2\tiny{$\pm$13.8} & 66.0\tiny{$\pm$9.6} \\
& CCPS & \textbf{8.0}\tiny{$\pm$5.7} & \textbf{20.2}\tiny{$\pm$3.8} & \textbf{69.5}\tiny{$\pm$8.6} & \textbf{49.4}\tiny{$\pm$15.9} & 69.3\tiny{$\pm$7.8} \\
\midrule
\multirow{10}{*}{\textbf{Qwen2.5-14B-Instruct}}
 & P(True) & 33.9\tiny{$\pm$7.6} & 36.9\tiny{$\pm$5.9} & 54.1\tiny{$\pm$7.4} & 46.3\tiny{$\pm$12.6} & 52.6\tiny{$\pm$5.0} \\
& P(IK) & 14.1\tiny{$\pm$9.9} & 26.3\tiny{$\pm$4.3} & 55.8\tiny{$\pm$12.7} & 42.8\tiny{$\pm$12.2} & 49.5\tiny{$\pm$1.9} \\
& LTS & 27.8\tiny{$\pm$5.4} & 32.8\tiny{$\pm$4.0} & 55.9\tiny{$\pm$5.6} & 49.2\tiny{$\pm$13.1} & 56.5\tiny{$\pm$6.0} \\
& IT & 33.6\tiny{$\pm$6.1} & 36.7\tiny{$\pm$4.8} & 55.0\tiny{$\pm$6.0} & 47.5\tiny{$\pm$13.9} & 54.0\tiny{$\pm$5.9} \\
& SAPLMA-M & 14.9\tiny{$\pm$10.8} & 26.4\tiny{$\pm$4.2} & 42.8\tiny{$\pm$12.3} & 42.8\tiny{$\pm$12.3} & 49.9\tiny{$\pm$0.9} \\
& SAPLMA-UM & 14.6\tiny{$\pm$9.9} & 26.1\tiny{$\pm$3.0} & 42.8\tiny{$\pm$12.3} & 42.7\tiny{$\pm$12.3} & 49.9\tiny{$\pm$0.9} \\
& SAPLMA-F & 14.7\tiny{$\pm$10.1} & 26.2\tiny{$\pm$3.3} & 42.8\tiny{$\pm$12.3} & 42.8\tiny{$\pm$12.4} & 49.9\tiny{$\pm$1.0} \\
& CT & 9.4\tiny{$\pm$5.6} & 22.6\tiny{$\pm$4.0} & 63.4\tiny{$\pm$8.4} & \textbf{61.7}\tiny{$\pm$14.3} & \textbf{69.3}\tiny{$\pm$7.7} \\
& LitCab & 34.4\tiny{$\pm$10.3} & 37.0\tiny{$\pm$7.3} & 49.4\tiny{$\pm$10.1} & 56.8\tiny{$\pm$13.4} & 62.5\tiny{$\pm$6.8} \\
& CCPS & \textbf{6.7}\tiny{$\pm$3.5} & \textbf{22.5}\tiny{$\pm$2.0} & \textbf{63.6}\tiny{$\pm$6.8} & 59.0\tiny{$\pm$12.7} & 66.6\tiny{$\pm$6.8} \\
\midrule
\multirow{10}{*}{\textbf{Mistral-Small-24B-Instruct-2501}}
 & P(True) & 28.0\tiny{$\pm$8.9} & 33.5\tiny{$\pm$6.7} & 55.5\tiny{$\pm$8.7} & 44.6\tiny{$\pm$13.3} & 49.8\tiny{$\pm$4.5} \\
& P(IK) & 19.9\tiny{$\pm$12.7} & 29.7\tiny{$\pm$7.4} & 52.5\tiny{$\pm$11.1} & 46.3\tiny{$\pm$14.4} & 52.7\tiny{$\pm$5.2} \\
& LTS & 19.4\tiny{$\pm$6.3} & 29.3\tiny{$\pm$4.0} & 55.2\tiny{$\pm$6.7} & 46.1\tiny{$\pm$13.8} & 50.8\tiny{$\pm$5.3} \\
& IT & 26.2\tiny{$\pm$7.9} & 32.5\tiny{$\pm$5.6} & 55.2\tiny{$\pm$7.4} & 45.5\tiny{$\pm$13.6} & 50.6\tiny{$\pm$4.5} \\
& SAPLMA-M & 15.1\tiny{$\pm$10.9} & 26.2\tiny{$\pm$3.3} & 42.6\tiny{$\pm$13.0} & 42.9\tiny{$\pm$12.9} & 50.2\tiny{$\pm$1.0} \\
& SAPLMA-UM & 15.2\tiny{$\pm$11.0} & 26.3\tiny{$\pm$3.5} & 42.6\tiny{$\pm$13.0} & 42.8\tiny{$\pm$13.0} & 50.1\tiny{$\pm$0.8} \\
& SAPLMA-F & 14.9\tiny{$\pm$10.9} & 26.2\tiny{$\pm$3.4} & 42.6\tiny{$\pm$13.0} & 42.7\tiny{$\pm$13.0} & 50.0\tiny{$\pm$1.6} \\
& CT & 10.8\tiny{$\pm$5.4} & 22.8\tiny{$\pm$3.4} & 62.2\tiny{$\pm$8.3} & 60.7\tiny{$\pm$15.8} & 68.2\tiny{$\pm$8.0} \\
& LitCab & 11.2\tiny{$\pm$5.0} & 24.6\tiny{$\pm$3.1} & 60.2\tiny{$\pm$6.8} & 60.5\tiny{$\pm$13.3} & 66.4\tiny{$\pm$6.5} \\
& CCPS & \textbf{6.8}\tiny{$\pm$2.6} & \textbf{20.8}\tiny{$\pm$2.6} & \textbf{67.6}\tiny{$\pm$6.0} & \textbf{64.7}\tiny{$\pm$13.2} & \textbf{71.4}\tiny{$\pm$6.8} \\
\midrule
\multirow{10}{*}{\textbf{Qwen2.5-32B-Instruct}}
 & P(True) & 36.3\tiny{$\pm$4.6} & 38.0\tiny{$\pm$3.7} & 54.8\tiny{$\pm$4.3} & 53.8\tiny{$\pm$12.9} & 57.1\tiny{$\pm$5.5} \\
& P(IK) & 13.1\tiny{$\pm$10.4} & 26.3\tiny{$\pm$4.5} & 52.8\tiny{$\pm$12.5} & 46.5\tiny{$\pm$12.3} & 49.9\tiny{$\pm$0.6} \\
& LTS & 29.5\tiny{$\pm$4.8} & 34.4\tiny{$\pm$3.4} & 53.7\tiny{$\pm$4.0} & 52.7\tiny{$\pm$13.3} & 55.5\tiny{$\pm$5.5} \\
& IT & 33.2\tiny{$\pm$7.1} & 37.3\tiny{$\pm$5.3} & 52.7\tiny{$\pm$6.8} & 49.0\tiny{$\pm$12.7} & 51.6\tiny{$\pm$4.5} \\
& SAPLMA-M & 13.6\tiny{$\pm$10.0} & 26.2\tiny{$\pm$3.7} & 46.2\tiny{$\pm$12.5} & 46.2\tiny{$\pm$12.6} & 49.9\tiny{$\pm$0.6} \\
& SAPLMA-UM & 13.7\tiny{$\pm$10.4} & 26.3\tiny{$\pm$4.1} & 46.1\tiny{$\pm$12.5} & 46.2\tiny{$\pm$12.5} & 49.8\tiny{$\pm$1.2} \\
& SAPLMA-F & 13.7\tiny{$\pm$10.2} & 26.3\tiny{$\pm$3.8} & 46.1\tiny{$\pm$12.5} & 46.2\tiny{$\pm$12.5} & 49.8\tiny{$\pm$0.8} \\
& CT & 22.9\tiny{$\pm$4.7} & 31.1\tiny{$\pm$3.5} & 57.1\tiny{$\pm$5.2} & 52.9\tiny{$\pm$12.8} & 56.3\tiny{$\pm$5.7} \\
& LitCab & 28.4\tiny{$\pm$8.1} & 33.2\tiny{$\pm$5.5} & 52.7\tiny{$\pm$8.5} & 60.2\tiny{$\pm$13.0} & 62.3\tiny{$\pm$7.6} \\
& CCPS & \textbf{8.7}\tiny{$\pm$4.9} & \textbf{23.3}\tiny{$\pm$2.1} & \textbf{62.6}\tiny{$\pm$6.8} & \textbf{62.0}\tiny{$\pm$11.8} & \textbf{66.4}\tiny{$\pm$5.8} \\
\bottomrule
\end{tabular}
}
\caption{Complete performance metrics for the MMLU-OE dataset. Arrows indicate whether lower ($\downarrow$) or higher ($\uparrow$) values are better. All values are percentages and show mean $\pm$ standard deviation. Best values per model are bolded.}
\label{tab:appendix_mmluoe}
\end{table*}

\subsection{Per-LLM Performance Bar Charts}
\begin{figure*}[ht]
  \centering
  \includegraphics[width=0.9\textwidth]{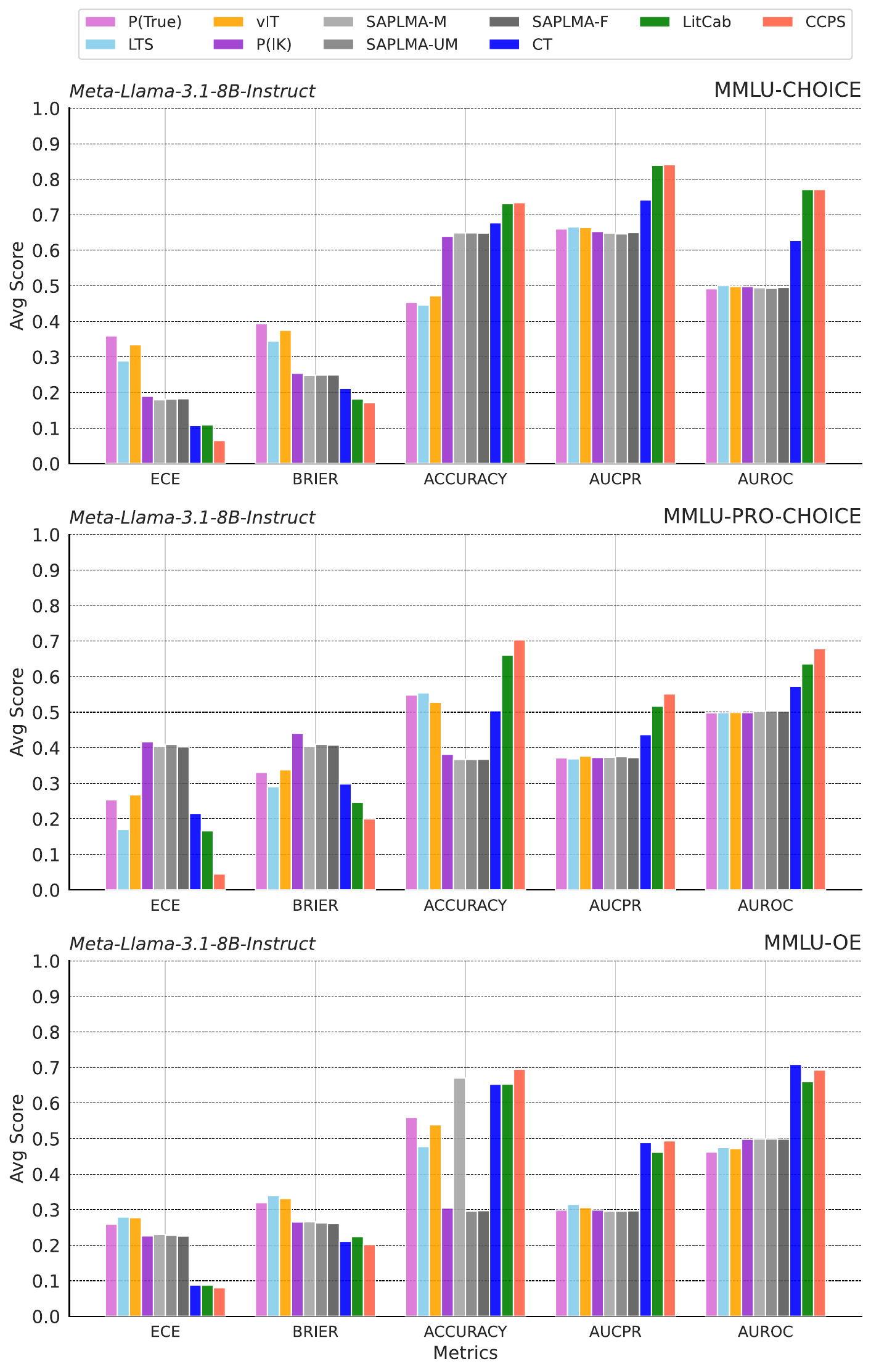}
  \caption{Performance comparison of confidence estimation methods on \texttt{Meta-Llama-3.1-8B-Instruct} across MMLU variants.}
  \label{fig:Meta-Llama-3.1-8B-Instruct_metric_bars}
\end{figure*}

\begin{figure*}[ht]
  \centering
  \includegraphics[width=0.9\textwidth]{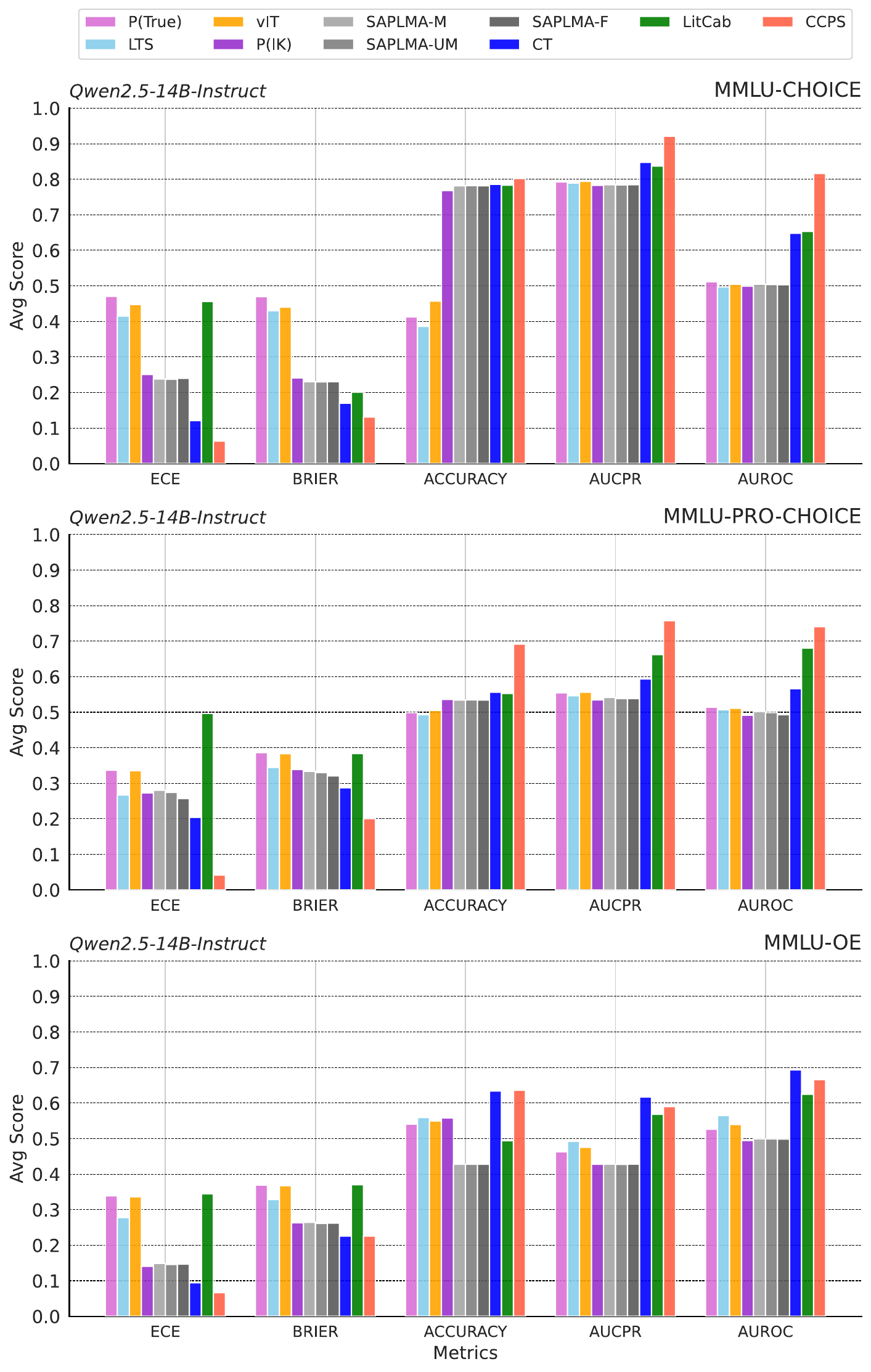}
  \caption{Performance comparison of confidence estimation methods on \texttt{Qwen2.5-14B-Instruct} across MMLU variants.}
  \label{fig:Qwen2.5-14B-Instruct_metric_bars}
\end{figure*}

\begin{figure*}[ht]
  \centering
  \includegraphics[width=0.9\textwidth]{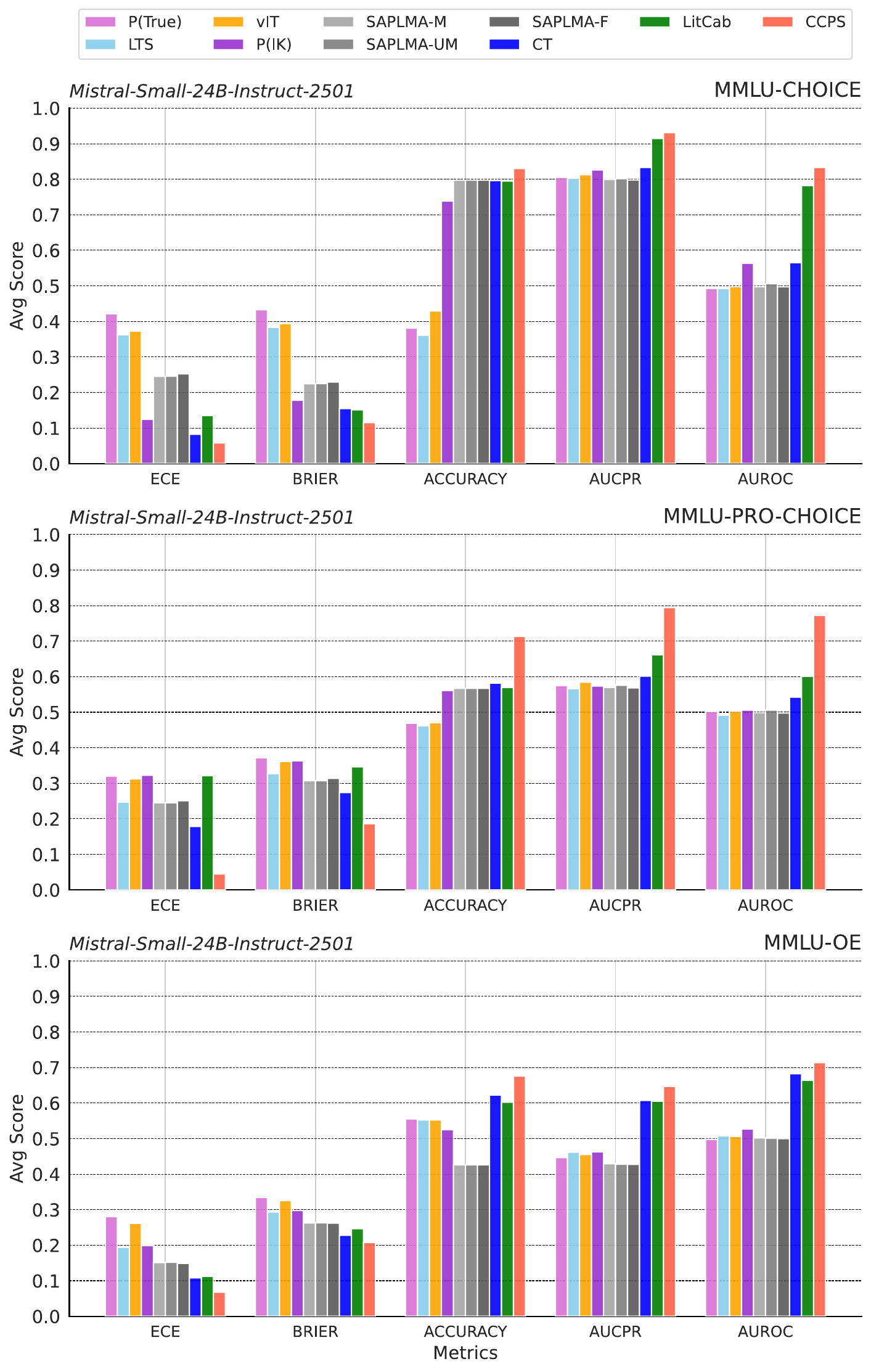}
  \caption{Performance comparison of confidence estimation methods on \texttt{Mistral-Small-24B-Instruct-2501} across MMLU variants.}
  \label{fig:Mistral-Small-24B-Instruct-2501_metric_bars}
\end{figure*}

\begin{figure*}[ht]
  \centering
  \includegraphics[width=0.9\textwidth]{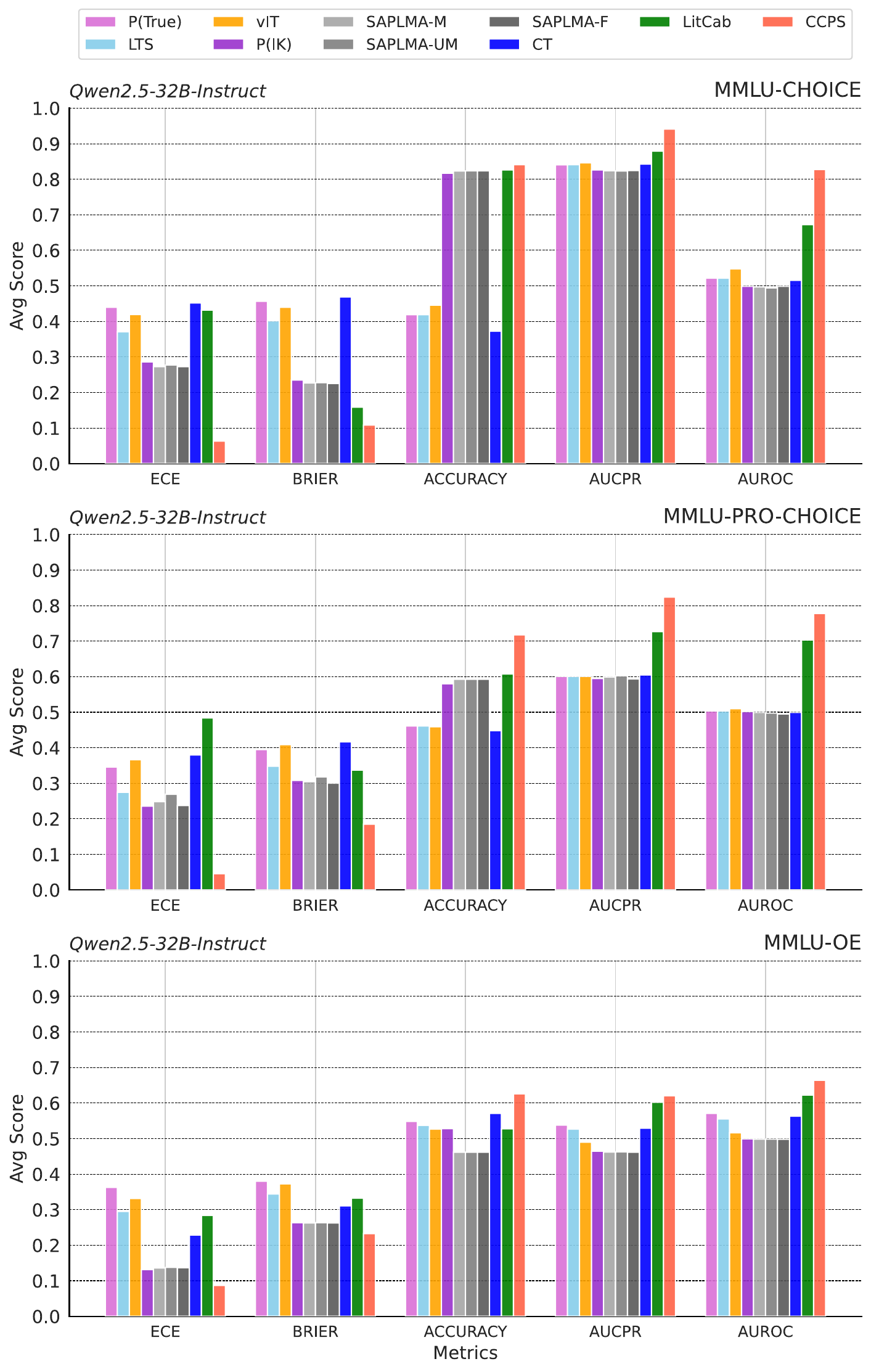}
  \caption{Performance comparison of confidence estimation methods on \texttt{Qwen2.5-32B-Instruct} across MMLU variants.}
  \label{fig:Qwen2.5-32B-Instruct_metric_bars}
\end{figure*}

For a visual comparison of method performance on each specific LLM, Figures \ref{fig:Meta-Llama-3.1-8B-Instruct_metric_bars}, \ref{fig:Qwen2.5-14B-Instruct_metric_bars}, \ref{fig:Mistral-Small-24B-Instruct-2501_metric_bars}, and \ref{fig:Qwen2.5-32B-Instruct_metric_bars} present bar charts. Each figure corresponds to one of the four LLMs used in our experiments, illustrating the performance of every confidence estimation method across the different MMLU variant datasets on all evaluation metrics.

\subsection{Calibration Curves}
\begin{figure*}[t]
  \centering
  \includegraphics[width=\textwidth]{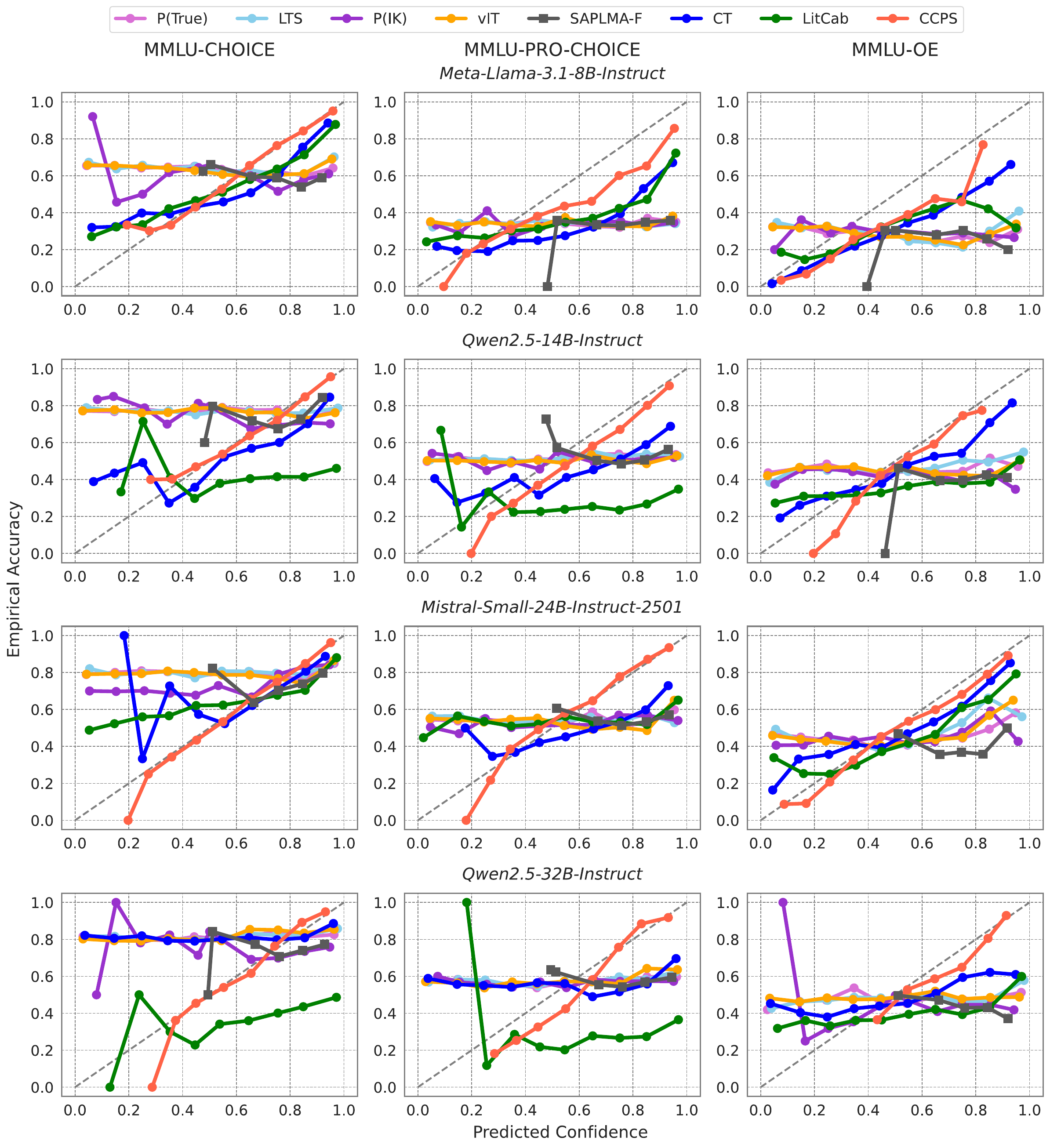}
  \caption{Calibration curves of confidence estimation methods across all models and MMLU variants.}
  \label{fig:calibration_curves_43}
\end{figure*}

\begin{figure*}[ht]
  \centering
  \includegraphics[width=0.75\textwidth]{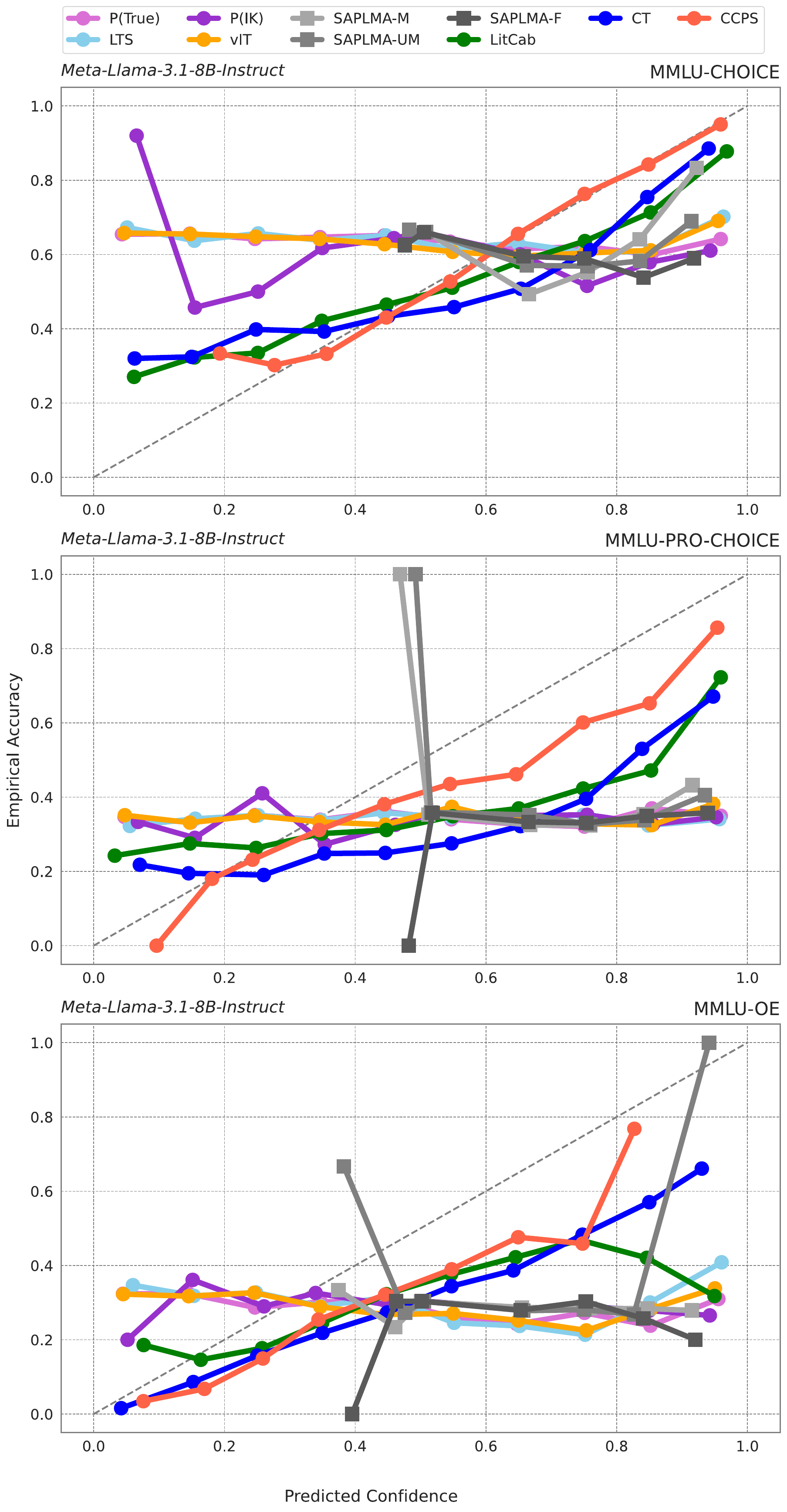}
  \caption{Calibration curves of confidence estimation methods on \texttt{Meta-Llama-3.1-8B-Instruct} across MMLU variants.}
  \label{fig:Meta-Llama-3.1-8B-Instruct_calibration}
\end{figure*}

\begin{figure*}[ht]
  \centering
  \includegraphics[width=0.75\textwidth]{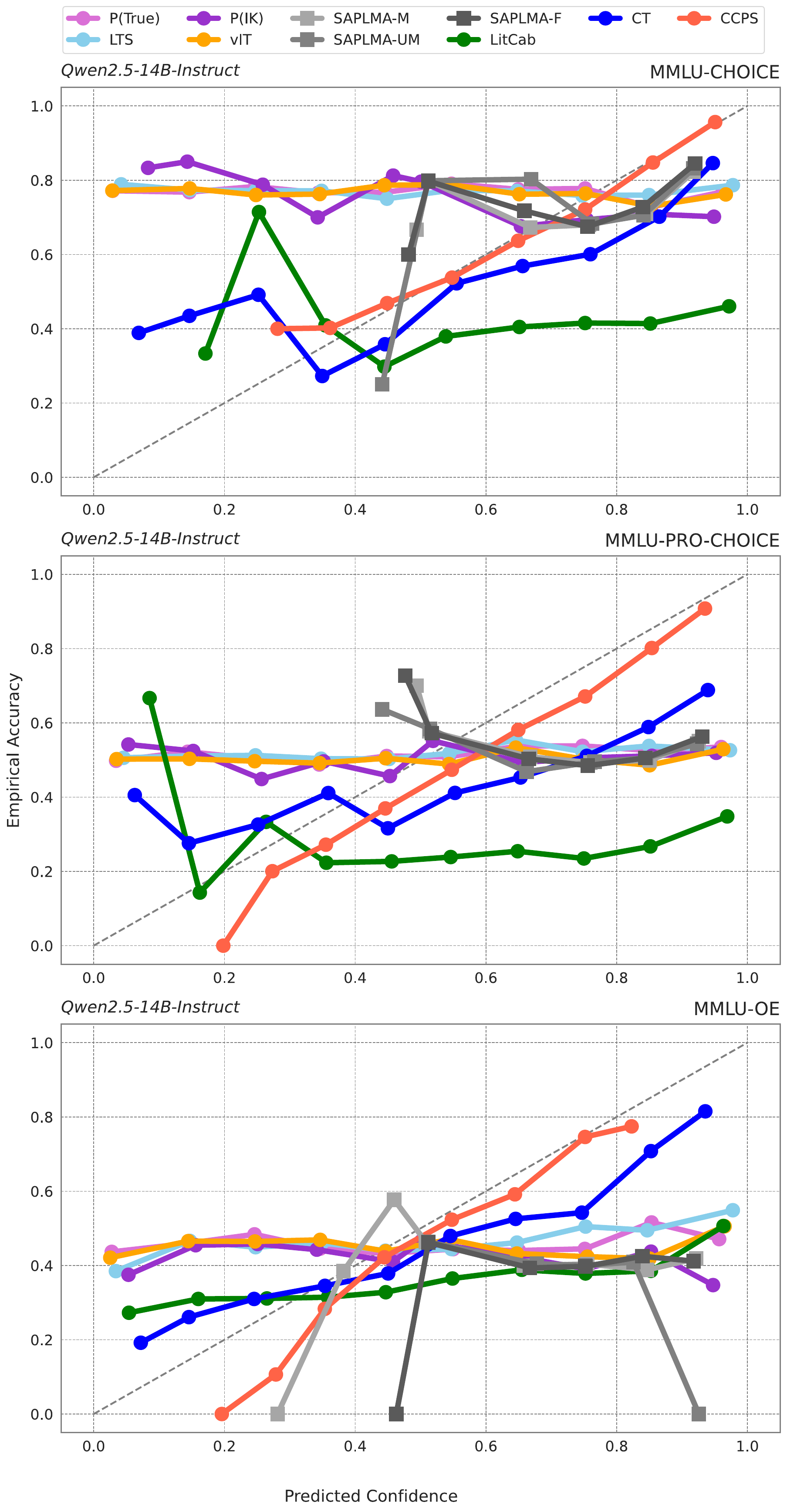}
  \caption{Calibration curves of confidence estimation methods on \texttt{Qwen2.5-14B-Instruct} across MMLU variants.}
  \label{fig:Qwen2.5-14B-Instruct_calibration}
\end{figure*}

\begin{figure*}[ht]
  \centering
  \includegraphics[width=0.75\textwidth]{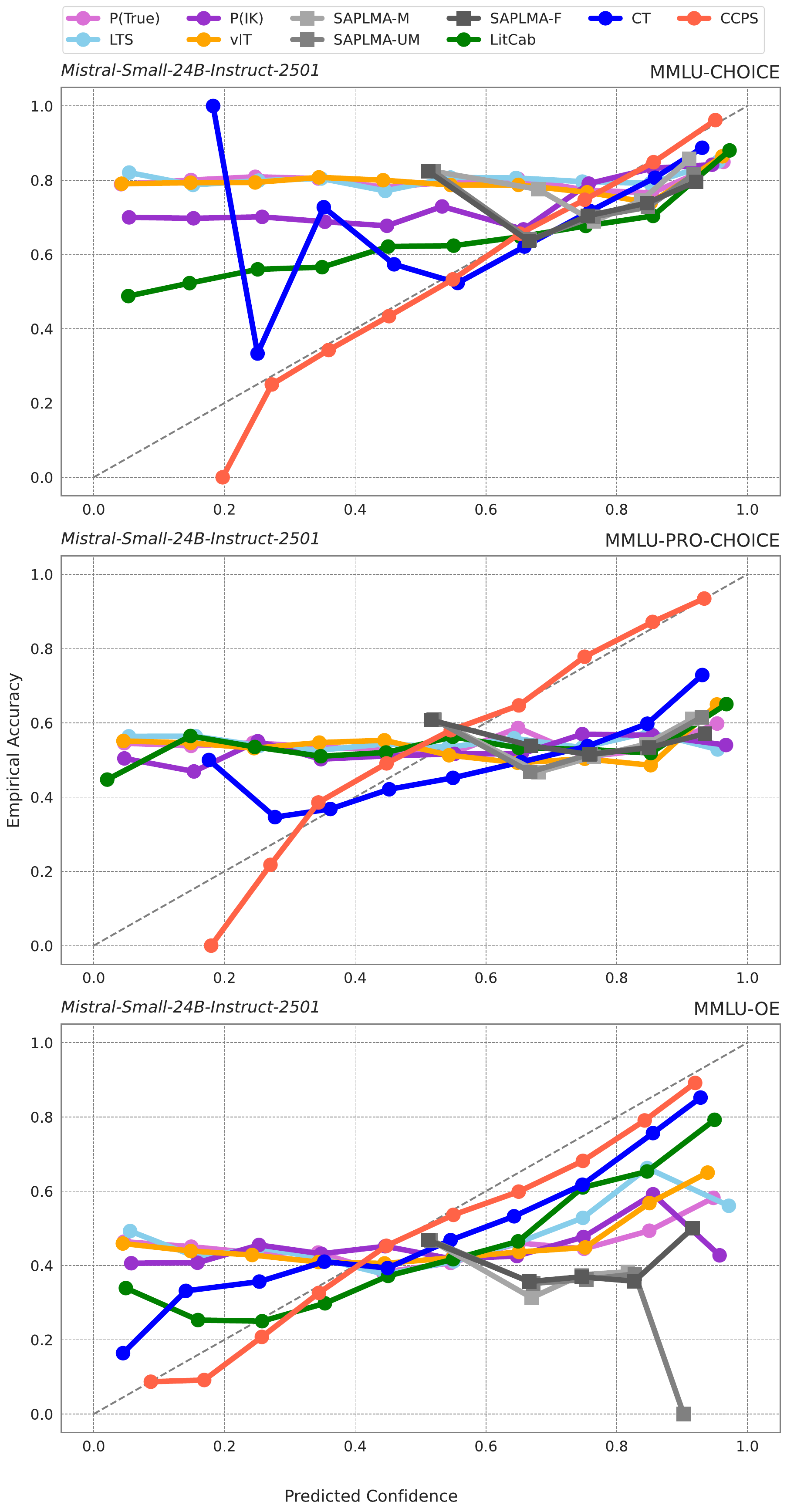}
  \caption{Calibration curves of confidence estimation methods on \texttt{Mistral-Small-24B-Instruct-2501} across MMLU variants.}
  \label{fig:Mistral-Small-24B-Instruct-2501_calibration}
\end{figure*}

\begin{figure*}[ht]
  \centering
  \includegraphics[width=0.75\textwidth]{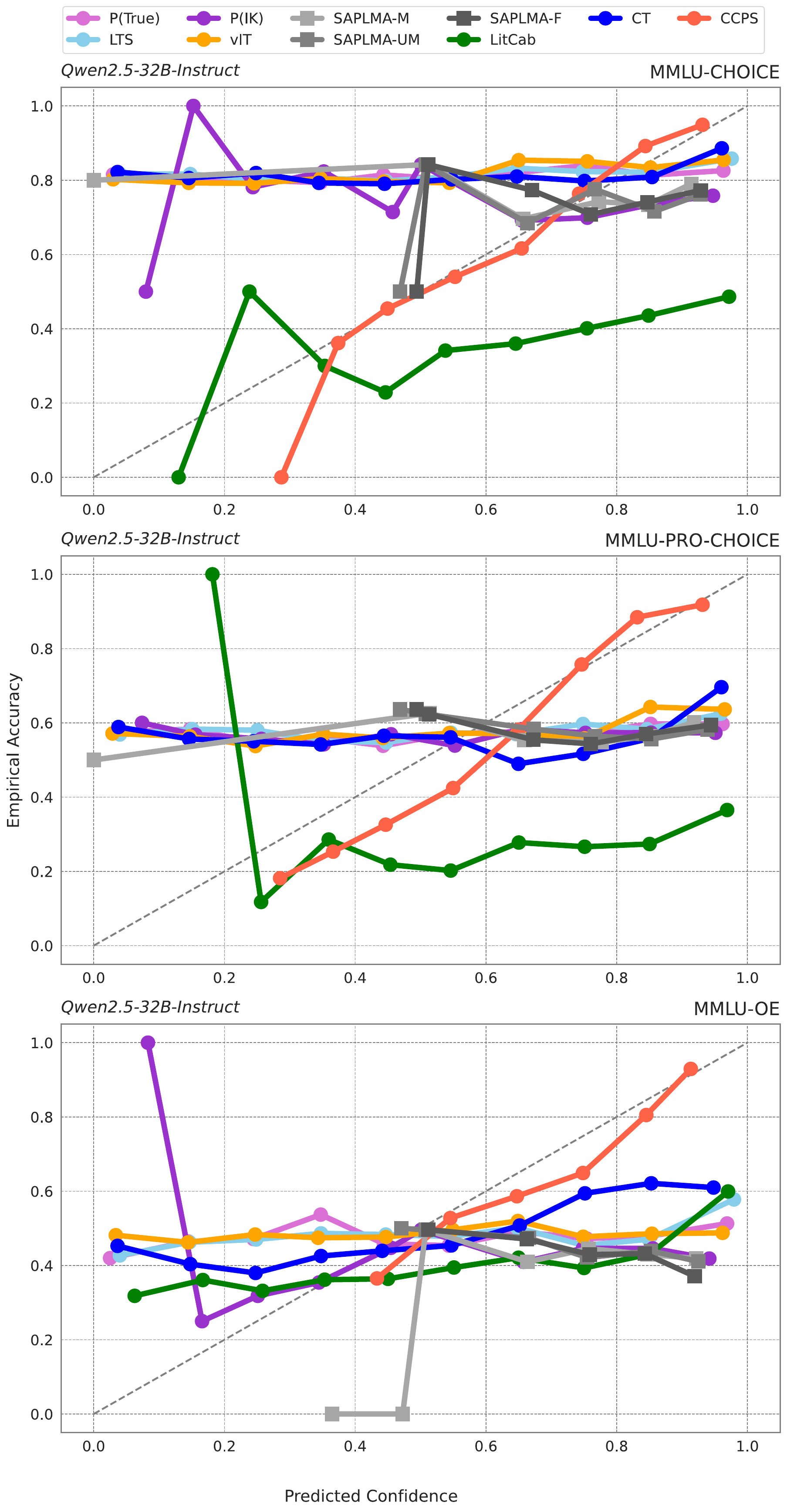}
  \caption{Calibration curves of confidence estimation methods on \texttt{Qwen2.5-32B-Instruct} across MMLU variants.}
  \label{fig:Qwen2.5-32B-Instruct_calibration}
\end{figure*}

To visually assess the calibration of the confidence scores produced by different methods, we provide calibration curves. Figure \ref{fig:calibration_curves_43} offers an overview, displaying calibration curves across all models and MMLU variants. Additionally, Figures \ref{fig:Meta-Llama-3.1-8B-Instruct_calibration}, \ref{fig:Qwen2.5-14B-Instruct_calibration}, \ref{fig:Mistral-Small-24B-Instruct-2501_calibration}, and \ref{fig:Qwen2.5-32B-Instruct_calibration} present detailed calibration curves for each specific LLM across the test datasets, allowing for a more granular inspection of calibration performance.

\subsection{Per-Task Performance Analysis}
\begin{figure*}[ht]
  \centering
  \includegraphics[width=\textwidth]{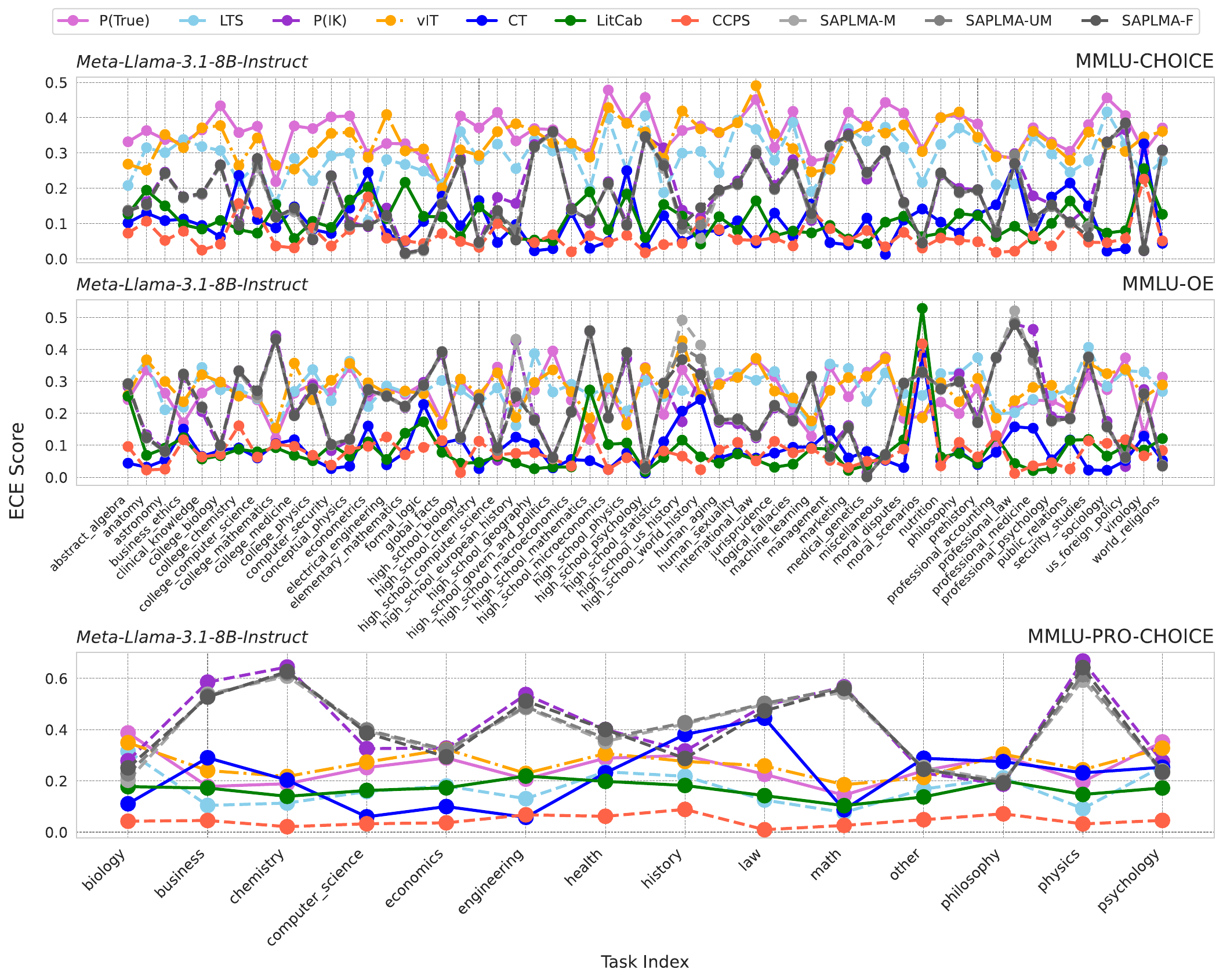}
  \caption{ECE comparison of confidence estimation methods on \texttt{Meta-Llama-3.1-8B-Instruct} across different tasks of MMLU variants.}
  \label{fig:llama_ece}
\end{figure*}

\begin{figure*}[ht]
  \centering
  \includegraphics[width=\textwidth]{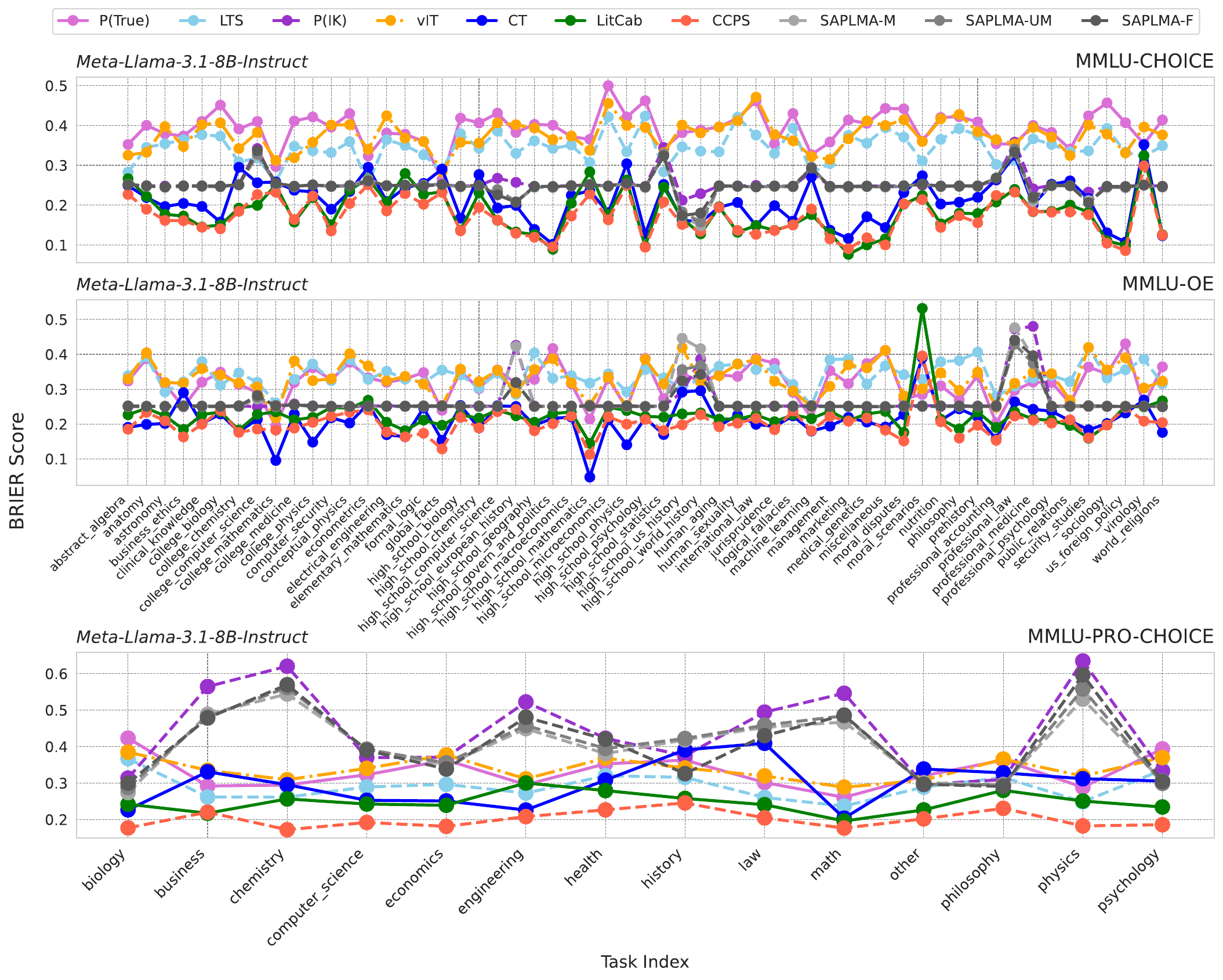}
  \caption{Brier score comparison of confidence estimation methods on \texttt{Meta-Llama-3.1-8B-Instruct} across different tasks of MMLU variants.}
  \label{fig:llama_brier}
\end{figure*}

\begin{figure*}[ht]
  \centering
  \includegraphics[width=\textwidth]{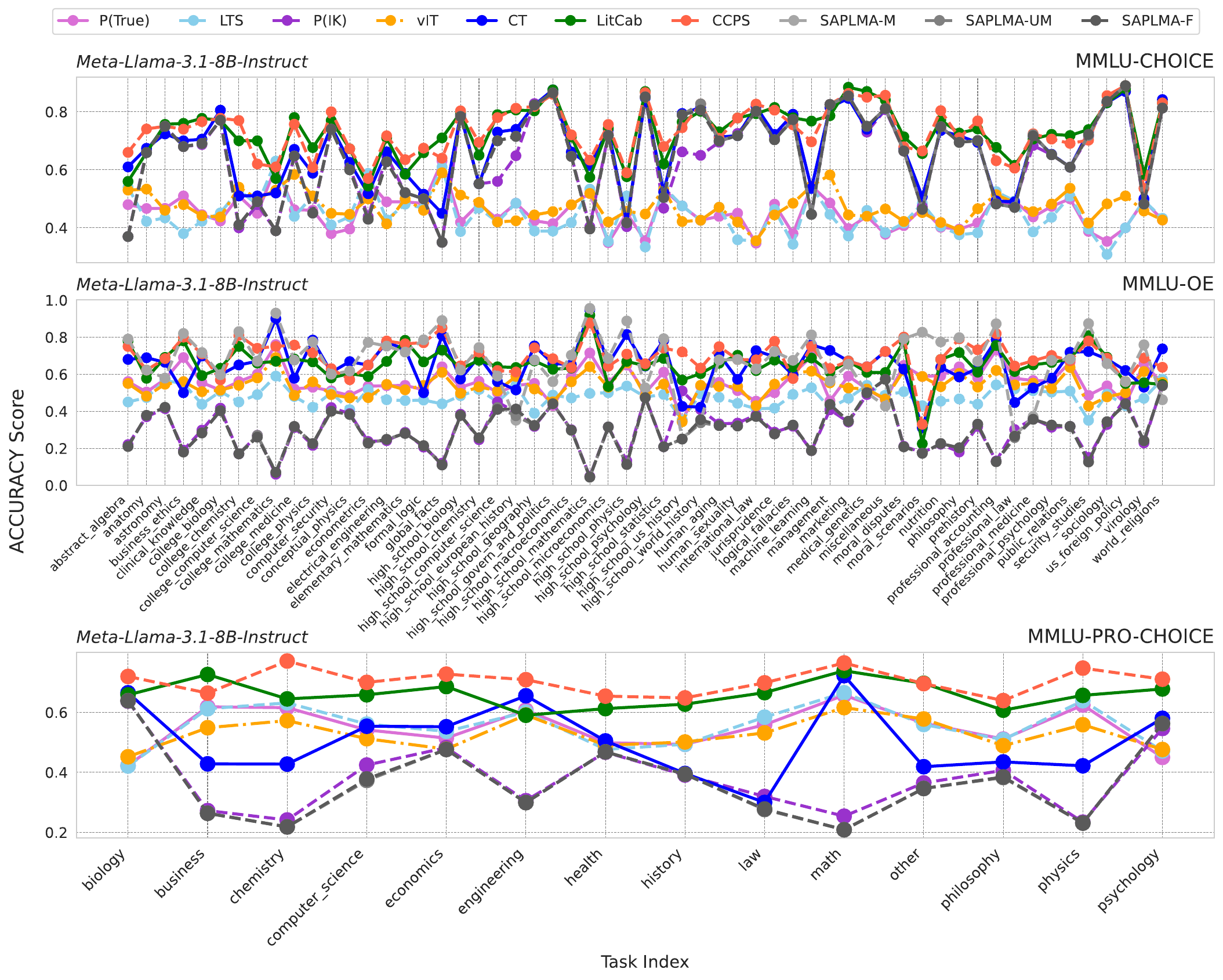}
  \caption{Accuracy (ACC) comparison of confidence estimation methods on \texttt{Meta-Llama-3.1-8B-Instruct} across different tasks of MMLU variants.}
  \label{fig:llama_acc}
\end{figure*}

\begin{figure*}[ht]
  \centering
  \includegraphics[width=\textwidth]{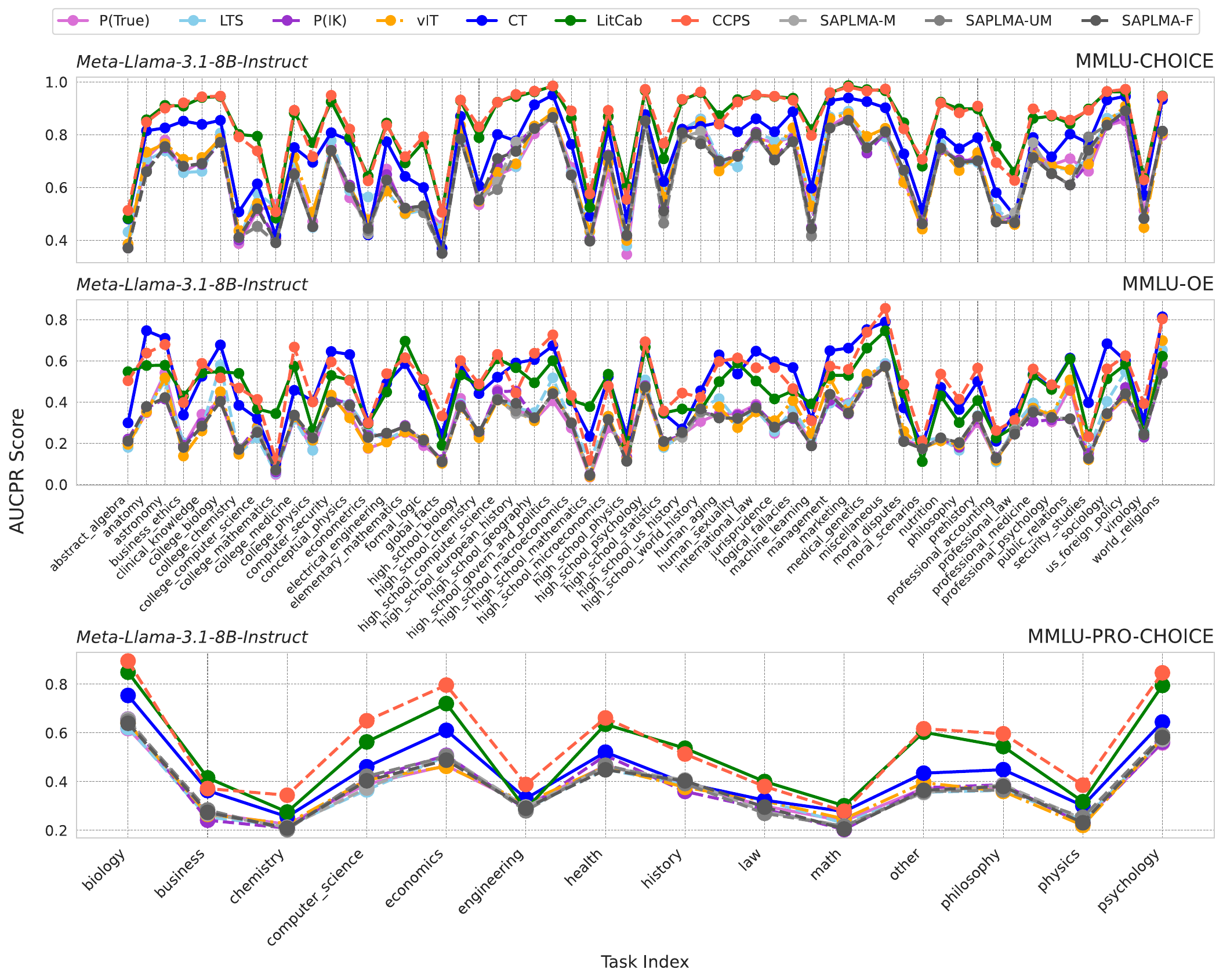}
  \caption{AUCPR comparison of confidence estimation methods on \texttt{Meta-Llama-3.1-8B-Instruct} across different tasks of MMLU variants.}
  \label{fig:llama_aucpr}
\end{figure*}

\begin{figure*}[ht]
  \centering
  \includegraphics[width=\textwidth]{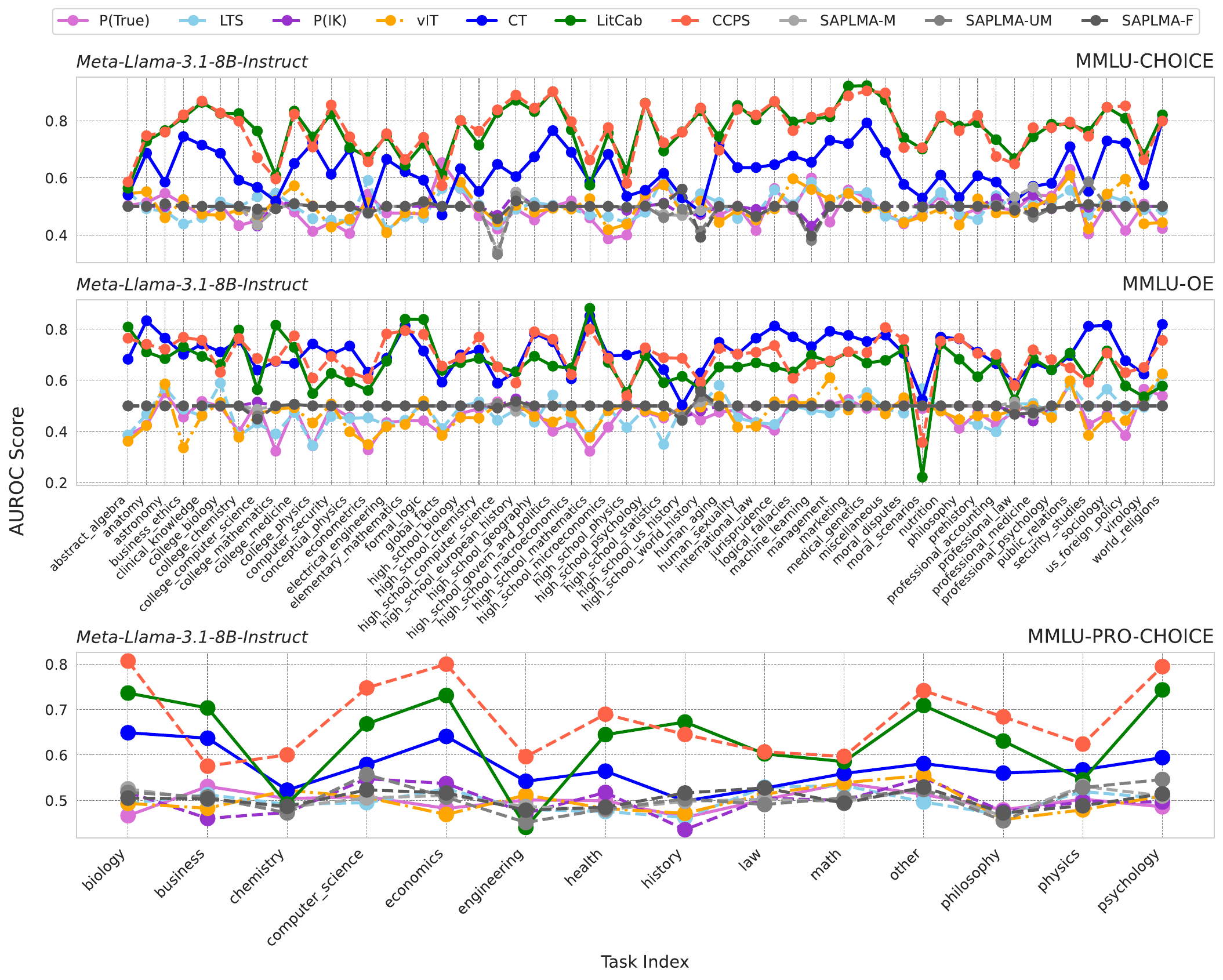}
  \caption{AUROC comparison of confidence estimation methods on \texttt{Meta-Llama-3.1-8B-Instruct} across different tasks of MMLU variants.}
  \label{fig:llama_auroc}
\end{figure*}

\begin{figure*}[ht]
  \centering
  \includegraphics[width=\textwidth]{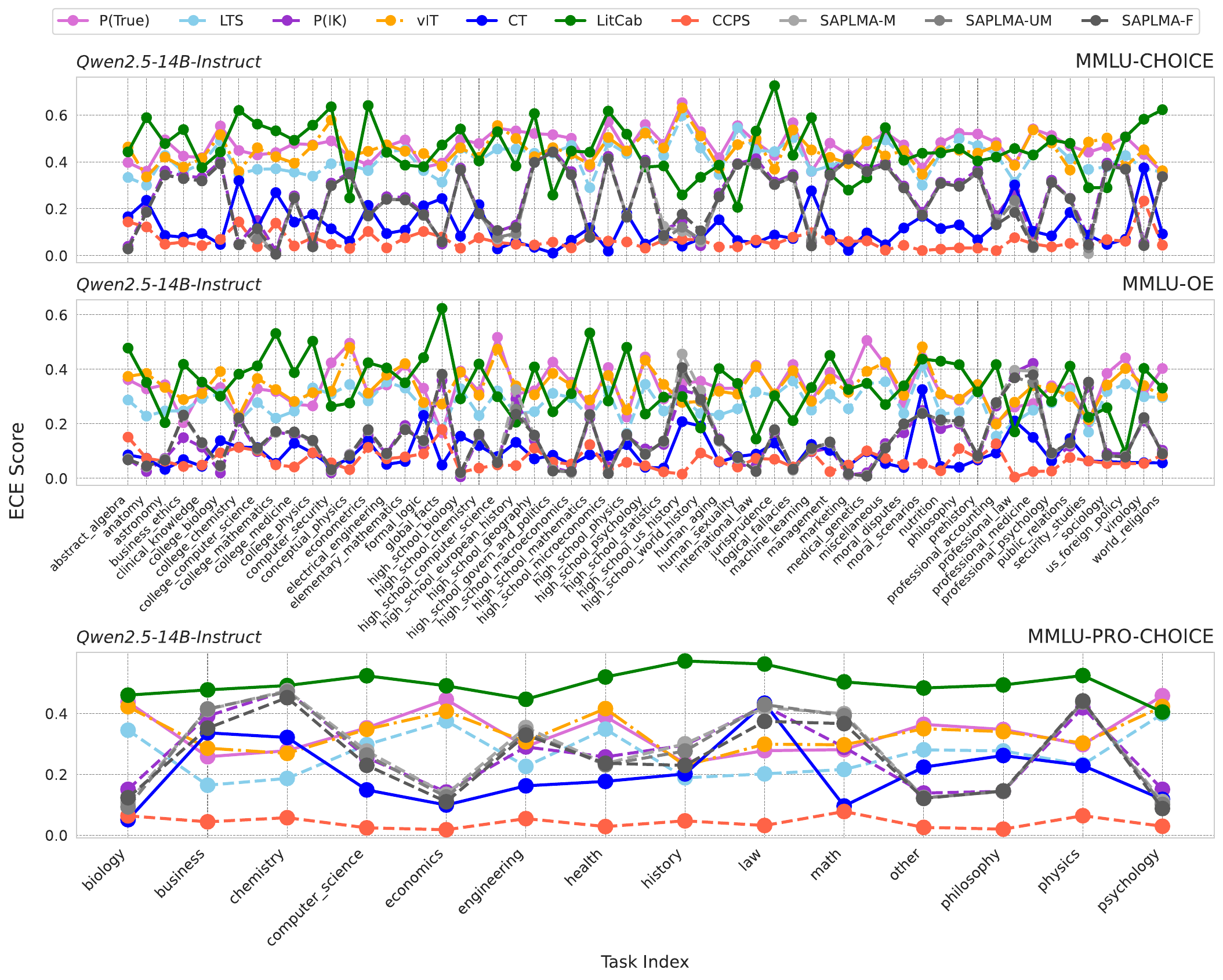}
  \caption{ECE comparison of confidence estimation methods on \texttt{Qwen2.5-14B-Instruct} across different tasks of MMLU variants.}
  \label{fig:qwen14_ece}
\end{figure*}

\begin{figure*}[ht]
  \centering
  \includegraphics[width=\textwidth]{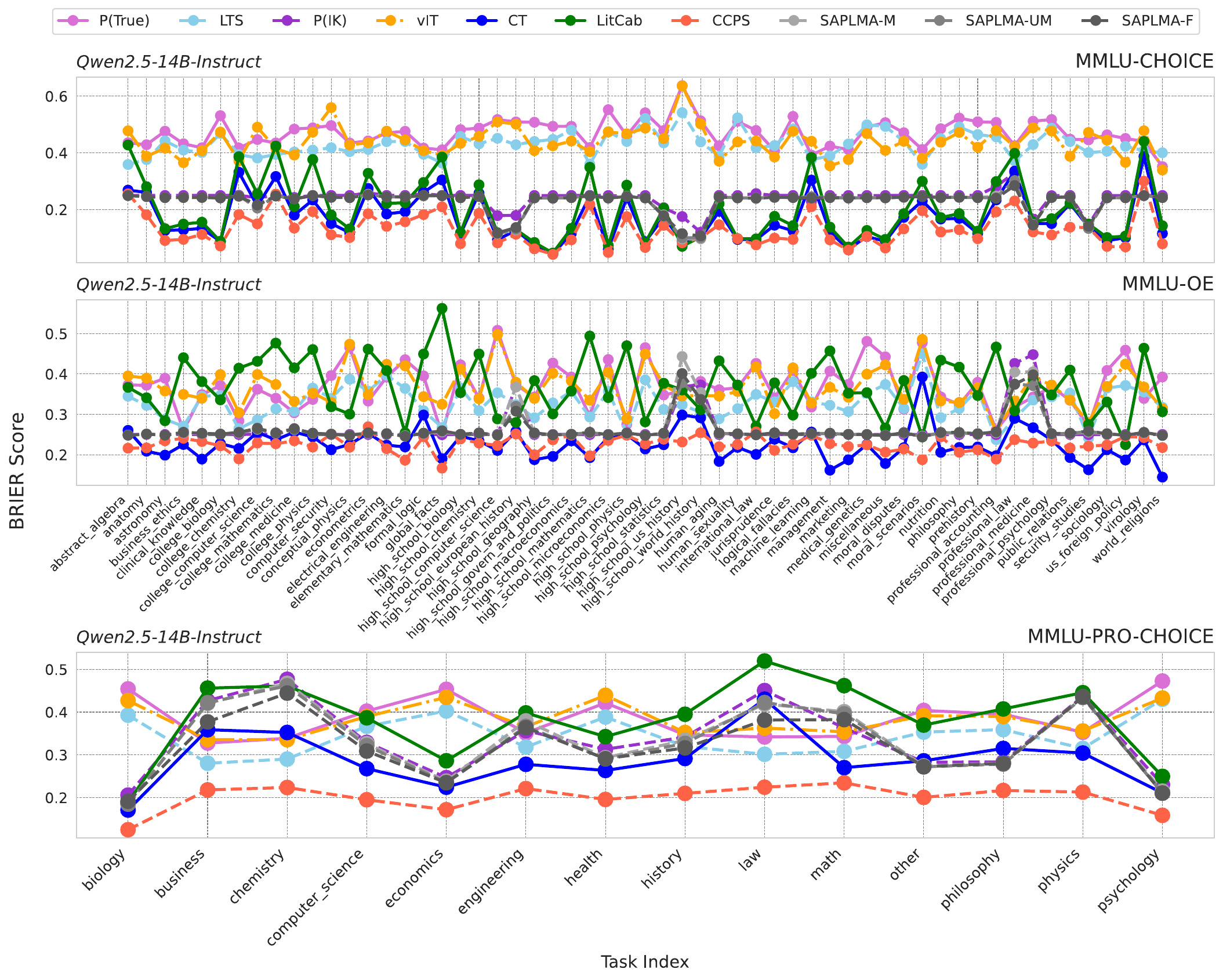}
  \caption{Brier score comparison of confidence estimation methods on \texttt{Qwen2.5-14B-Instruct} across different tasks of MMLU variants.}
  \label{fig:qwen14_brier}
\end{figure*}

\begin{figure*}[ht]
  \centering
  \includegraphics[width=\textwidth]{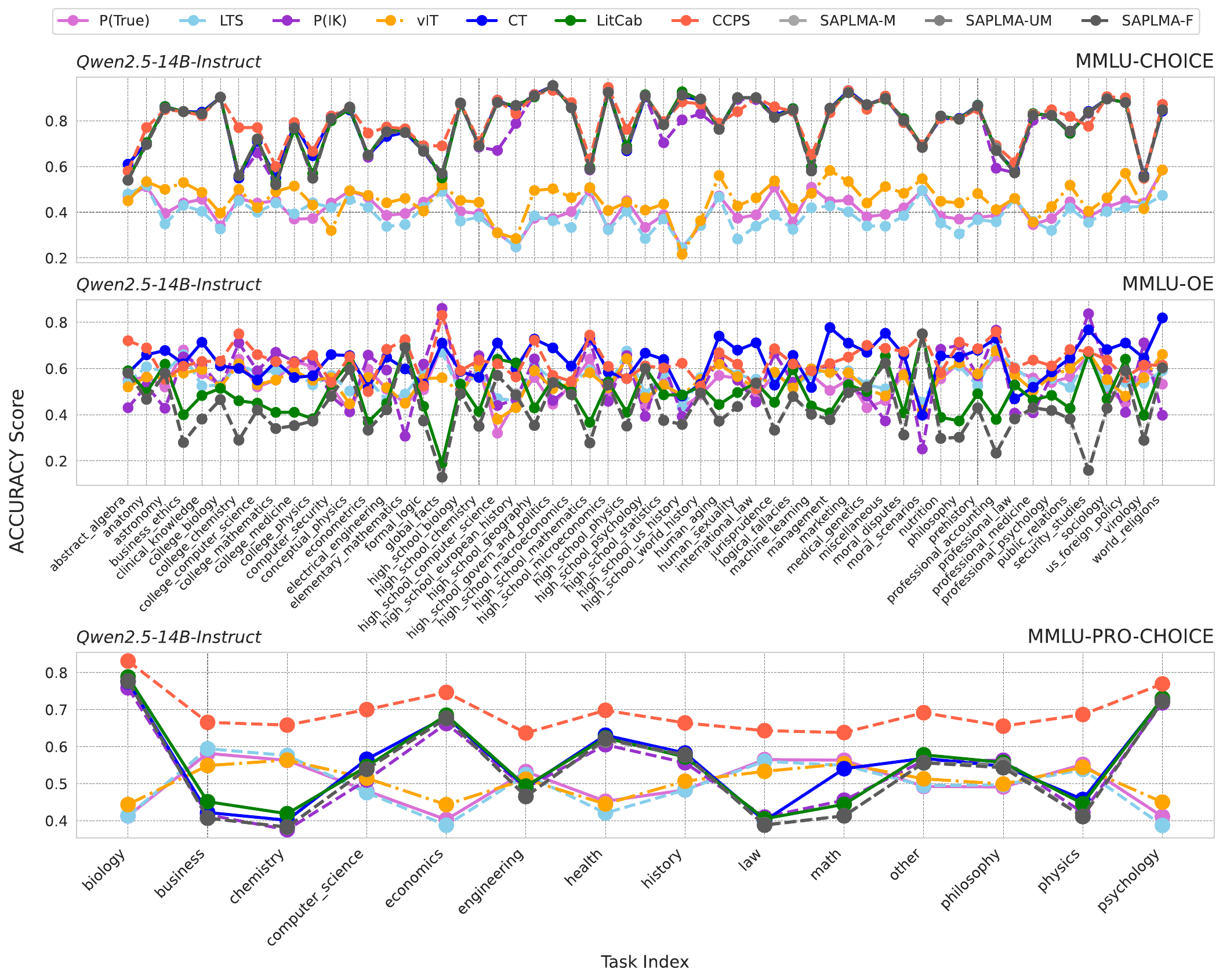}
  \caption{Accuracy (ACC) comparison of confidence estimation methods on \texttt{Qwen2.5-14B-Instruct} across different tasks of MMLU variants.}
  \label{fig:qwen14_acc}
\end{figure*}

\begin{figure*}[ht]
  \centering
  \includegraphics[width=\textwidth]{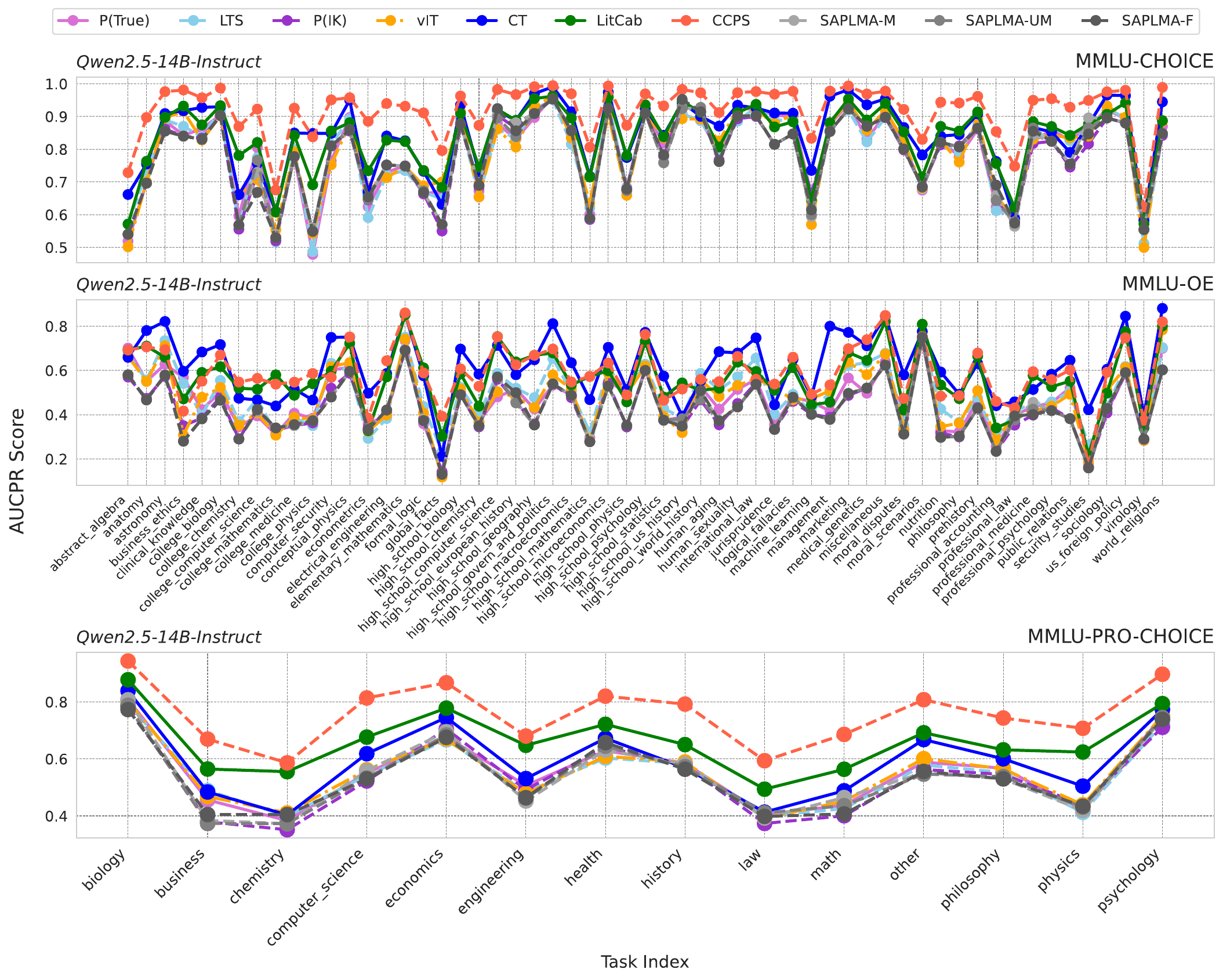}
  \caption{AUCPR comparison of confidence estimation methods on \texttt{Qwen2.5-14B-Instruct} across different tasks of MMLU variants.}
  \label{fig:qwen14_aucpr}
\end{figure*}

\begin{figure*}[ht]
  \centering
  \includegraphics[width=\textwidth]{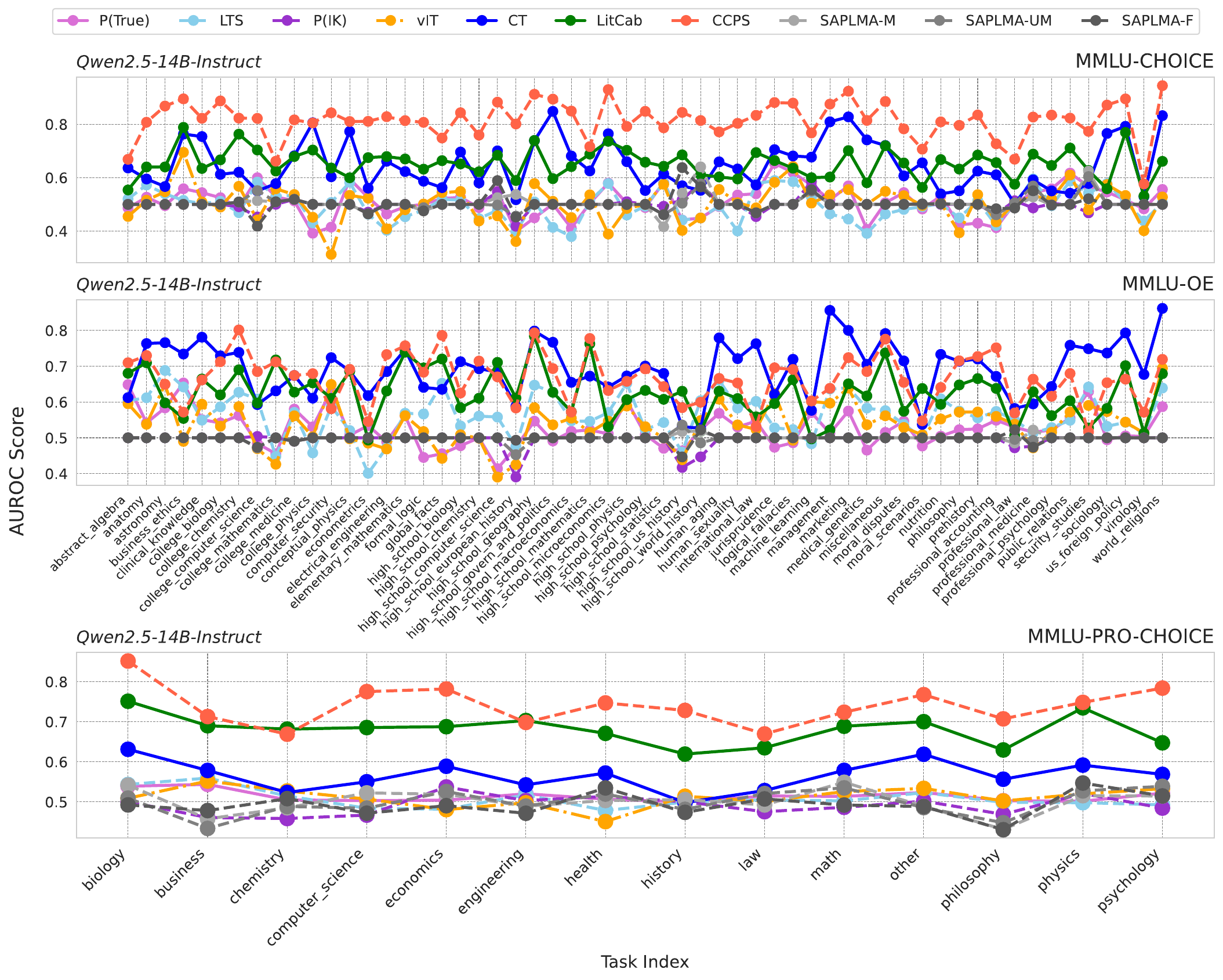}
  \caption{AUROC comparison of confidence estimation methods on \texttt{Qwen2.5-14B-Instruct} across different tasks of MMLU variants.}
  \label{fig:qwen14_auroc}
\end{figure*}

\begin{figure*}[ht]
  \centering
  \includegraphics[width=\textwidth]{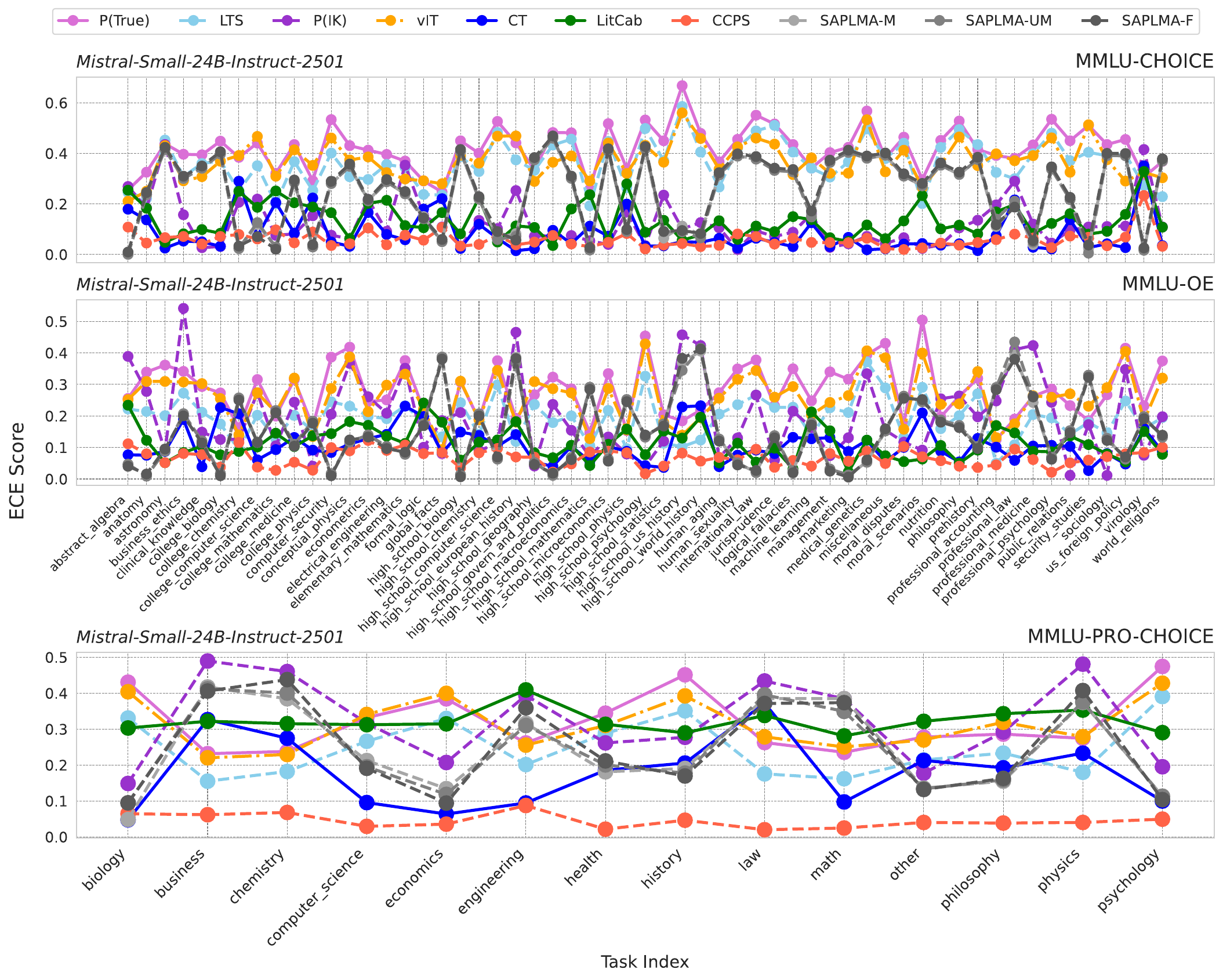}
  \caption{ECE comparison of confidence estimation methods on \texttt{Mistral-Small-24B-Instruct-2501} across different tasks of MMLU variants.}
  \label{fig:mistral_ece}
\end{figure*}

\begin{figure*}[ht]
  \centering
  \includegraphics[width=\textwidth]{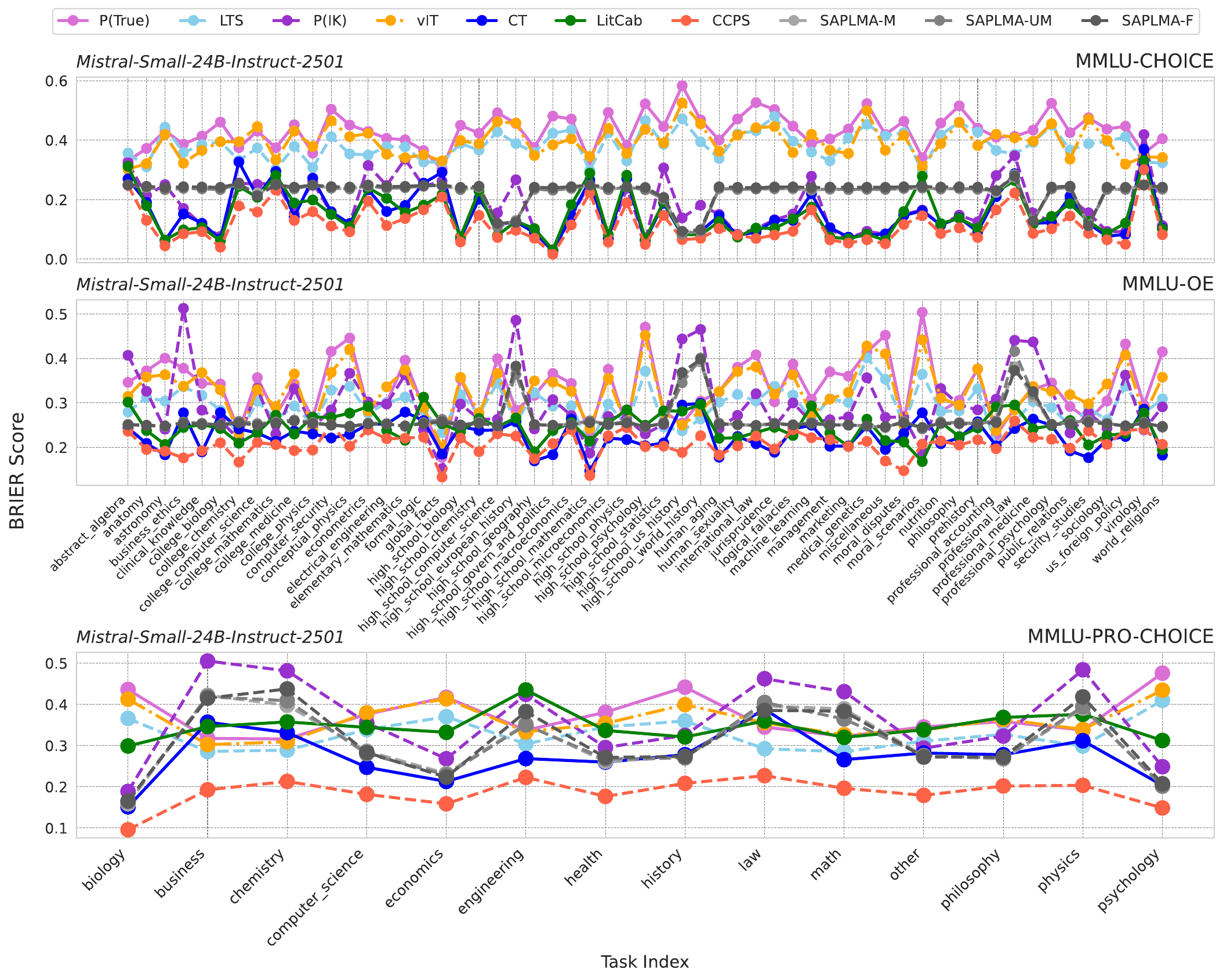}
  \caption{Brier score comparison of confidence estimation methods on \texttt{Mistral-Small-24B-Instruct-2501} across different tasks of MMLU variants.}
  \label{fig:mistral_brier}
\end{figure*}

\begin{figure*}[ht]
  \centering
  \includegraphics[width=\textwidth]{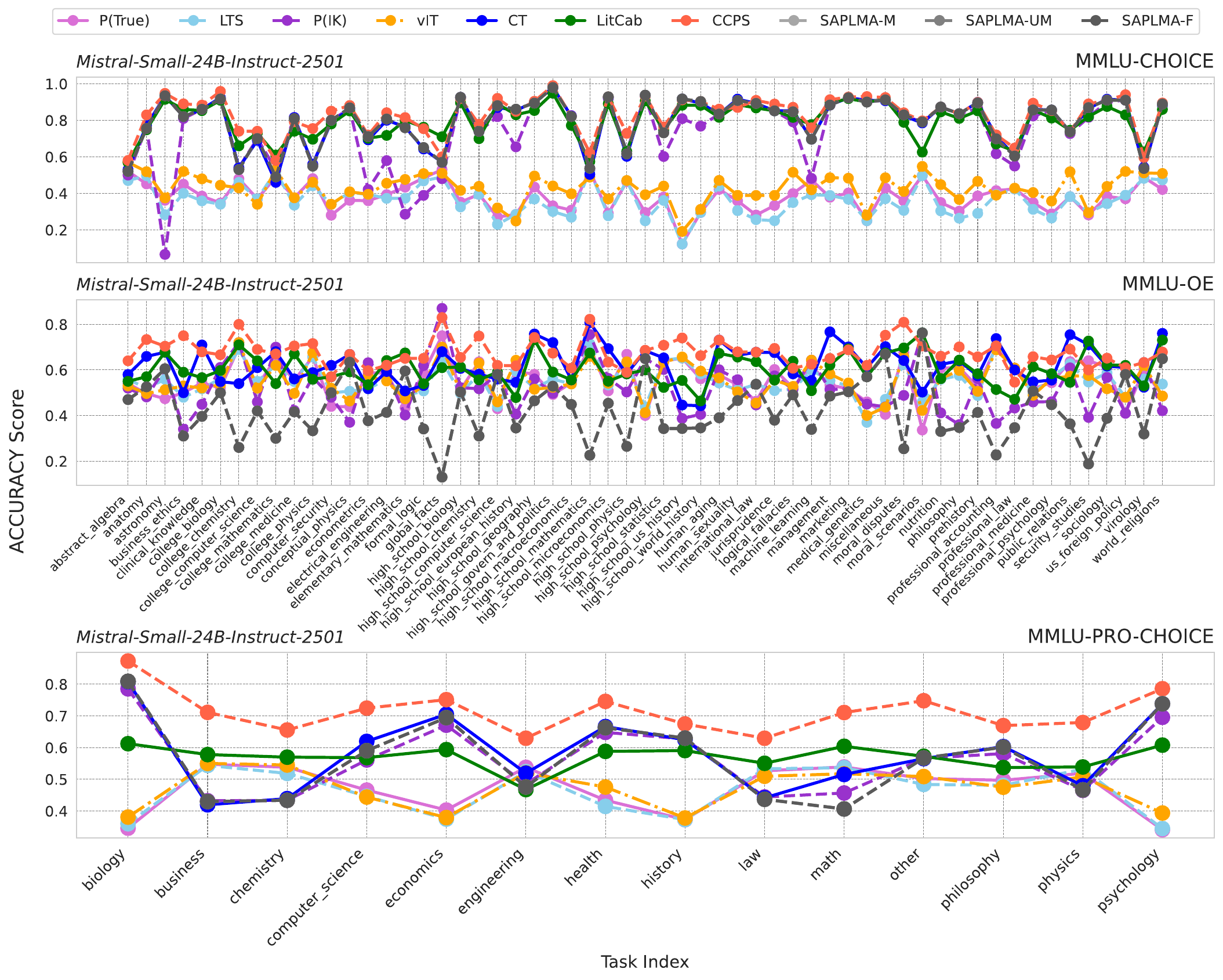}
  \caption{Accuracy (ACC) comparison of confidence estimation methods on \texttt{Mistral-Small-24B-Instruct-2501} across different tasks of MMLU variants.}
  \label{fig:mistral_acc}
\end{figure*}

\begin{figure*}[ht]
  \centering
  \includegraphics[width=\textwidth]{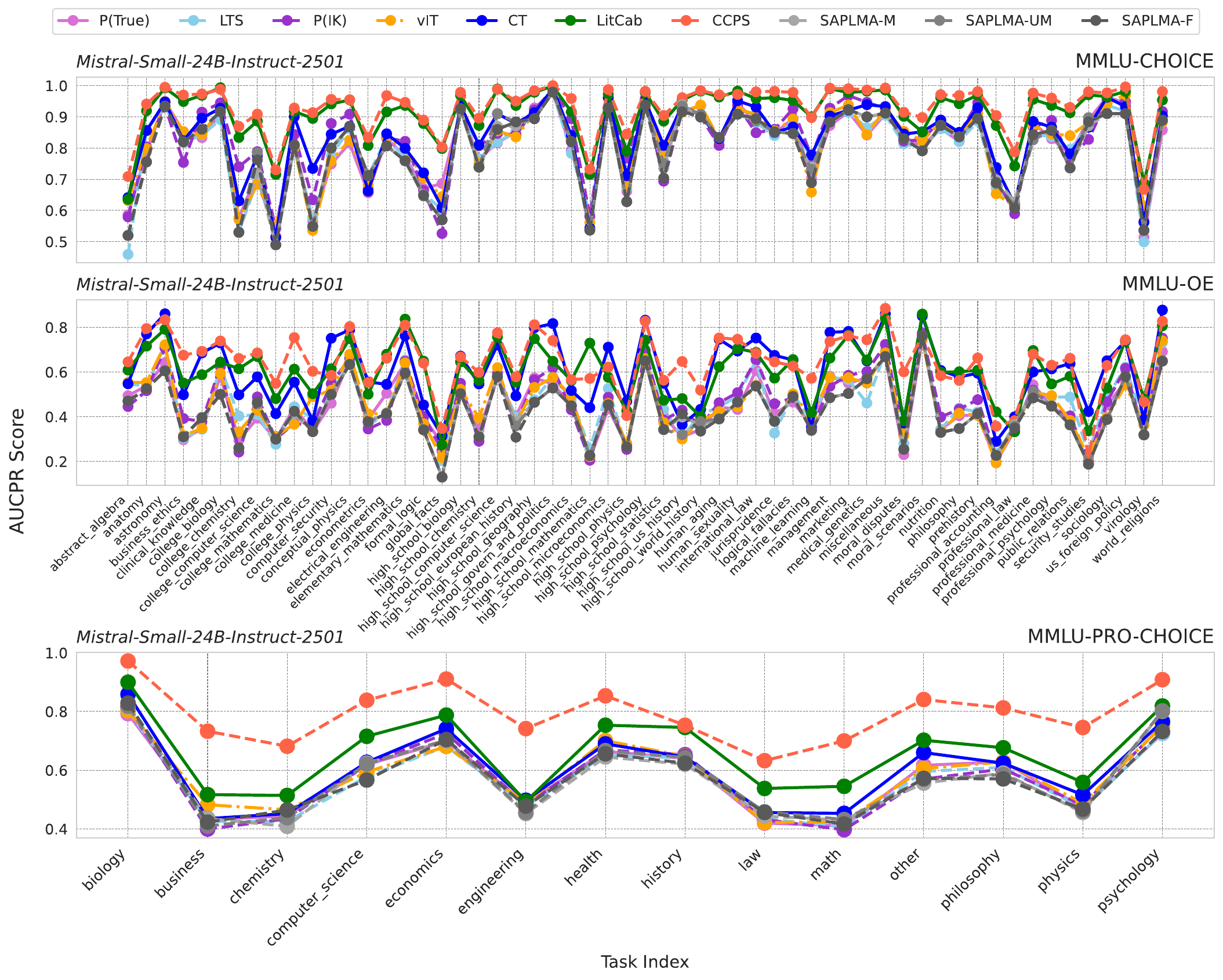}
  \caption{AUCPR comparison of confidence estimation methods on \texttt{Mistral-Small-24B-Instruct-2501} across different tasks of MMLU variants.}
  \label{fig:mistral_aucpr}
\end{figure*}

\begin{figure*}[ht]
  \centering
  \includegraphics[width=\textwidth]{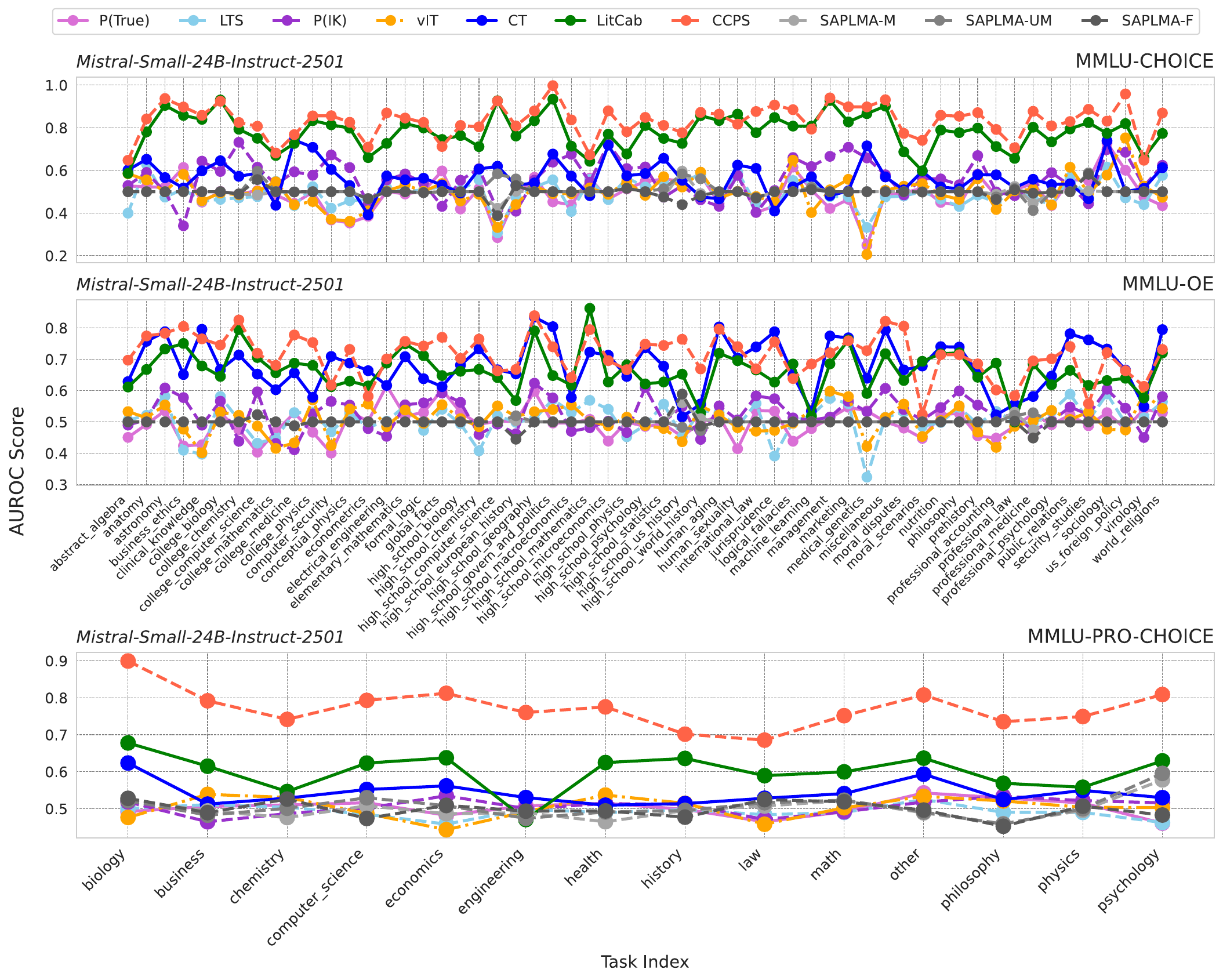}
  \caption{AUROC comparison of confidence estimation methods on \texttt{Mistral-Small-24B-Instruct-2501} across different tasks of MMLU variants.}
  \label{fig:mistral_auroc}
\end{figure*}

\begin{figure*}[ht]
  \centering
  \includegraphics[width=\textwidth]{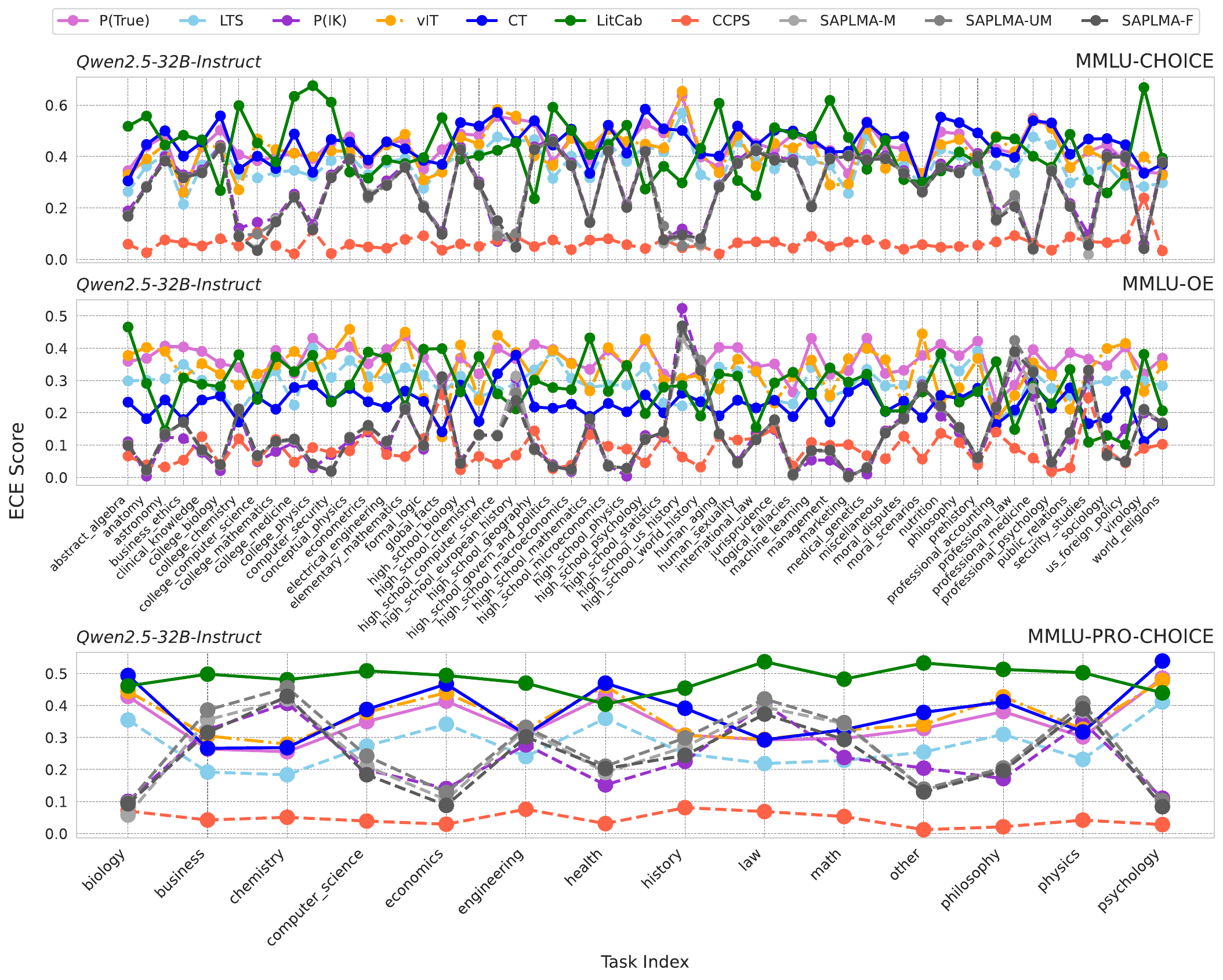}
  \caption{ECE comparison of confidence estimation methods on \texttt{Qwen2.5-32B-Instruct} across different tasks of MMLU variants.}
  \label{fig:qwen32_ece}
\end{figure*}

\begin{figure*}[ht]
  \centering
  \includegraphics[width=\textwidth]{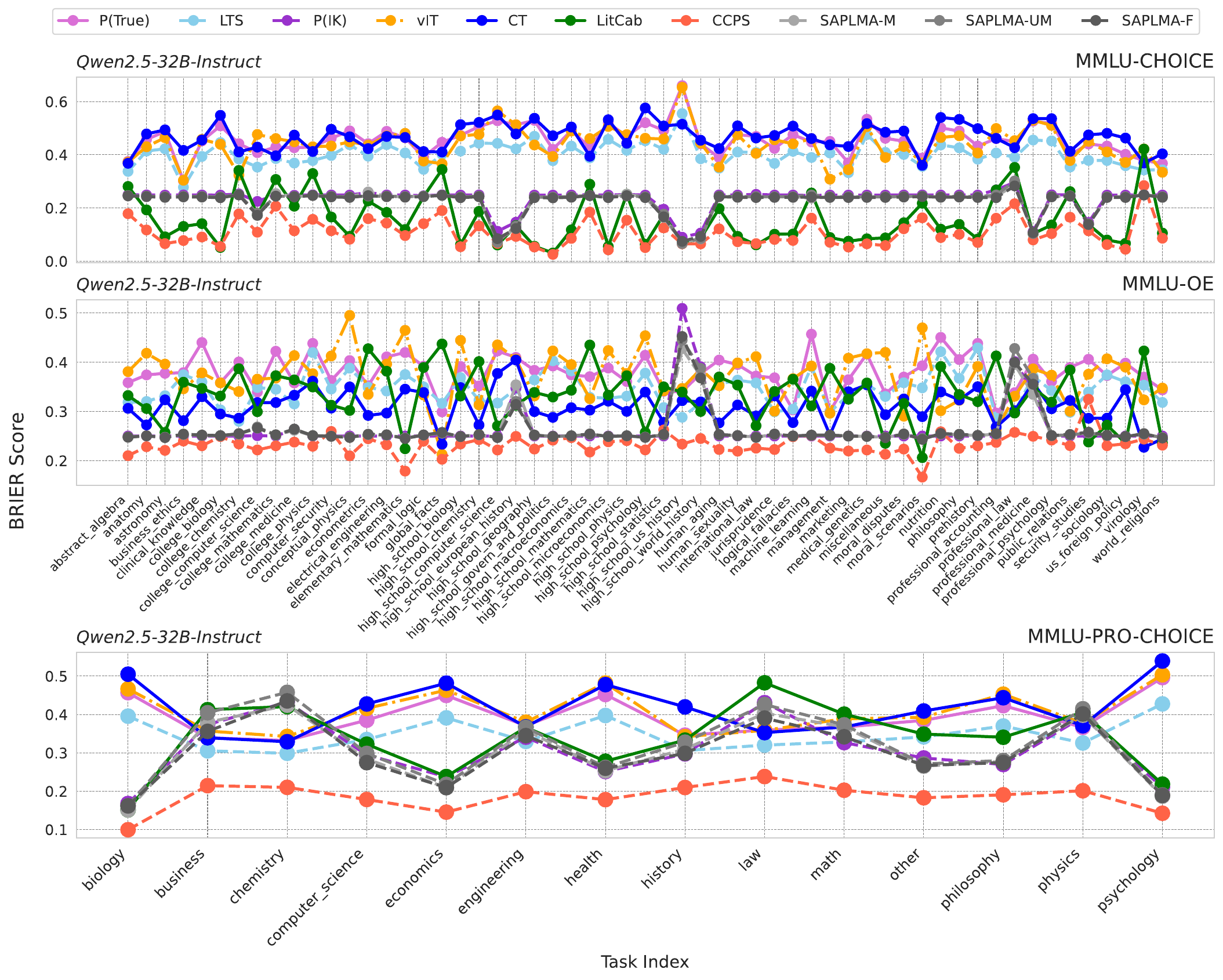}
  \caption{Brier score comparison of confidence estimation methods on \texttt{Qwen2.5-32B-Instruct} across different tasks of MMLU variants.}
  \label{fig:qwen32_brier}
\end{figure*}

\begin{figure*}[ht]
  \centering
  \includegraphics[width=\textwidth]{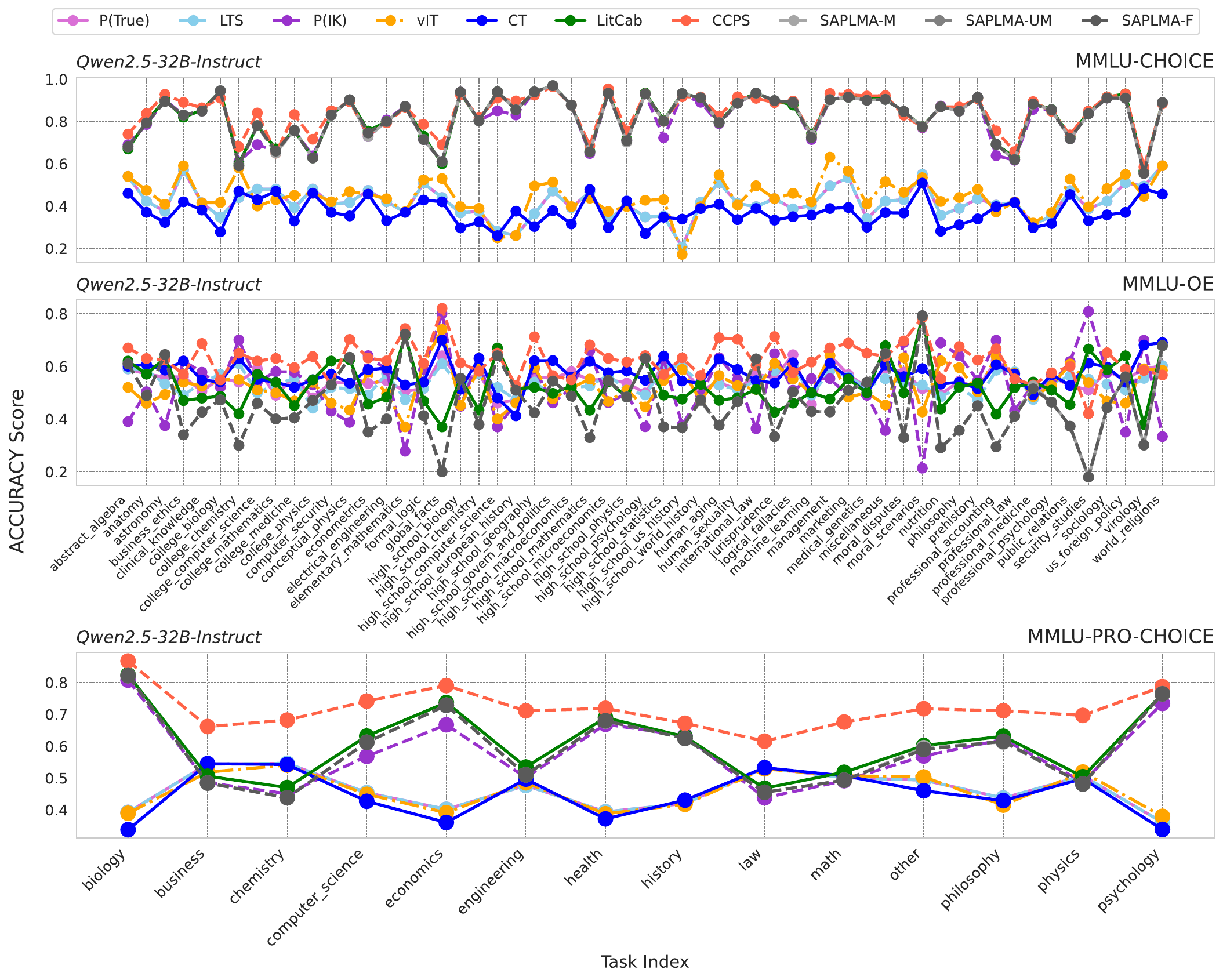}
  \caption{Accuracy (ACC) comparison of confidence estimation methods on \texttt{Qwen2.5-32B-Instruct} across different tasks of MMLU variants.}
  \label{fig:qwen32_acc}
\end{figure*}

\begin{figure*}[ht]
  \centering
  \includegraphics[width=\textwidth]{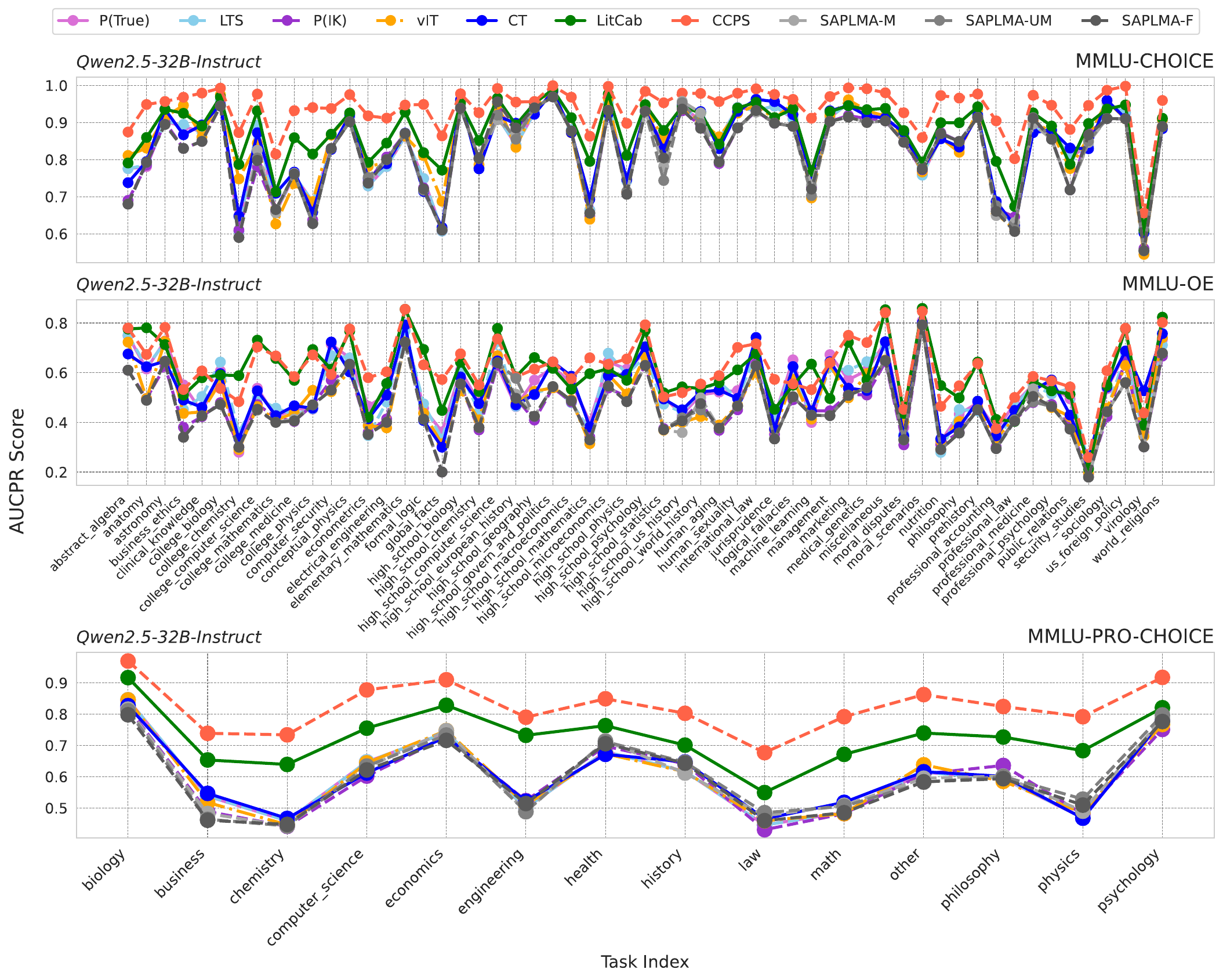}
  \caption{AUCPR comparison of confidence estimation methods on \texttt{Qwen2.5-32B-Instruct} across different tasks of MMLU variants.}
  \label{fig:qwen32_aucpr}
\end{figure*}

\begin{figure*}[ht]
  \centering
  \includegraphics[width=\textwidth]{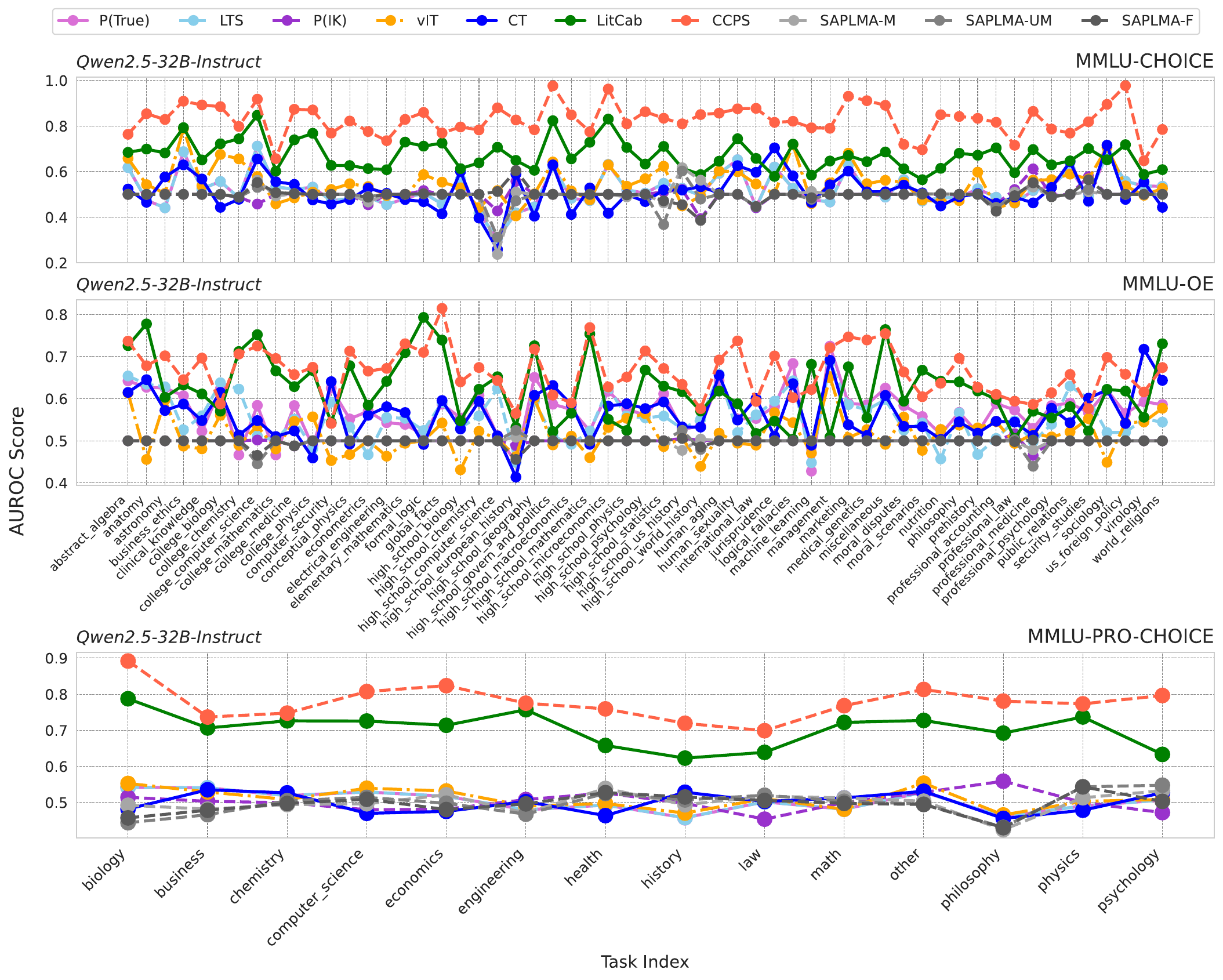}
  \caption{AUROC comparison of confidence estimation methods on \texttt{Qwen2.5-32B-Instruct} across different tasks of MMLU variants.}
  \label{fig:qwen32_auroc}
\end{figure*}

For an in-depth understanding of performance at a finer granularity, this section provides per-task results. Figures \ref{fig:llama_ece} through \ref{fig:qwen32_auroc} illustrate the comparative performance of all methods on every individual task within the MMLU datasets for each of the four base LLMs, across all evaluation metrics (ECE, Brier score, ACC, AUCPR, and AUROC).

\section{Ablation Studies}
\label{app:ablation_studies}

\subsection{Token Importance in OE Models}
\label{app:token_importance}
To address the concern that using features from all tokens in an OE generation might be excessive, we conducted an ablation study to determine whether our convolutional architecture learns to prioritize semantically meaningful tokens over less informative "filler" words.

For this analysis, we used our final trained OE CCPS models. We performed a \textit{Token Masking Impact Analysis}, which involved the following steps:
\begin{enumerate}
    \item For each sample in our test set, we first obtained the baseline confidence score from the full, unmasked sequence of token features.
    \item We then systematically masked one token at a time by zeroing out its entire feature vector and re-calculated the confidence score with the masked sequence.
    \item The "impact" of each token was measured as the absolute difference between the baseline confidence and the confidence score from the sequence where that token was masked.
    \item Finally, using Part-of-Speech (POS) tagging, we categorized each token as either a \textit{"Content" word} (e.g., nouns, verbs, adjectives, adverbs) or a \textit{"Function" word} (e.g., determiners, prepositions, pronouns) and compared the average impact scores for each category.
\end{enumerate}

The results, summarized in Table~\ref{tab:impact_content_function}, confirm that our convolutional architecture effectively learns to prioritize semantically meaningful tokens. Across all four LLMs, masking Content Words had a statistically significant and substantially larger impact on the final confidence score than masking Function Words. For example, for the \texttt{Qwen2.5-14B} and \texttt{Qwen2.5-32B} models, the impact of content words was over 75\% greater than that of function words. This analysis demonstrates that the model's sensitivity is not uniform across all tokens; the convolutional architecture effectively learns to place greater weight on semantically rich words while attenuating the influence of less informative ones, thus validating our approach to using the full feature sequence.

\begin{table*}[t]
\centering
\small
\begin{tabular}{lccc}
\toprule
\textbf{Base LLM} & \textbf{Mean Impact of Content} & \textbf{Mean Impact of Function} & \textbf{Content vs. Function Impact Lift} \\
\midrule
Meta-Llama-3.1-8B & 0.0182 & 0.0125 & +45.6\% \\
Qwen2.5-14B       & 0.0160 & 0.0091 & +75.8\% \\
Mistral-Small-24B & 0.0385 & 0.0258 & +49.2\% \\
Qwen2.5-32B       & 0.0156 & 0.0088 & +77.2\% \\
\bottomrule
\end{tabular}
\caption{Impact of masking content vs. function words on the final confidence score. ``Impact Lift'' shows the percentage increase in impact when masking a content word compared to a function word.}
\label{tab:impact_content_function}
\end{table*}

\subsection{Comparison with Self-Consistency Baseline}
\label{app:self_consistency_comparison}
In response to reviewer feedback, and to provide a more comprehensive comparison against methods that rely on output sampling, we implemented and evaluated the Self-Consistency (SC) baseline~\cite{selfconsistency}. The SC method estimates confidence by generating multiple answer samples for a given question and using the consensus or frequency of the most common answer as the confidence score. We tested the SC method using our full experimental setup across all MMLU variants.

The results of this comparison are presented in Table~\ref{tab:ccps_sc_unified}. The findings show that while SC is a competitive baseline, CCPS consistently outperforms it, particularly in calibration metrics. As shown, CCPS achieves substantially lower ECE and Brier scores across all models and datasets, indicating significantly better calibration. Furthermore, CCPS generally demonstrates superior discriminative power, leading in AUCPR and AUROC in nearly all cases. These results further validate the effectiveness of our internal stability probing approach compared to methods based on external output consistency.

\begin{table*}[h!]
\centering
\caption{Unified comparison of CCPS vs. Self-Consistency (SC) across three datasets: \textbf{MMLU-CHOICE}, \textbf{MMLU-PRO-CHOICE}, and \textbf{MMLU-OE}. Values are percentages. Best results per model are bolded.}
\label{tab:ccps_sc_unified}
\small
\resizebox{\textwidth}{!}{%
\begin{tabular}{lllcccccc}
\toprule
\textbf{Dataset} & \textbf{Model} & \textbf{Method} & \textbf{ECE $\downarrow$} & \textbf{Brier $\downarrow$} & \textbf{ACC $\uparrow$} & \textbf{AUCPR $\uparrow$} & \textbf{AUROC $\uparrow$} \\
\midrule
\multirow{8}{*}{\textbf{MMLU-CHOICE}} 
 & \multirow{2}{*}{\texttt{Meta-Llama-3.1-8B}} 
   & SC    & 23.0 & 22.6 & 69.1 & 77.0 & 71.5 \\
 &   & \textbf{CCPS} & \textbf{6.5} & \textbf{17.1} & \textbf{73.4} & \textbf{84.1} & \textbf{77.1} \\
\cmidrule{2-8}
 & \multirow{2}{*}{\texttt{Qwen2.5-14B}} 
   & SC    & 46.1 & 20.8 & 78.1 & 79.9 & 55.5 \\
 &   & \textbf{CCPS} & \textbf{6.3} & \textbf{13.1} & \textbf{80.2} & \textbf{92.1} & \textbf{81.6} \\
\cmidrule{2-8}
 & \multirow{2}{*}{\texttt{Mistral-Small-24B}} 
   & SC    & 22.3 & 14.7 & 81.2 & 88.0 & 74.2 \\
 &   & \textbf{CCPS} & \textbf{5.8} & \textbf{11.5} & \textbf{83.0} & \textbf{93.1} & \textbf{83.3} \\
\cmidrule{2-8}
 & \multirow{2}{*}{\texttt{Qwen2.5-32B}} 
   & SC    & 26.3 & 16.4 & 82.6 & 84.5 & 56.8 \\
 &   & \textbf{CCPS} & \textbf{6.3} & \textbf{10.8} & \textbf{84.1} & \textbf{94.1} & \textbf{82.8} \\
\midrule
\multirow{8}{*}{\textbf{MMLU-PRO-CHOICE}} 
 & \multirow{2}{*}{\texttt{Meta-Llama-3.1-8B}} 
   & SC    & 19.3 & 29.8 & 60.3 & 51.9 & \textbf{68.7} \\
 &   & \textbf{CCPS} & \textbf{4.5} & \textbf{20.0} & \textbf{70.4} & \textbf{55.2} & 67.9 \\
\cmidrule{2-8}
 & \multirow{2}{*}{\texttt{Qwen2.5-14B}} 
   & SC    & 47.1 & 41.1 & 54.8 & 58.5 & 58.7 \\
 &   & \textbf{CCPS} & \textbf{4.2} & \textbf{20.1} & \textbf{69.2} & \textbf{75.8} & \textbf{74.0} \\
\cmidrule{2-8}
 & \multirow{2}{*}{\texttt{Mistral-Small-24B}} 
   & SC    & 20.2 & 25.0 & 67.2 & 72.7 & 73.0 \\
 &   & \textbf{CCPS} & \textbf{4.5} & \textbf{18.6} & \textbf{71.3} & \textbf{79.5} & \textbf{77.2} \\
\cmidrule{2-8}
 & \multirow{2}{*}{\texttt{Qwen2.5-32B}} 
   & SC    & 45.4 & 36.0 & 60.5 & 64.4 & 59.5 \\
 &   & \textbf{CCPS} & \textbf{4.6} & \textbf{18.5} & \textbf{71.8} & \textbf{82.4} & \textbf{77.8} \\
\midrule
\multirow{8}{*}{\textbf{MMLU-OE}} 
 & \multirow{2}{*}{\texttt{Meta-Llama-3.1-8B}} 
   & SC    & 11.2 & 21.2 & \textbf{71.2} & 46.6 & 67.0 \\
 &   & \textbf{CCPS} & \textbf{8.0} & \textbf{20.2} & 69.5 & \textbf{49.4} & \textbf{69.3} \\
\cmidrule{2-8}
 & \multirow{2}{*}{\texttt{Qwen2.5-14B}} 
   & SC    & 22.5 & 33.0 & 56.6 & 51.7 & 59.8 \\
 &   & \textbf{CCPS} & \textbf{6.7} & \textbf{22.5} & \textbf{63.6} & \textbf{59.0} & \textbf{66.6} \\
\cmidrule{2-8}
 & \multirow{2}{*}{\texttt{Mistral-Small-24B}} 
   & SC    & 14.0 & 24.1 & 65.8 & 58.9 & 67.2 \\
 &   & \textbf{CCPS} & \textbf{6.8} & \textbf{20.8} & \textbf{67.6} & \textbf{64.7} & \textbf{71.4} \\
\cmidrule{2-8}
 & \multirow{2}{*}{\texttt{Qwen2.5-32B}} 
   & SC    & 23.7 & 33.6 & 56.0 & 54.3 & 59.8 \\
 &   & \textbf{CCPS} & \textbf{8.7} & \textbf{23.3} & \textbf{62.6} & \textbf{62.0} & \textbf{66.4} \\
\bottomrule
\end{tabular}
}
\end{table*}

\subsection{Disentangling Feature Contributions}
\label{app:disentangling_feature_contributions}
To assess whether the observed performance gains of CCPS arise from its novel perturbation mechanism or merely from classification on features extracted from the LLM’s unperturbed representations, we conducted a comprehensive ablation study. This study aims to isolate and quantify the contribution of our perturbation-derived features.

While our main results demonstrate CCPS's superior performance over other established probing baselines like SAPLMA, we designed this ablation to provide a more direct, controlled comparison within our own framework. We created and evaluated three variants of the CCPS model:
\begin{itemize}
    \item \textbf{Original Only ($O$):} A CCPS classifier trained exclusively on the 12 features derived from the LLM's initial, unperturbed hidden state (e.g., \texttt{original\_entropy}, \texttt{original\_log\_prob\_actual}). This serves as our non-perturbation-based probe baseline.
    \item \textbf{Perturbation Only ($P$):} A CCPS classifier trained exclusively on the 63 features derived from the perturbation process and its effects (e.g., \texttt{epsilon\_to\_flip\_token}, \texttt{pei\_value\_token}, perturbed state statistics, and comparison features).
    \item \textbf{Full ($F$):} Our complete CCPS model, which uses all 75 features (Original + Perturbation).
\end{itemize}
For each variant, the model architecture and training process were kept identical, with only the input dimension of the first layer adjusted to match the feature set size. The results for each LLM across the three MMLU variants are presented in Table~\ref{tab:ece_auroc_unified}. The results of this ablation study clearly demonstrate that our perturbation-based features are the primary driver of CCPS's strong performance.

\begin{table*}[t!]
\centering
\small
\resizebox{0.8\textwidth}{!}{%
\begin{tabular}{lllcc}
\toprule
\textbf{Dataset} & \textbf{Model} & \textbf{Variant} & \textbf{ECE $\downarrow$} & \textbf{AUROC $\uparrow$} \\
\midrule
\multirow{12}{*}{\textbf{\texttt{MMLU-CHOICE}}} 
 & \multirow{3}{*}{\texttt{Meta-Llama-3.1-8B-Instruct}} & $O$ & 17.1 & 56.3 \\
 & & $P$ & 11.3 & 63.0 \\
 & & \textbf{$F$} & \textbf{6.5} & \textbf{77.1} \\
\cmidrule{2-5}
 & \multirow{3}{*}{\texttt{Qwen2.5-14B-Instruct}} & $O$ & 12.1 & 62.8 \\
 & & $P$ & 8.6 & 74.7 \\
 & & \textbf{$F$} & \textbf{6.4} & \textbf{81.6} \\
\cmidrule{2-5}
 & \multirow{3}{*}{\texttt{Mistral-Small-24B-Instruct-2501}} & $O$ & 14.8 & 73.2 \\
 & & $P$ & 8.1 & 76.5 \\
 & & \textbf{$F$} & \textbf{5.9} & \textbf{83.3} \\
\cmidrule{2-5}
 & \multirow{3}{*}{\texttt{Qwen2.5-32B-Instruct}} & $O$ & 15.6 & 63.0 \\
 & & $P$ & 8.9 & 77.8 \\
 & & \textbf{$F$} & \textbf{6.3} & \textbf{82.9} \\
\midrule
\multirow{12}{*}{\textbf{\texttt{MMLU-PRO-CHOICE}}} 
 & \multirow{3}{*}{\texttt{Meta-Llama-3.1-8B-Instruct}} & $O$ & 17.1 & 58.6 \\
 & & $P$ & 6.1 & 61.2 \\
 & & \textbf{$F$} & \textbf{4.5} & \textbf{67.9} \\
\cmidrule{2-5}
 & \multirow{3}{*}{\texttt{Qwen2.5-14B-Instruct}} & $O$ & 21.0 & 59.0 \\
 & & $P$ & 11.0 & 70.7 \\
 & & \textbf{$F$} & \textbf{4.2} & \textbf{74.0} \\
\cmidrule{2-5}
 & \multirow{3}{*}{\texttt{Mistral-Small-24B-Instruct-2501}} & $O$ & 18.3 & 54.5 \\
 & & $P$ & 8.9 & 63.1 \\
 & & \textbf{$F$} & \textbf{4.5} & \textbf{77.2} \\
\cmidrule{2-5}
 & \multirow{3}{*}{\texttt{Qwen2.5-32B-Instruct}} & $O$ & 23.9 & 77.7 \\
 & & $P$ & 6.5 & 77.5 \\
 & & \textbf{$F$} & \textbf{4.6} & \textbf{78.0} \\
\midrule
\multirow{12}{*}{\textbf{\texttt{MMLU-OE}}} 
 & \multirow{3}{*}{\texttt{Meta-Llama-3.1-8B-Instruct}} & $O$ & 14.6 & 48.7 \\
 & & $P$ & 10.2 & 61.4 \\
 & & \textbf{$F$} & \textbf{8.0} & \textbf{69.3} \\
\cmidrule{2-5}
 & \multirow{3}{*}{\texttt{Qwen2.5-14B-Instruct}} & $O$ & 15.2 & 60.1 \\
 & & $P$ & 9.6 & 61.0 \\
 & & \textbf{$F$} & \textbf{6.8} & \textbf{66.6} \\
\cmidrule{2-5}
 & \multirow{3}{*}{\texttt{Mistral-Small-24B-Instruct-2501}} & $O$ & 10.8 & 50.1 \\
 & & $P$ & 9.7 & 66.3 \\
 & & \textbf{$F$} & \textbf{6.8} & \textbf{71.4} \\
\cmidrule{2-5}
 & \multirow{3}{*}{\texttt{Qwen2.5-32B-Instruct}} & $O$ & 23.4 & 52.2 \\
 & & $P$ & 12.5 & 59.1 \\
 & & \textbf{$F$} & \textbf{8.7} & \textbf{66.4} \\
\bottomrule
\end{tabular}
}
\caption{Ablation results of disentangling feature contributions. Variants are Original Only ($O$), Perturbation Only ($P$), and Full ($F$). Best performance per model is bolded.}
\label{tab:ece_auroc_unified}
\end{table*}

\vspace{5pt}
\noindent\textbf{Perturbation Features are the Dominant Performance Driver.}
As shown across all datasets, the $P$ model consistently and substantially outperforms the $O$ probe. This is particularly evident in ECE, where the perturbation-derived features consistently yield much better calibration than features from the unperturbed state. For example, on \texttt{MMLU-PRO-CHOICE} with \texttt{Qwen2.5-14B}, the $P$ model cuts the ECE in half (21.0\% $\rightarrow$ 11.0\%) and boosts AUROC by over 11 percentage points (59.0\% $\rightarrow$ 70.7\%) compared to the $O$ model. This directly isolates and confirms the significant contribution of our core perturbation mechanism.

\vspace{5pt}
\noindent\textbf{The Value of Perturbation Increases with Task Difficulty.}
The performance gap between the perturbation-based models and the simple probe widens on more challenging datasets. While the $O$ probe is a reasonable baseline on standard \texttt{MMLU-CHOICE}, its performance degrades considerably on the more difficult \texttt{MMLU-PRO-CHOICE} and \texttt{MMLU-OE} tasks. In contrast, the $P$ model remains robust, showing much smaller performance degradation. For instance, with \texttt{Meta-Llama-3.1-8B}, the AUROC gap between $P$ and $O$ is 6.7 pp on \texttt{MMLU-CHOICE}, but this gap widens to 12.7 pp on the more challenging \texttt{MMLU-OE}. This strongly suggests that as task complexity increases and simple signals like initial log-probabilities become less reliable, the deeper stability signals captured by CCPS's perturbation mechanism become critical for robust confidence estimation.

\vspace{5pt}
\noindent\textbf{Features Combine Synergistically.}
Finally, the Full ($F$) model, which combines both feature sets, consistently achieves the best performance across all metrics and datasets. This indicates that the original state features, while less powerful on their own, provide complementary information that further refines the predictions made using the dominant perturbation-based features. This study provides clear evidence that the performance gains of CCPS are fundamentally driven by the novel perturbation mechanism we introduce, sharpening its distinction from prior probing methods.

\subsection{Feature Importance Analysis with SHAP}
\label{app:shap}
\begin{figure*}[ht]
  \centering
  \includegraphics[width=0.9\textwidth]{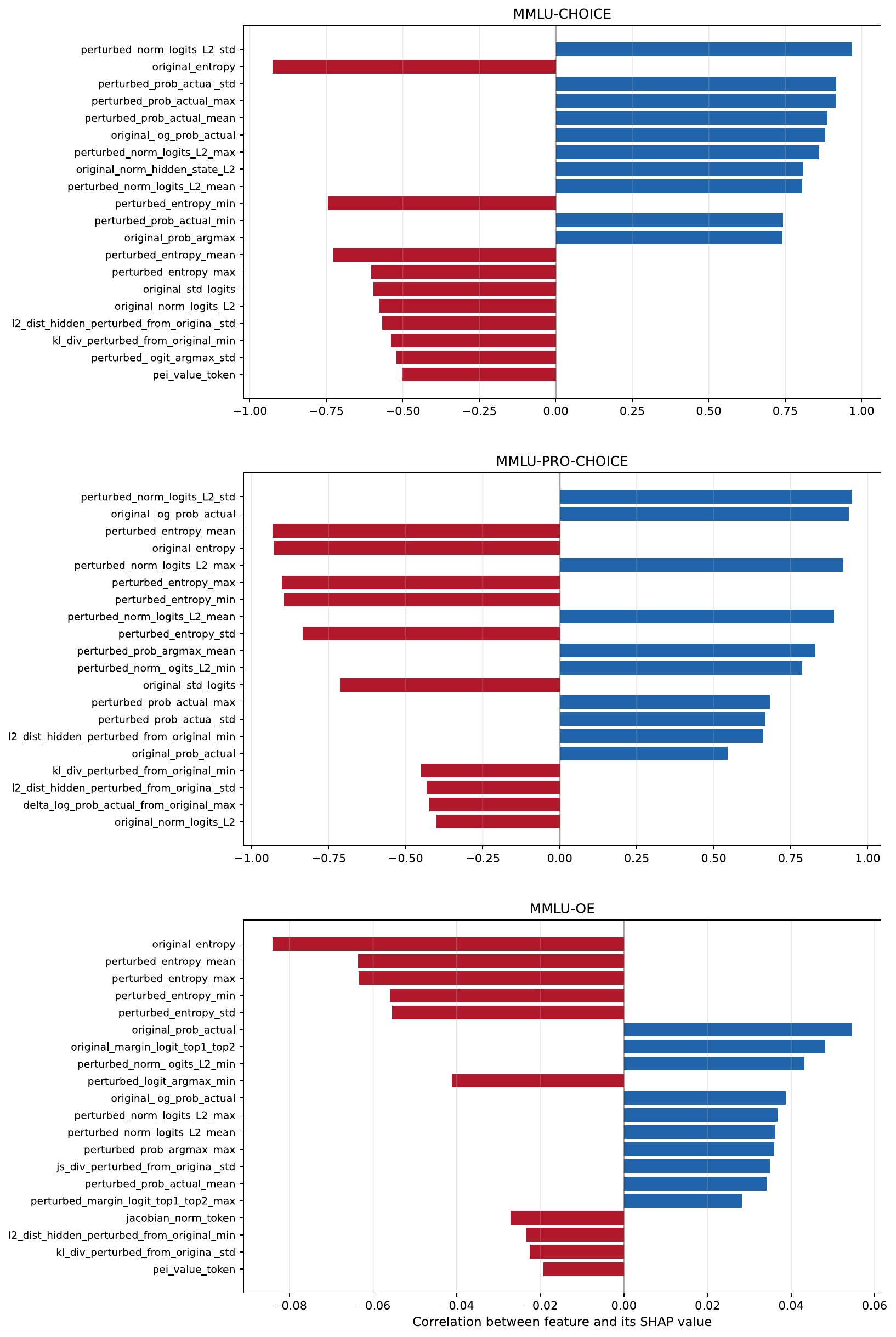}
  \caption{Correlations between feature values and SHAP scores in CCPS on \texttt{Meta-Llama-3.1-8B-Instruct} across all datasets. Blue bars denote positive correlations (higher feature values increase prediction ACC), and red bars denote negative correlations.}
  \label{fig:llama_shap}
\end{figure*}

\begin{figure*}[ht]
  \centering
  \includegraphics[width=0.9\textwidth]{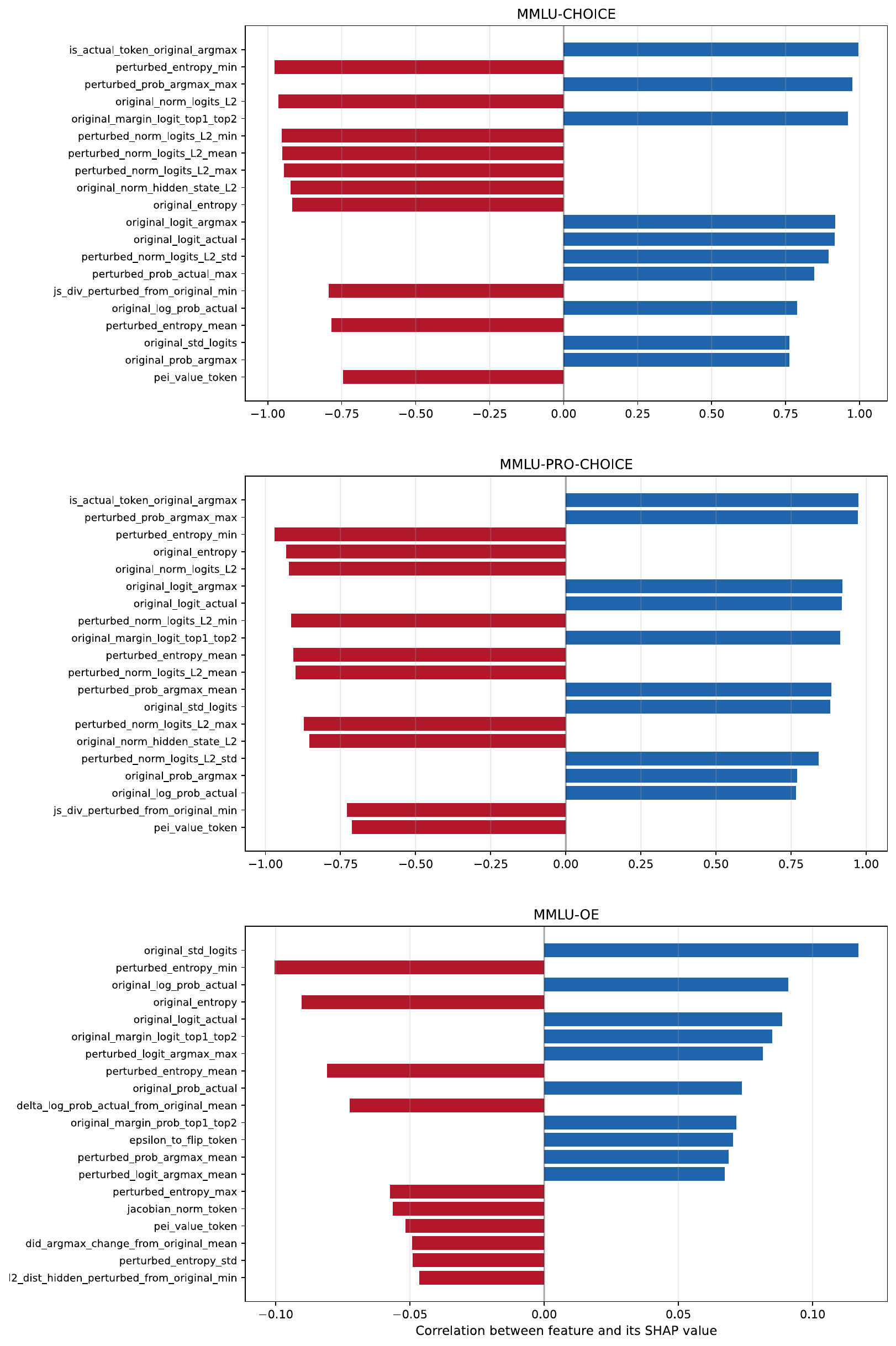}
  \caption{Correlations between feature values and SHAP scores in CCPS on \texttt{Qwen2.5-14B-Instruct} across all datasets. Blue bars denote positive correlations (higher feature values increase prediction ACC), and red bars denote negative correlations.}
  \label{fig:qwen14_shap}
\end{figure*}

\begin{figure*}[ht]
  \centering
  \includegraphics[width=0.9\textwidth]{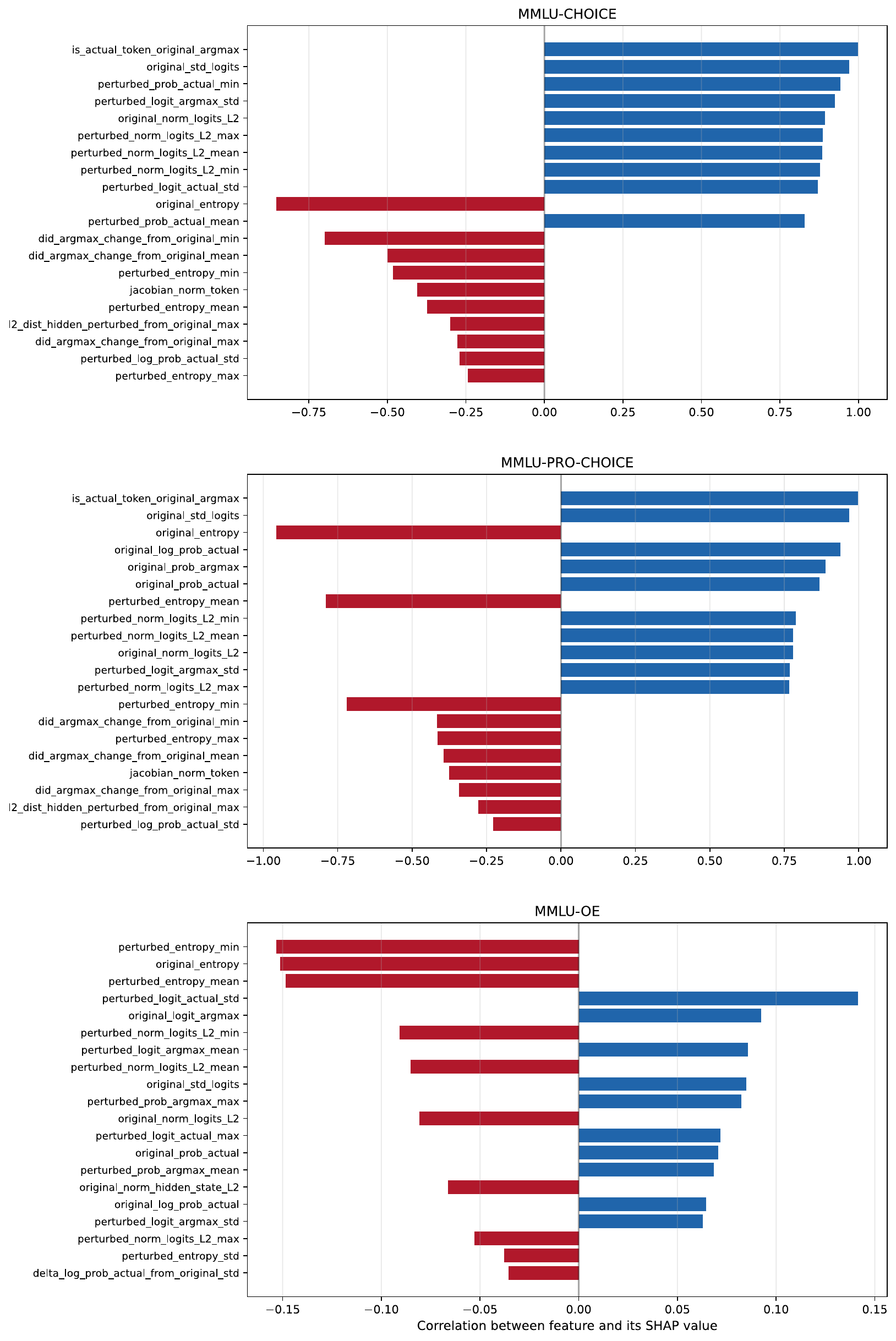}
  \caption{Correlations between feature values and SHAP scores in CCPS on \texttt{Mistral-Small-24B-Instruct-2501} across all datasets. Blue bars denote positive correlations (higher feature values increase prediction ACC), and red bars denote negative correlations.}
  \label{fig:mistral_shap}
\end{figure*}

\begin{figure*}[ht]
  \centering
  \includegraphics[width=0.9\textwidth]{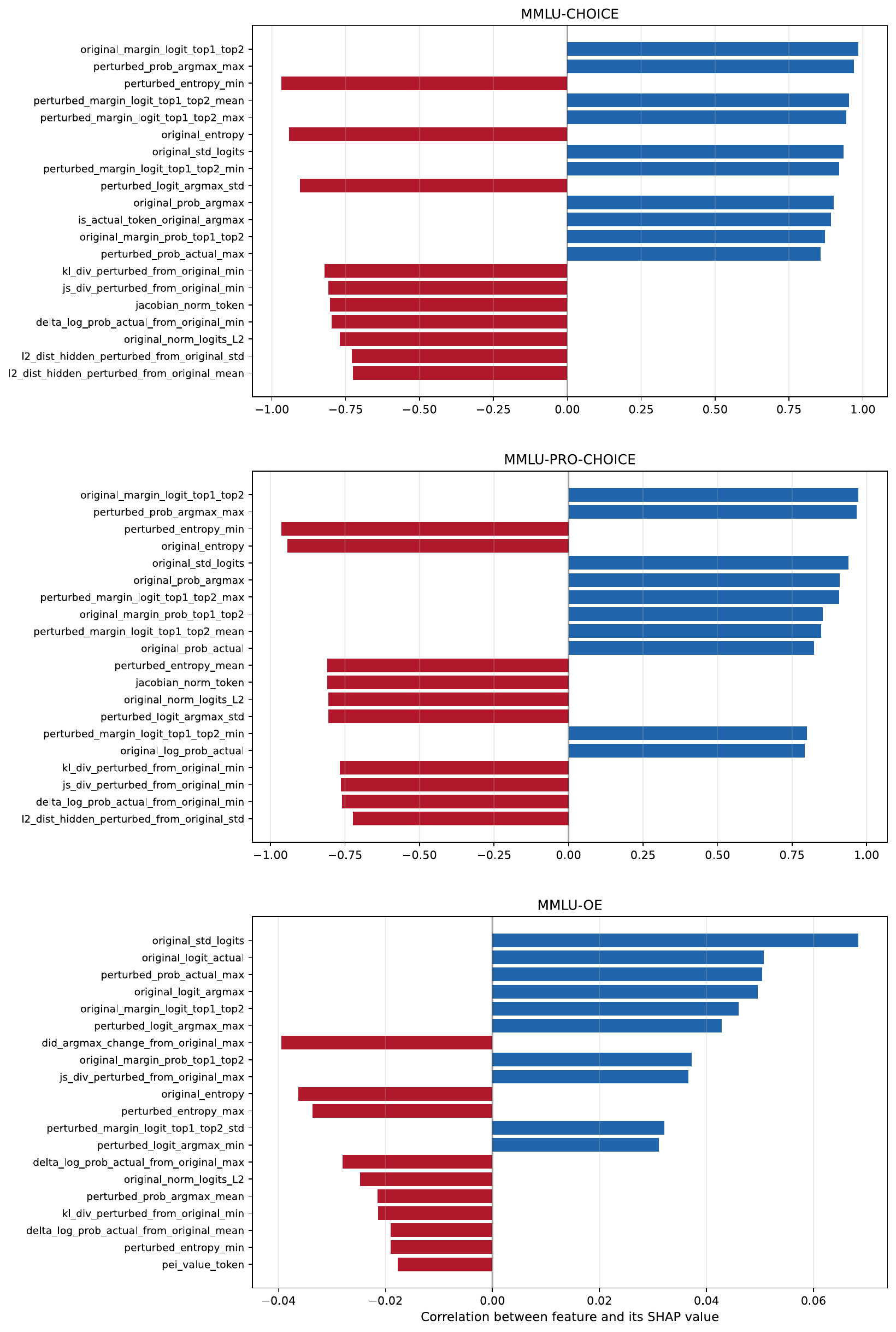}
  \caption{Correlations between feature values and SHAP scores in CCPS on \texttt{Qwen2.5-32B-Instruct} across all datasets. Blue bars denote positive correlations (higher feature values increase prediction ACC), and red bars denote negative correlations.}
  \label{fig:qwen32_shap}
\end{figure*}

To investigate the contributions of various engineered features to the predictions of our CCPS model, we employed SHAP (SHapley Additive exPlanations) \cite{shapley} (MIT License). This analysis utilized a model wrapper around our trained CCPS classifiers and a subset of the respective training data as background references for the \texttt{shap.KernelExplainer} with a logit link function. For Multiple-Choice (MC) models, which take a single feature vector as input, SHAP values directly indicate the importance of each of the $D_f$ features. The resulting "Feature-SHAP Correlation" plots (Figures \ref{fig:llama_shap} through \ref{fig:qwen32_shap} for MC model results) visualize the Pearson correlation between scaled feature values and their SHAP values, where colors typically distinguish positive and negative correlations, indicating how feature magnitudes influence the prediction towards correctness.

Due to the sequential nature of inputs (a matrix of feature vectors per token) for Open-Ended (OE) models, SHAP analysis was adapted to assess feature importance across the initial $N$ tokens (e.g., $N=10$) of an answer. For each feature type, SHAP values were computed based on its influence at these initial positions and then averaged across these $N$ positions to derive an overall impact score. Consequently, the "Feature-SHAP Correlation" plots for OE models (also presented in Figures \ref{fig:llama_shap} through \ref{fig:qwen32_shap} for the respective LLMs' OE results) illustrate the correlation between these position-averaged feature values and their corresponding position-averaged SHAP values.

\subsection{Leave-One-Out Feature Ablation}
\label{app:loo_feature_ablation}
To investigate the utility of our 75 features and address the possibility of redundancy, we performed a comprehensive leave-one-out feature ablation study. For this analysis, we retrained our CCPS model 75 times for each of the four base LLMs, each time with one feature removed, and evaluated the performance on both \texttt{MMLU-CHOICE} and the more challenging \texttt{MMLU-PRO-CHOICE} test sets. The importance of each feature was quantified by the resulting drop in AUROC when it was excluded from the model. Table~\ref{tab:influential_features} summarizes the five most and least influential features shared across all LLMs for each dataset, based on the average drop in AUROC.

\begin{table*}[t]
\centering
\small
\begin{tabular}{l p{0.8\textwidth}}
\toprule
\textbf{Test Dataset} & \textbf{Influential Features} \\
\midrule
\texttt{\textbf{MMLU-CHOICE}} & 
\textbf{Top 5 Most Influential:} \\
& \texttt{perturbed\_logit\_argmax\_mean} \\
& \texttt{did\_argmax\_change\_from\_original\_max} \\
& \texttt{did\_argmax\_change\_from\_original\_mean} \\
& \texttt{pei\_value\_token} \\
& \texttt{perturbed\_logit\_argmax\_std} \\
& \textbf{Top 5 Least Influential:} \\
& \texttt{l2\_dist\_hidden\_perturbed\_from\_original\_std} \\
& \texttt{cosine\_sim\_hidden\_perturbed\_to\_original\_std} \\
& \texttt{original\_norm\_logits\_L2} \\
& \texttt{is\_actual\_token\_original\_argmax} \\
& \texttt{jacobian\_norm\_token} \\
\midrule
\texttt{\textbf{MMLU-PRO-CHOICE}} & 
\textbf{Top 5 Most Influential:} \\
& \texttt{js\_div\_perturbed\_from\_original\_mean} \\
& \texttt{cosine\_sim\_logits\_perturbed\_to\_original\_mean} \\
& \texttt{js\_div\_perturbed\_from\_original\_max} \\
& \texttt{kl\_div\_perturbed\_from\_original\_std} \\
& \texttt{did\_argmax\_change\_from\_original\_mean} \\
& \textbf{Top 5 Least Influential:} \\
& \texttt{perturbed\_prob\_actual\_std} \\
& \texttt{original\_prob\_actual} \\
& \texttt{jacobian\_norm\_token} \\
& \texttt{is\_actual\_token\_original\_argmax} \\
& \texttt{original\_entropy} \\
\bottomrule
\end{tabular}
\caption{Top 5 most and least influential features identified for MMLU-CHOICE and MMLU-PRO-CHOICE. Each feature is shown on a separate line.}
\label{tab:influential_features}
\end{table*}

\vspace{5pt}
\noindent\textbf{The Ablation Confirms that Features Measuring the Response to Perturbation are Most Important} \quad Our analysis confirms the utility of our feature set, as removing any single feature did not improve performance across all model-task combinations, indicating a well-designed, non-redundant set. On \texttt{MMLU-CHOICE}, the most impactful features directly measure the \textit{outcome of the perturbation}, such as whether the model's top prediction changed (\texttt{did\_argmax\_change...}). This validates our core hypothesis that the stability of the LLM's prediction under a targeted challenge is a key confidence signal.

\vspace{5pt}
\noindent\textbf{The Nature of Informative Confidence Signals Shifts with Increasing Task Difficulty} \quad Interestingly, when evaluating on the more challenging \texttt{MMLU-PRO-CHOICE} dataset, the nature of the most influential features shifts. The most vital shared features become those measuring the \textit{holistic distributional change under perturbation}, such as \texttt{js\_div\_perturbed\allowbreak\_from\_original\_mean} (the divergence between original and perturbed output distributions) and \texttt{cosine\_sim\_logits\_perturbed\allowbreak\_to\_original\_mean} (the similarity between original and perturbed logit vectors). This key insight suggests that on harder problems, quantifying the degree of representational shift appears to be a more robust indicator of confidence than tracking the stability of just the single top prediction.
\end{document}